\documentclass[letterpaper]{article} %
\usepackage{aaai2027}
\nocopyright
\usepackage{framed}
\usepackage[hyphens]{url}  %
\usepackage{amssymb}
\usepackage{graphicx} %
\usepackage[disable]{todonotes}
\usepackage[most]{tcolorbox}
\usepackage{xparse}
\usepackage{array}
\newcolumntype{L}[1]{>{\raggedright\arraybackslash}p{#1}}

\newcounter{prompttemplate}[section]
\renewcommand{\theprompttemplate}{%
  \thesection.\arabic{prompttemplate}%
}

\NewDocumentEnvironment{promptbox}{O{} m m}{%
  \refstepcounter{prompttemplate}%
  \label{#3}%
  \begin{tcolorbox}[
    enhanced,
    breakable,
    colback=gray!3,
    colframe=black!50,
    boxrule=0.5pt,
    arc=1mm,
    left=5pt,
    right=5pt,
    top=5pt,
    bottom=5pt,
    fonttitle=\bfseries,
    title={Prompt~\theprompttemplate: #2},
    #1
  ]
}{%
  \end{tcolorbox}
}

\usepackage{natbib}  %
\usepackage{arydshln}
\usepackage{caption} %
\usepackage{algorithm}
\usepackage{algorithmic}
\usepackage{amsmath}
\usepackage{newfloat}
\usepackage{listings}
\DeclareCaptionStyle{ruled}{labelfont=normalfont,labelsep=colon,strut=off} %
\floatstyle{ruled}
\newfloat{listing}{tb}{lst}{}
\floatname{listing}{Listing}

\usepackage{booktabs}
\usepackage{multirow}
\usepackage{adjustbox}
\usepackage{threeparttable}
\usepackage{makecell}
\usepackage{siunitx}
\usepackage{xcolor}
\newif\ifstructurenotes
\structurenotesfalse
\definecolor{neutral}{rgb}{0.20,0.45,0.65}

\usepackage{xspace}

\newcommand{\alps}{\texttt{ALPS}\xspace}
\newcommand{\alpsfull}{\textit{\textbf{A}lign \textbf{L}ong-term by \textbf{P}lanning \textbf{S}hort-term}\xspace}

\newcommand{\dlim}{\mathcal{D}^{\text{suff.}}}
\newcommand{\dtest}{\mathcal{D}^{\text{test}}}
\newcommand{\dfew}{\mathcal{D}^{\text{scarce}}}
\newcommand{\tuglet}{\texttt{TugLet}}
\newcommand{\pharmasim}{\texttt{PharmaSim}}

\newcommand{\sd}[1]{${}_{\pm#1}$}
\definecolor{addblue}{rgb}{0.05,0.30,0.65} %

\newcommand{\td}{TD\xspace}
\newcommand{\tr}{TR\xspace}
\newcommand{\density}{\textit{density}\xspace}
\newcommand{\coverage}{\textit{coverage}\xspace}

\title{To Align Long, Plan Short: Simulating Diverse Profile-Aligned Students by Realizing Short-Term Plans in Long-Horizon Learning Environments}
\author{
    Bahar Radmehr\textsuperscript{\rm 1},
    Jiaxu Zhao\textsuperscript{\rm 1},
    Fatma Bet\"{u}l G\"{u}re\c{s}\textsuperscript{\rm 1},
    Adish Singla\textsuperscript{\rm 2},
    Tanja K\"{a}ser\textsuperscript{\rm 1}
}
\affiliations{
    \textsuperscript{\rm 1}EPFL, Lausanne, Switzerland\\
    \textsuperscript{\rm 2}MPI-SWS, Saarbr\"{u}cken, Germany\\
    \{bahar.radmehr, jiaxu.zhao, fatma.gures, tanja.kaeser\}@epfl.ch, adishs@mpi-sws.org
}

\begin{document}

\maketitle

\begin{abstract}
Simulated students offer a scalable way to test learning platforms and AI tutors before deployment with real students and to augment scarce data for studying learning at scale. Existing LLM-based simulators primarily target short, language-centered interactions. In open-ended learning environments, however, student behavior emerges from patterns across long-horizon interaction trajectories, where turn-level prompting can drift from an assigned behavior profile. We introduce \alps{} (\alpsfull), a hierarchical simulated student conditioned on observable, environment-specific behavior profiles. Rather than selecting every action independently, \alps{} alternates between planning profile-aligned task segments and generating the corresponding actions. We evaluate \alps{} in both data-sufficient and data-scarce settings, where supervision comes either from real student trajectories or from a teacher LLM. Across both settings, we compare in-context learning (ICL), supervised fine-tuning (SFT), reward-ranked fine-tuning (RAFT), and their sequential combination (SFT$\rightarrow$RAFT). Experiments in an inquiry-based physics game and a pharmacy simulation evaluate behavioral \density{} and \coverage{} against held-out real data, and profile consistency using a human-aligned LLM judge. Across both settings, \alps{} consistently outperforms turn-level generation, achieving comparable \density{} and judge pass rates under scarce supervision, although \coverage{} remains lower. These results show that hierarchical planning enables faithful, profile-aligned simulation over long interaction horizons, even with limited real student data.
\end{abstract}
\section{Introduction}
Interactive learning technologies increasingly rely on AI to provide personalized support, evaluate learning experiences, and guide adaptive interventions. Developing and validating such systems requires realistic student behavior: students who explore, pursue different strategies, and make mistakes, rather than merely records of success and failure. Collecting such behavior at scale is difficult. Classroom access, cost, and privacy constraints result in datasets that are small and skewed, with rare behavior patterns, often those of the students an intervention aims to support, being the least represented \cite{DBLP:journals/aiedu/BakerH22}.
Simulated learners, computational models that emulate how students interact with learning environments
\cite{DBLP:journals/aiedu/KaserA24}, offer a scalable alternative for testing learning platforms and adaptive interventions before deployment
\cite{VanLehn1994,DBLP:conf/sigcse/RobinsonJR18,DBLP:conf/edm/NazaretskyHA19}
and augmenting datasets with rare but pedagogically important behaviors
\cite{bjet/AI-for-data-generation,lak/synthetic-data,DBLP:conf/lak/LiuDVKJS25}. To fulfill this promise, simulated learners must be \textit{faithful}, reproducing realistic student behavior, and \textit{diverse}, representing the range of learner behaviors. 

Earlier simulated learners were based on cognitive, rule-based, or probabilistic models learned from interaction logs
\cite{DBLP:conf/edm/LiCKM11,DBLP:conf/edm/MacLellanHPK16,
Corbett2005KnowledgeTM,DBLP:conf/edm/FauconKD16,
DBLP:conf/edm/BotelhoAH16}. While these approaches provide interpretable and controllable models of student behavior, they typically rely on a small number of predefined states or parameters, limiting their ability to capture the richness of real student behavior
\cite{DBLP:journals/aiedu/KaserA24}. More recently, large language models (LLMs) have enabled richer simulated learners. Existing LLM-based simulators, however, have primarily focused on short, language-centered interactions, including tutoring dialogues
\cite{DBLP:conf/chi/JinYPLWK25,neurips2023gaied_38_schmucker,
DBLP:conf/naacl/ZhangZYGZHJCLLHL25,DBLP:conf/chi/XuWPDHZ25,
DBLP:journals/corr/abs-2601-04025,DBLP:journals/corr/abs-2605-30051},
misconceptions and errors
\cite{Ross2025,Fernandez2024,Sonkar2024,Daheim2024},
assessment responses
\cite{Benedetto2024,DBLP:journals/corr/abs-2507-05129,Nguyen2025,HeYueya2024,Lu2024}, and
mathematical or programming practices
\cite{Ross2025b,DBLP:conf/edm/NguyenTS24} (see Appendix~\ref{ap:related}). Whether they can generate coherent, diverse behavioral trajectories in longer horizons remains largely unexplored.

Our work focuses on simulating learners in open-ended learning environments (OELEs), where students engage in long exploratory interaction trajectories rather than short exchanges. In this setting, behavior emerges from patterns across entire trajectories rather than individual actions. This shift from turn-level responses to trajectory-level behavior introduces \emph{three} key challenges. First, maintaining a consistent behavior profile becomes more difficult. Because trajectories span hundreds of interactions, LLMs tend to drift from their assigned profiles \cite{DBLP:journals/corr/abs-2511-00222}.  Second, existing work typically conditions simulated learners on generic attributes such as knowledge states, personality traits, or demographics \cite{liu-etal-2024-personality,DBLP:conf/chi/JinYPLWK25}. These attributes describe the student, but not how the student behaves in a particular learning environment, leaving no observable basis for verifying generated trajectories. Many pedagogically meaningful behaviors are also environment-specific, arising from the available actions and constraints rather than general learner characteristics. Finally, the scarcity of student data that motivates simulated learners also constrains their development. New environments often provide only a handful of trajectories or none at all, making it difficult to learn environment-specific behavior profiles. Existing work shows that prompting performs poorly under such limited supervision \cite{DBLP:journals/corr/abs-2601-04025}, leaving the development of faithful and diverse simulated learners an open problem.

To address these challenges, we formulate student simulation as profile-conditioned generation of long interaction trajectories. Rather than conditioning on generic learner attributes, we derive environment-specific behavior profiles from real student trajectories (or adopt established profiles from prior work), providing an observable target for both generation and evaluation. We introduce \alps{} (\alpsfull), a hierarchical simulator inspired by planning-based agents for long-horizon tasks \cite{Sutton1999Options,He2024WordsToActions,Zhao2024EPO}.  Instead of generating trajectories action by action, \alps{} decomposes them into semantically meaningful segments. A segment planner specifies the behavior for each upcoming segment, and an action generator realizes this plan via individual actions in the environment. This hierarchical decomposition maintains coherent behavior over long trajectories while preserving flexibility within each segment. We adapt \alps{} to target profiles under two supervision settings: a \emph{data-sufficient} setting, where real trajectories provide direct supervision, and a \emph{data-scarce} setting, where a teacher model supplies supervision. In both settings, we compare in-context learning (ICL), supervised fine-tuning (SFT), and reward-ranked fine-tuning (RAFT)~\cite{DBLP:journals/tmlr/Dong0GZCPDZS023}.

We evaluate \alps{} in two OELEs, an inquiry-based physics game (\tuglet)~\cite{DBLP:conf/lak/KaserHS17} and a pharmacy simulation (\pharmasim)~\cite{PharmaSimText2025}, against turn-level generation. We evaluate behavioral \density{} and \coverage{} against held-out real data, and profile consistency using a human-aligned LLM judge. Across both environments and supervision settings, \alps{} consistently produces more profile-consistent trajectories than turn-level generation, with further gains from adaptation.
Our main contributions are:
\begin{enumerate}
\item \textbf{A formulation of student simulation as profile-conditioned
generation} of long interaction trajectories, using observable, environment-specific behavior profiles.
\item \textbf{\alps{}, a hierarchical planner-generator architecture} that maintains coherent behavior over long trajectories through segment-level planning.
\item \textbf{Adaptation methods for two data settings}, pairing ICL, SFT, and RAFT with real or teacher supervision.
\item \textbf{A comprehensive empirical evaluation} demonstrating improved \density{} and profile consistency across two OELEs in data-sufficient and data-scarce settings.
\end{enumerate}
Our implementation and evaluation code will be released publicly.

\section{Method}
\label{sec:method}
\begin{figure*}[t]
    \centering
    \includegraphics[width=\linewidth]{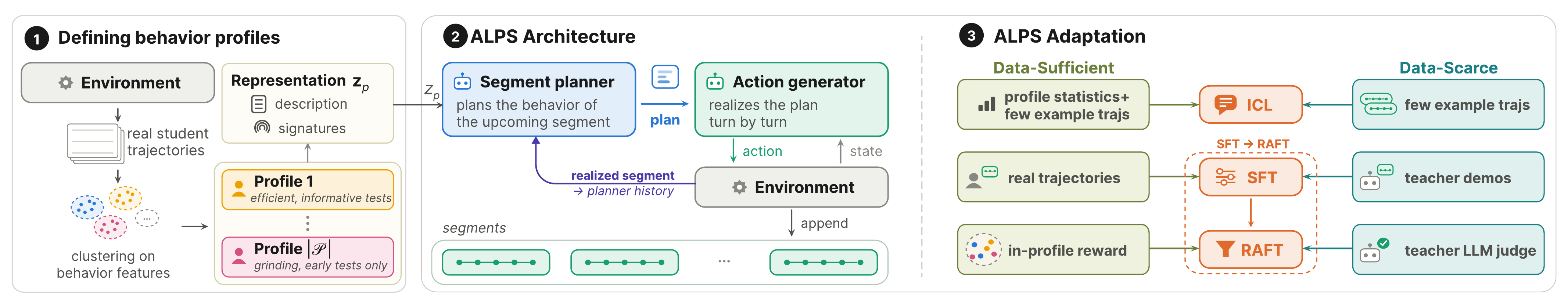}
    \captionsetup{skip=3pt}
    \caption{Overview of the proposed simulation framework. (1) Real student trajectories are clustered into environment-specific behavior profiles. (2) Conditioned on a target profile, \alps{} generates long trajectories through a hierarchical planner-generator architecture. (3) When real trajectories are available, \alps{} is adapted under both data-sufficient and data-scarce settings using ICL, SFT, and RAFT.}
\label{fig:overview}
\end{figure*}

Our goal is to generate simulated student populations that faithfully reproduce the behavioral diversity observed in OELEs. We first derive environment-specific behavior profiles by clustering student trajectories using trajectory-level features, following prior work on student profiling \cite{Amershi_Conati_2009,DBLP:journals/aiedu/KaserS20,DBLP:conf/aied/ShvedBMKK24}. These profiles condition \alps{}, which combines a hierarchical planner-generator architecture for long-horizon trajectory generation with adaptation for data-scarce and data-sufficient settings. We study adaptation through ICL, SFT, and RAFT.

\subsection{Defining Behavior Profiles}
\label{sec:behavior-profiles}
We derive behavior profiles from a set
$\dlim=\{\tau_i\}_{i=1}^{N}$ of real student trajectories, available in the data-sufficient setting (Figure~\ref{fig:overview}, step~1). Following the behavior-based clustering approach of \cite{DBLP:journals/aiedu/KaserS20}, each trajectory is represented using action-specific behavioral features capturing the intensity, quality, and temporal distribution of student actions. We compute action-specific similarities from these features using Euclidean distance, embed their sum into a Euclidean space using multidimensional scaling, and cluster with $k$-means, selecting the number of clusters via the Bayesian Information Criterion (BIC) averaged over $200$ seeds.
Each trajectory is thereby assigned a behavior profile
$p \in \mathcal{P}$, where $\mathcal{P}$ denotes the environment's set
of profiles. 

We then represent each behavior profile $p$ as $z_p = (d_p, \mathcal{S}_p)$ to condition the simulated student. Here, $d_p$ is a concise description of the behavioral patterns characterizing the profile and $\mathcal{S}_p$ is a set of
behavioral signatures. We distinguish two types of signatures: \emph{hard signatures}, behaviors exhibited by all students in the profile, and \emph{soft signatures}, tendencies shared by at least two thirds of the students, a threshold fixed \emph{a priori}. We prompt an LLM with cluster statistics and representative trajectories to
generate an initial description and set of signatures, which a human domain expert subsequently reviews and refines. For environments with profiles established in prior work, we reuse the published descriptions as $d_p$ and derive the behavioral signatures using the same procedure.

\subsection{\alps}
\label{sec:alps}

\alps{} simulates a student of a target behavior profile $p$ in a long-horizon learning environment. It consists of a hierarchical planner-generator architecture that generates coherent trajectory segments while maintaining alignment with $z_p$, together with adaptation for both supervision settings.

\subsubsection{Architecture.}
\label{sec:alps-structure}
As shown in Figure~\ref{fig:overview} (Step~2), \alps{} generates long trajectories by alternating between a segment planner, which maintains consistency with the target behavior profile, and an action generator, which realizes each planned segment through individual actions. A trajectory
\(
\tau=(s_1,a_1,\ldots,s_T,a_T)
\)
is generated segment by segment. A deterministic, environment-specific segmentation map partitions the trajectory into contiguous segments
\(
\operatorname{seg}_{\mathcal{E}}(\tau)=(\xi_1,\ldots,\xi_M),
\)
whose boundaries follow the natural structure of the learning task, such as transitions between environment modes or specific action types. The rules are fixed across profiles and applied identically to real and generated trajectories, producing only a handful of segments even for trajectories containing hundreds of actions.

At the beginning of segment $\xi_k$, the segment planner generates a structured behavior plan 
\(
B_k \sim \pi_{\mathrm{plan}}
(\cdot \mid z_p,s_{t_k},\hat{B}_{<k}),
\)
where $s_{t_k}$ is the current environment state and $\hat{B}_{<k}$ summarizes the behavior generated in preceding segments (Prompt ~\ref{prompt:segment-planner}). The environment description is provided as fixed context and is omitted from the notation. The plan specifies target behavior statistics, whether the trajectory should terminate, and the segment termination condition, while leaving action selection to the generator. To ensure valid plans, we deterministically correct occasional inconsistencies, for example when the planned number of relevant actions of a type exceeds the planned total number of
actions of that type.

Within segment $\xi_k$, the action generator selects one action per turn:
\(
a_t \sim \pi_{\mathrm{gen}}
\bigl(\cdot \mid B_k,s_t,\mathcal{A}(s_t),\hat{B}_{k,<t}\bigr),
\)
where $\mathcal{A}(s_t)$ contains the actions available in the current state and $\hat{B}_{k,<t}$ summarizes the behavior generated within the current segment (Prompt ~\ref{prompt:action-generator}). The generator is conditioned on the target profile only indirectly through $B_k$, which serves as a soft behavioral target because some planned actions may be infeasible in the current state. If the generated action is invalid, the model retries once; if the retry also fails, it uniformly samples an allowed action from $\mathcal{A}(s_t)$ that does not terminate the trajectory. At segment completion, the generated actions are deterministically summarized into $\hat{B}_k$ for the next planner call. The planner and generator alternate until the planner terminates the trajectory or the action budget is exhausted. Appendix~\ref{app:segmentation} describes the environment-specific segmentation rules and behavior plan fields.

\subsubsection{Adaptation.}
\label{sec:alps-adaptation}
Adaptation uses the available real data to improve alignment with the target behavior profile beyond prompting with $z_p$ alone (Figure~\ref{fig:overview}, Step~3). We compare ICL, SFT, and RAFT, applying RAFT either to the base model or to the SFT-adapted model (SFT$\rightarrow$RAFT). For SFT and RAFT, we train one LoRA adapter per profile, component, and environment. The three methods differ between data settings only in the context provided for ICL, the demonstrations used for SFT, and the reward used by RAFT.

ICL requires no training and conditions the planner or action generator directly on examples from the corresponding data setting. For SFT, each demonstration trajectory $\tau$ is partitioned into segments using the rules from Section~\ref{sec:alps-structure}. Each segment $\xi_k$ provides one planner example, with a target plan $B_k$ extracted from the observed actions. Each action $a_t$ provides one action-generator example conditioned on $B_k$ and the interaction context at turn $t$. Because trajectories differ in length, we up-sample underrepresented later segments, and train both components with token-level cross-entropy.

RAFT uses the same example format in a generate--filter--train cycle. The simulator generates complete trajectories, and a reward selects those that match the target profile. After removing near-duplicates in the feature space, the accepted ones are converted into training examples. The planner target for each accepted segment $\xi_k$ is its generated summary $\hat{B}_k$, so the planner learns from generated rather than intended behavior. Action-generator examples are retained only when $\hat{B}_k$ matches the original plan $B_k$ within a tolerance, so the generator learns from segments that followed their plans.

In the data-sufficient setting, we augment the profile representation $z_p$ with quantitative summary statistics estimated from the real trajectories in $\dlim$, for example the range of test counts associated with grinding behavior. ICL also conditions the planner on few-shot examples from the intended profile. The profile-labeled trajectories in $\dlim$ provide the demonstrations for SFT and the data for constructing the \emph{in-profile} reward for RAFT. We implement the in-profile reward using a weighted $k$-nearest-neighbor (KNN) classifier over the behavioral feature space (Section~\ref{sec:behavior-profiles}). We fit the classifier on $\dlim$, with neighbor votes weighted by inverse class frequency and inverse distance and $k$ selected by leave-one-out cross-validation on balanced accuracy. \emph{In-profile} accepts a generated trajectory $\tau$ when its classifier score for the target profile exceeds $0.5$ and it passes a within-profile outlier gate. The gate requires that the distance from $\tau$ to the real trajectories of the target profile does not exceed the 95th percentile of the distances observed among real students of that profile.

The data-scarce setting provides three real trajectories per profile in $\dfew$. Here, ICL conditions both components on the segmented trajectories in $\dfew$ as in-context examples. Because little real supervision is available, a teacher LLM provides the reward for RAFT and demonstrations for SFT.

As reward, the teacher provides the \emph{teacher reward} (\tr). Given $z_p$, the target examples in $\dfew$, contrast examples from the two most similar profiles, and a candidate trajectory $\tau$, it evaluates the trajectory using a fixed checklist of binary questions. The questions cover behavioral realism, overall profile match, and the presence of signatures $\mathcal{S}_p$. Following recent work, each answer includes a one-sentence justification citing relevant trajectory segments because checklist- and rubric-based evaluation is more reliable than scalar judgments \cite{gunjal2025rubrics}. It also receives a feature table comparing the candidate with target and contrast examples, avoiding the need for the LLM to count actions from the trajectory. No feedback is returned to the generator (Prompt~\ref{prompt:teacher-assessment}).
Because the LLM may misjudge individual items, \tr tolerates a small number of failed checks per question category. We calibrate this tolerance on $\dfew$ per environment. Automated aggregation applies the same acceptance rule across profiles while keeping the reason for each rejection inspectable (See~\ref{app:rewards}). The same teacher generates demonstrations for SFT. Given $z_p$ and the segmented target-profile examples in $\dfew$, it generates candidate trajectories using the planner--generator architecture (Prompt~\ref{prompt:teacher-demonstration}). Candidates accepted by \tr form the teacher demonstration set $\td$ used for SFT.

\section{Experimental Protocol}
\label{sec:experimental-protocol}
We evaluate \alps{} in two OELEs under the
data-sufficient and data-scarce settings described in Section~\ref{sec:method}. For each condition, we generate simulated populations for every behavior profile and compare them with held-out real students using measures of \density{}, \coverage{} and an LLM judge evaluation.

\subsection{Learning Environments and Data}
\label{sec:environments-data}
All experiments use real interaction data from two open-ended environments that differ in interaction style and trajectory length: \tuglet{}, an inquiry-based game in which students experiment with tug-of-war contests to discover that motion occurs toward the side with the greater net force~\cite{DBLP:journals/aiedu/KaserS20}, and \pharmasim{}, a simulated pharmacy environment for practicing diagnostic conversations~\cite{PharmaSimText2025}. Trajectories span tens of actions in \pharmasim{} (median 31, maximum 61) and up to 185 actions in \tuglet{} (median 41, 90th percentile 124), requiring profiles to remain coherent over interactions of very different lengths. \tuglet{} contains six behavior profiles adopted from prior work~\cite{DBLP:journals/aiedu/KaserS20}, whereas we derived the five \pharmasim{} profiles using the procedure of Section~\ref{sec:behavior-profiles} (See Appendix~\ref{app:environments}).

The \tuglet{} dataset comprises trajectories from 249 students, and the \pharmasim{} dataset  from 234 students. We partition each into a development set $\dlim$ (140 students in both environments) and a held-out test set $\dtest$ (109 students in \tuglet{}, 94 in \pharmasim{}), stratified by the profile labels so that every profile contains at least five test students. The development set $\dlim$ provides all profile definitions, demonstrations, and rewards. For the data-scarce setting, we construct $\dfew$ by sampling three trajectories per profile from $\dlim$. The held-out set $\dtest$ is used exclusively for evaluation. Its trajectories are labeled by a weighted KNN classifier (Section~\ref{sec:alps-adaptation}) trained on $\dlim$, and no test-set information is used for training, adaptation, or rewards.

\subsection{Models and Conditions}
\label{sec:models-conditions}
We use Qwen3-8B~\cite{DBLP:journals/corr/abs-2505-09388} as the
simulated student. GPT-OSS-120B~\cite{DBLP:journals/corr/abs-2508-10925} serves as the teacher LLM in the data-scarce setting and as a prompted ICL reference student. All adapters are trained on Qwen3-8B.

We compare \alps{} against a \emph{turn-level student}, a single LLM invoked with a fresh prompt at each turn to select the next action. Each
prompt contains $z_p$, the environment description, the current state,
the valid actions, and a running summary of the generated behavior. The
turn-level student has no segment planner and is adapted using the same
methods as \alps{}, with training
examples constructed per turn (Prompt~\ref{prompt:turn-level}).
We evaluate ICL, SFT, RAFT, and SFT$\rightarrow$RAFT using the
setting-specific context, demonstrations, and rewards defined in Section \ref{sec:alps-adaptation}.

\subsection{Adaptation and Generation}
\label{sec:protocol}
For each fine-tuning run, we split the available examples 80/20 into training and validation sets and retain the checkpoint with the lowest validation loss. RAFT uses a single generate--filter--train cycle with the setting-specific reward, accepting up to 100 trajectories per profile and architecture. In the data-scarce setting, candidates are generated using the examples in $\dfew$, but these are excluded from the training inputs. Adapted models are therefore trained and evaluated without in-context examples.

For each condition, we generate 100 trajectories per profile at temperature $0.7$ using identical decoding parameters. A trajectory terminates when the learning task ends or the environment's action budget is exhausted. Trajectories reaching the budget limit are completed using the same deterministic procedure for both architectures. We repeat the full pipeline three times with different training, generation, and evaluation seeds and report the mean and standard deviation.

\subsection{Evaluation}
\label{sec:evaluation}
We measure behavioral \density{} and \coverage{} in the feature space of
Section~\ref{sec:behavior-profiles} using Clipped Density (\density{}) and Clipped
Coverage (\coverage{}), two standard metrics for evaluating generative
models~\cite{salvy2026enhanced}. Density measures how frequently generated trajectories fall within the local neighborhoods of real trajectories.
Coverage measures the fraction of real trajectories whose neighborhood
contains at least one generated trajectory. We use $N{=}3$ neighbors by
default and reduce this to $N{=}1$ for profiles with fewer than 10 test
students. For profiles with few students, larger neighborhoods overlap substantially, causing coverage to saturate even when generated trajectories cover only a small portion of the profile. We report both the mean across profiles and the worst-profile score, since strong average performance can hide failure on an individual profile.

Feature-based metrics may miss unnatural behavior within individual
actions. We therefore also evaluate every generated trajectory $\tau$ using the
strict verdict
\(
V = G \cdot H \cdot R
\) (See Appendix~\ref{app:judge}).
The blind profile-identification component $G$ measures whether a human-aligned LLM judge correctly infers the intended profile without seeing its label. The signature component $H$ verifies that $\tau$ satisfies every signature of the intended profile, while the rule-based component $R$ detects mechanical generation artifacts using deterministic checks. Two domain experts constructed the artifact list by reviewing generated trajectories and verifying each irregularity against the real data. The judge prompts, inputs, and components are independent of the adaptation reward \tr{}, ensuring no condition is evaluated using its training signal (Prompt~\ref{prompt:evaluation-judge}).

The thresholds in $R$ are set at the observed extremes of real students, and the checks in $H$ are calibrated so that every real student passes the complete verdict. We report the pass rate on generated trajectories, i.e., the fraction satisfying the complete profile and realism rubric.

We aligned the LLM-based components with human judgments using 50 generated trajectories per environment, sampled across architectures and adaptation methods. Two raters independently annotated 25 \pharmasim{} and 18 \tuglet{} trajectories. Human inter-rater agreement, measured by Cohen's $\kappa$, was high for blind profile identification (Cohen's $\kappa=1.00$ and $0.87$) and full signature coverage ($\kappa=0.91$ and $1.00$), respectively. One rater annotated the remaining trajectories. The judge agreed with human blind profile labels on all annotated trajectory in both environments ($\kappa=1.00$). Agreement on hard-signature checks was high as well ($\kappa=0.83$ and $0.94$).

\section{Results}
\label{sec:results}
We evaluated \alps{} across two learning environments under two data settings. \textbf{Exp1} considered the \emph{data-sufficient} setting, where real profile-labeled trajectories directly supervised model adaptation. \textbf{Exp2} considered the \emph{data-scarce} setting, where only three real trajectories per profile were available and a teacher LLM supplied the remaining supervision. In both experiments, we compared turn-level generation and \alps{} using ICL, SFT, RAFT, and SFT$\rightarrow$RAFT. For each experiment, we report feature-based metrics computed on the held-out $\dtest$  along with the human-aligned LLM judge results and ablation studies.

 \begin{table*}[t]
    \centering
    \footnotesize
    \setlength{\tabcolsep}{3pt}
    \renewcommand{\arraystretch}{0.9}
    \setlength{\aboverulesep}{0.05ex}
    \setlength{\belowrulesep}{0.15ex}
    {\footnotesize
    \begin{tabular*}{\textwidth}{@{\extracolsep{\fill}} c l l l l cc cc cc @{}}
        \toprule[1.2pt]
        & Model & Architecture & Adaptation & Signal Source &
        \multicolumn{2}{c}{Clipped Density$\uparrow$} &
        \multicolumn{2}{c}{Clipped Coverage$\uparrow$} &
        \multicolumn{2}{c}{Judge pass rate $\uparrow$} \\
        \cmidrule(lr){6-7} \cmidrule(lr){8-9} \cmidrule(l){10-11}
        & & & & & mean & worst & mean & worst & mean & worst \\
        \midrule[1pt]
\multirow{12}{*}{\rotatebox[origin=c]{90}{\textit{PharmaSim}}}
        & \multirow{2}{*}{GPT-OSS} & Turn-level & ICL & 3-shot
        & 22.5\sd{0.3} & 0.0\sd{0.0}
        & 2.2\sd{0.7} & 0.0\sd{0.0}
        & 39.6\sd{0.2} & 0.0\sd{0.0} \\
        \cmidrule(lr){4-11}
        & & \alps & ICL & 3-shot
        & \textbf{76.0\sd{2.3}}$^\ast$ & \textbf{47.9\sd{4.5}}
        & \textbf{9.7\sd{0.3}}$^\ast$ & \textbf{3.0\sd{0.0}}
        & \textbf{95.6\sd{0.8}}$^\ast$ & \textbf{86.9\sd{1.5}} \\
        \cmidrule(l){3-11}
        & \multirow{9}{*}{Qwen3} & \multirow{4}{*}{Turn-level}
        & ICL & 3-shot
        & 37.4\sd{0.6} & 0.0\sd{0.0}
        & 4.2\sd{1.3} & 2.0\sd{1.4}
        & 46.6\sd{0.3} & 1.0\sd{0.8} \\
        & & & RAFT & in-profile & 21.1\sd{2.5} & 0.0\sd{0.0} & 13.7\sd{6.0} & 0.0\sd{0.0} & 28.1\sd{2.2} & 0.0\sd{0.0} \\
        & & & SFT & $\dlim$ & 37.7\sd{4.3} & 8.6\sd{5.2} & 43.0\sd{6.5} & 6.3\sd{2.8} & 53.2\sd{5.6} & 23.8\sd{11.3} \\
        & & & SFT${\scriptstyle\rightarrow}$RAFT & $\dlim{,}$ in-profile & 29.3\sd{8.8} & 2.9\sd{2.9} & 27.2\sd{14.1} & 2.9\sd{2.4} & 44.3\sd{7.7} & 10.9\sd{5.2} \\
        \cmidrule(lr){4-11}
        & & \multirow{4}{*}{\alps} & ICL & 3-shot
        & 30.7\sd{0.8} & 5.2\sd{2.6}
        & 8.3\sd{1.4} & 3.0\sd{0.0}
        & 41.9\sd{1.2} & 8.7\sd{3.4} \\
        & & & RAFT & in-profile & 36.6\sd{5.9}$^\ast$ & 8.3\sd{3.3} & 7.4\sd{1.3} & 3.0\sd{0.0} & 68.2\sd{2.2}$^\ast$ & 37.7\sd{5.0} \\
        & & & SFT & $\dlim$ & 64.9\sd{6.0}$^\ast$ & 30.8\sd{6.3} & 51.0\sd{7.4} & 16.2\sd{0.9} & 72.7\sd{5.2}$^\ast$ & 34.7\sd{18.7} \\
        & & & SFT${\scriptstyle\rightarrow}$RAFT & $\dlim{,}$ in-profile & \textbf{69.7\sd{3.1}}$^\ast$ & \textbf{33.4\sd{5.8}} & \textbf{58.1\sd{2.8}}$^\ast$ & \textbf{21.2\sd{5.0}} & \textbf{75.8\sd{3.8}}$^\ast$ & \textbf{41.6\sd{20.4}} \\
        \specialrule{1.2pt}{0.2ex}{0.2ex}
\multirow{12}{*}{\rotatebox[origin=c]{90}{\textit{\tuglet{}}}}
        & \multirow{2}{*}{GPT-OSS} & Turn-level & ICL & 3-shot & 48.7\sd{4.0} & 2.0\sd{2.0} & \textbf{48.8\sd{7.8}} & 3.0\sd{0.0} & 20.6\sd{0.3} & 0.0\sd{0.0} \\
        \cmidrule(lr){4-11}
        & & \alps & ICL & 3-shot & \textbf{83.9\sd{1.1}}$^\ast$ & \textbf{42.4\sd{6.5}} & 34.9\sd{6.7} & \textbf{9.9\sd{0.0}} & \textbf{69.7\sd{0.8}}$^\ast$ & \textbf{11.0\sd{1.6}} \\
        \cmidrule(l){3-11}
        & \multirow{9}{*}{Qwen3} & \multirow{4}{*}{Turn-level} & ICL & 3-shot & 48.4\sd{2.3} & 0.4\sd{0.6} & \textbf{69.1\sd{0.2}} & 1.0\sd{1.4} & 6.9\sd{0.9} & 0.0\sd{0.0} \\
        & & & RAFT & in-profile & 5.6\sd{1.0} & 0.0\sd{0.0} & 5.3\sd{0.5} & 0.0\sd{0.0} & 5.2\sd{0.5} & 0.0\sd{0.0} \\
        & & & SFT & $\dlim$ & 28.9\sd{1.2} & 0.0\sd{0.0} & 11.0\sd{0.4} & 0.7\sd{0.9} & 19.2\sd{1.1} & 0.0\sd{0.0} \\
        & & & SFT${\scriptstyle\rightarrow}$RAFT & $\dlim{,}$ in-profile & 31.8\sd{4.3} & 2.4\sd{3.4} & 10.8\sd{0.9} & 3.0\sd{2.9} & 23.8\sd{3.1} & 0.3\sd{0.5} \\
        \cmidrule(lr){4-11}
        & & \multirow{4}{*}{\alps} & ICL & 3-shot & 15.7\sd{0.2} & 4.2\sd{2.4} & 13.5\sd{5.4} & 3.0\sd{0.0} & 15.4\sd{0.4} & 3.3\sd{1.2} \\
        & & & RAFT & in-profile & 34.8\sd{4.4}$^\ast$ & 2.1\sd{1.3} & 7.5\sd{1.3}$^\ast$ & 4.0\sd{0.8} & 25.9\sd{3.2}$^\ast$ & 4.3\sd{2.5} \\
        & & & SFT & $\dlim$ & 46.3\sd{6.3}$^\ast$ & \textbf{22.2\sd{4.6}} & 47.7\sd{5.1}$^\ast$ & \textbf{16.8\sd{4.0}} & \textbf{47.6\sd{2.2}}$^\ast$ & \textbf{27.7\sd{4.8}} \\
        & & & SFT${\scriptstyle\rightarrow}$RAFT & $\dlim{,}$ in-profile & \textbf{52.9\sd{7.1}}$^\ast$ & 9.1\sd{4.1} & 46.4\sd{7.3}$^\ast$ & 5.0\sd{0.8} & 41.2\sd{5.3}$^\ast$ & 11.0\sd{7.3} \\
        \bottomrule[1.2pt]
    \end{tabular*}
    }%
    \captionsetup{skip=3pt}
    \caption{\textbf{Exp1: Data-sufficient setting.}
Simulation quality in \pharmasim{} and \tuglet{}, reported as percentages
(mean$_{\pm\mathrm{std}}$ over three seeds). \emph{Mean} averages performance
across behavior profiles, and \emph{worst} reports the lowest-performing
profile. $\dlim$ denotes the real profile-labeled trajectories used for SFT,
and \emph{in-profile} denotes trajectories accepted by the RAFT reward and
used for fine-tuning. Bold and underlined values indicate the best and
second-best results, respectively, computed separately for GPT-OSS and Qwen3
within each environment. Almost all column-wise best result is achieved by an
\alps{} configuration. $\ast$ indicates that \alps{} significantly outperforms its turn-level counterpart ($p<0.05$).}
    \label{tab:results-limited}
\end{table*}

\begin{table*}[t]
    \centering
    \footnotesize
    \setlength{\tabcolsep}{3pt}
    \renewcommand{\arraystretch}{0.94}
    \setlength{\aboverulesep}{0.05ex}
    \setlength{\belowrulesep}{0.15ex}
    {\footnotesize
    \begin{tabular*}{\textwidth}{@{\extracolsep{\fill}} c l l l l cc cc cc @{}}
        \toprule[1.2pt]
        & Model & Architecture & Adaptation & Signal Source &
        \multicolumn{2}{c}{Clipped Density$\uparrow$} &
        \multicolumn{2}{c}{Clipped Coverage$\uparrow$} &
        \multicolumn{2}{c}{Judge pass rate $\uparrow$} \\
        \cmidrule(lr){6-7} \cmidrule(lr){8-9} \cmidrule(l){10-11}
        & & & & & mean & worst & mean & worst & mean & worst \\
        \midrule[1pt]

        \multirow{12}{*}{\rotatebox[origin=c]{90}{\textit{PharmaSim}}}
        & \multirow{2}{*}{GPT-OSS}
        & Turn-level
        & ICL
        & $\dfew$
        & 22.9\sd{0.2}
        & 0.0\sd{0.0}
        & 1.9\sd{0.2}
        & 0.0\sd{0.0}
        & 39.7\sd{0.1}
        & 0.0\sd{0.0} \\

        \cmidrule(lr){4-11}

        & & \alps
        & ICL
        & $\dfew$
        & \textbf{83.7\sd{0.7}}$^\ast$
        & \textbf{18.4\sd{3.3}}
        & \textbf{25.9\sd{0.2}}$^\ast$
        & \textbf{3.0\sd{0.0}}
        & \textbf{90.5\sd{0.7}}
        & \textbf{67.0\sd{2.2}} \\

        \cmidrule(l){3-11}

        & \multirow{9}{*}{Qwen3}
        & \multirow{4}{*}{Turn-level}
        & ICL
        & $\dfew$
        & 36.8\sd{0.4}
        & 0.0\sd{0.0}
        & 9.5\sd{8.7}
        & 0.7\sd{0.9}
        & 46.3\sd{0.8}
        & 0.7\sd{0.5} \\

        & & & RAFT
        & TR
        & 42.0\sd{1.4}
        & 0.0\sd{0.0}
        & 14.0\sd{7.2}
        & 2.0\sd{0.0}
        & 66.3\sd{2.9}
        & 0.7\sd{0.5} \\

        & & & SFT
        & TD
        & 30.0\sd{4.8}
        & 0.0\sd{0.0}
        & 2.4\sd{0.2}
        & 0.7\sd{0.9}
        & 46.3\sd{0.7}
        & 0.7\sd{0.5} \\

        & & & SFT${\scriptstyle\rightarrow}$RAFT
        & TD${\scriptstyle\rightarrow}$TR
        & 30.5\sd{7.3}
        & 0.0\sd{0.0}
        & 2.3\sd{0.2}
        & 0.7\sd{0.9}
        & 46.4\sd{0.7}
        & 0.7\sd{0.5} \\

        \cmidrule(lr){4-11}

        & & \multirow{4}{*}{\alps}
        & ICL
        & $\dfew$
        & 37.4\sd{1.7}
        & 3.2\sd{0.9}
        & 5.1\sd{0.7}
        & \textbf{3.0\sd{0.0}}
        & 42.0\sd{0.2}
        & 8.0\sd{0.8} \\

        & & & RAFT
        & TR
        & 61.4\sd{1.1}
        & 17.7\sd{11.1}
        & 14.8\sd{7.4}
        & 2.6\sd{0.5}
        & 69.2\sd{1.2}
        & 38.0\sd{7.8} \\

        & & & SFT
        & TD
        & \textbf{82.2\sd{2.2}}
        & \textbf{18.4\sd{3.8}}
        & \textbf{26.0\sd{0.5}}
        & \textbf{3.0\sd{0.0}}
        & 96.1\sd{2.0}
        & 86.3\sd{6.1} \\

        & & & SFT${\scriptstyle\rightarrow}$RAFT
        & TD${\scriptstyle\rightarrow}$TR
        & 80.5\sd{2.2}
        & 13.5\sd{4.9}
        & 25.0\sd{0.2}
        & \textbf{3.0\sd{0.0}}
        & \textbf{96.9\sd{1.5}}
        & \textbf{90.7\sd{3.9}} \\

        \specialrule{1.2pt}{0.2ex}{0.2ex}

        \multirow{12}{*}{\rotatebox[origin=c]{90}{\textit{\tuglet{}}}}
        & \multirow{2}{*}{GPT-OSS}
        & Turn-level
        & ICL
        & $\dfew$
        & 31.3\sd{5.4}
        & \textbf{2.4\sd{0.5}}
        & 11.7\sd{6.7}
        & \textbf{3.0\sd{0.0}}
        & 20.5\sd{1.2}
        & 0.0\sd{0.0} \\

        \cmidrule(lr){4-11}

        & & \alps
        & ICL
        & $\dfew$
        & \textbf{58.9\sd{1.5}}$^\ast$
        & 0.0\sd{0.0}
        & \textbf{16.6\sd{6.9}}$^\ast$
        & \textbf{3.0\sd{0.0}}
        & \textbf{43.2\sd{1.0}}$^\ast$
        & \textbf{3.0\sd{0.8}} \\

        \cmidrule(l){3-11}

        & \multirow{9}{*}{Qwen3}
        & \multirow{4}{*}{Turn-level}
        & ICL
        & $\dfew$
        & 13.0\sd{0.7}
        & 0.0\sd{0.0}
        & 7.2\sd{0.4}
        & 0.0\sd{0.0}
        & 17.6\sd{1.5}
        & 0.0\sd{0.0} \\

        & & & RAFT
        & TR
        & 20.9\sd{4.1}
        & 0.0\sd{0.0}
        & 28.7\sd{3.1}
        & 1.0\sd{1.4}
        & 14.2\sd{2.3}
        & 0.0\sd{0.0} \\

        & & & SFT
        & TD
        & 21.1\sd{3.4}
        & 0.0\sd{0.0}
        & 3.8\sd{0.8}
        & 0.0\sd{0.0}
        & 32.7\sd{3.6}
        & 0.0\sd{0.0} \\

        & & & SFT${\scriptstyle\rightarrow}$RAFT
        & TD${\scriptstyle\rightarrow}$TR
        & 17.0\sd{1.3}
        & 0.0\sd{0.0}
        & 6.1\sd{1.3}
        & 0.7\sd{0.9}
        & 32.3\sd{4.1}
        & 0.0\sd{0.0} \\

        \cmidrule(lr){4-11}

        & & \multirow{4}{*}{\alps}
        & ICL
        & $\dfew$
        & 21.5\sd{1.7}$^\ast$
        & 1.0\sd{1.5}
        & 7.4\sd{0.0}
        & 1.0\sd{1.4}
        & 15.6\sd{0.7}
        & 0.0\sd{0.0} \\

        & & & RAFT
        & TR
        & 34.5\sd{6.2}$^\ast$
        & 1.0\sd{1.5}
        & 13.4\sd{7.3}
        & 1.0\sd{1.4}
        & 22.9\sd{0.6}
        & 0.0\sd{0.0} \\

        & & & SFT
        & TD
        & \textbf{57.5\sd{3.1}}$^\ast$
        & 2.4\sd{1.9}
        & \textbf{35.8\sd{7.6}}$^\ast$
        & \textbf{3.3\sd{0.5}}
        & 49.3\sd{5.3}$^\ast$
        & 12.0\sd{6.7} \\

        & & & SFT${\scriptstyle\rightarrow}$RAFT
        & TD${\scriptstyle\rightarrow}$TR
        & 57.1\sd{2.2}$^\ast$
        & \textbf{2.9\sd{2.4}}
        & 29.7\sd{7.7}
        & \textbf{3.3\sd{0.5}}
        & \textbf{51.1\sd{5.3}}
        & \textbf{12.7\sd{4.2}} \\

        \bottomrule[1.2pt]
    \end{tabular*}
    }
\captionsetup{skip=3pt}
\caption{\textbf{Exp2: Data-Scarce Setting.}
Simulation quality in \pharmasim{} and \tuglet{} when only three real
profile-labeled trajectories per profile are available and a teacher
LLM provides additional supervision. Results are reported as
percentages (mean$_{\pm\mathrm{std}}$ over three seeds). \emph{Mean}
averages performance across behavior profiles, and \emph{worst}
reports the lowest-performing profile. $\dfew$ denotes the three real
trajectories per profile, TD teacher-generated demonstrations, and TR
the teacher reward. Bold values are computed separately for GPT-OSS and
Qwen3 within each environment. Almost all column-wise best values are
achieved by an \alps{} configuration. $\ast$ indicates that \alps{} significantly outperforms its turn-level counterpart ($p<0.05$).}

    \label{tab:results-fewshot}
\end{table*}

\subsection{Data-Sufficient Setting (Exp1)}
\label{sec:results-data-sufficient}
First, we examined how architecture and adaptation affect profile alignment when real profile-labeled trajectories are available for direct supervision. Table~\ref{tab:results-limited} reports \density{}, \coverage{} and judge pass rate for turn-level generation and \alps{} under ICL, SFT, RAFT, and SFT$\rightarrow$RAFT. Per-profile results and statistical tests are reported in Appendix Tables~\ref{tab:per-profile-pharmasim},~\ref{tab:per-profile-tuglet}, and~\ref{tab:stats-rq12}. Across both environments, an \alps{} configuration achieves the strongest balance of \density{} and \coverage{} while maintaining realistic behavior.

Without fine-tuning, \alps{} significantly improves \density{} and judge pass rate over turn-level generation for GPT-OSS in both environments, with judge pass rates roughly tripling. It also significantly improves GPT-OSS \coverage{} on \pharmasim{}, although \coverage{} decreases on \tuglet{}. Unlike \alps{}, turn-level GPT-OSS fails completely on at least one difficult or long-horizon profile in each environment, driving its worst-profile scores to zero. None of the architecture comparisons is significantly positive for Qwen3.

For GPT-OSS, \alps{} limits profile drift by maintaining the assigned profile through a small number of segment-level decisions rather than re-establishing it at every action. The trajectory-prefix analysis in Appendix~\ref{app:drift} supports this explanation: GPT-OSS \alps{} becomes increasingly profile-aligned as trajectories unfold, while turn-level generation stagnates or declines toward chance. For Qwen3 on \tuglet{}, turn-level generation achieves higher \density{} and \coverage{} but a very low judge pass rate, indicating mechanistic agreement with aggregate features rather than realistic profile behavior. Removing the three long examples makes Qwen3 \alps{} significantly outperform turn-level generation in both \density{} and judge pass rate (Appendix \ref{app:icl}), suggesting that the added context overloads the 8B model and makes segment planning less reliable.

We next examined how fine-tuning adaptations improved the prompted simulators. Among the adaptation methods, every improvement that remains significant after correction comes from adding SFT to the prompted model. No comparison involving RAFT is significantly positive.
Compared with ICL, SFT improves all three metrics for \alps{} in both environments, with significant gains in \coverage{}, its weakest metric under ICL. Relative to turn-level SFT, \alps{} significantly improves \density{} and judge pass rate in both environments, as well as \coverage{} on \tuglet{}. Its \coverage{} gain on \pharmasim{} is not significant, but, together with the substantially higher \density{}, indicates a population that covers real students more faithfully. SFT therefore complements the hierarchical architecture by reducing mechanistic behavior and strengthening profile-aligned planning, making SFT-adapted \alps{} the only configuration that performs strongly across all three metrics. It also changes the comparison with the much larger prompted GPT-OSS: although GPT-OSS achieves higher per-trajectory \density{}, adapted Qwen3 covers more real students within each profile in both environments.

To isolate the source of the SFT gains, we separately ablate the planner and generator adapters (Appendix Table~\ref{tab:ablation}). The planner accounts for most profile alignment: replacing it with its prompted version returns \pharmasim{} to prompted density and coverage and sharply reduces all metrics on \tuglet{}. This larger drop likely reflects co-adaptation, as the adapted generator receives plans that differ from those produced by the adapted planner during training. The generator primarily supports \coverage{}: replacing it reduces \coverage{} in both environments and worst-profile \density{}, while mean \density{} changes little. Thus, the planner maintains profile alignment, while the generator broadens coverage of real students within each profile.

RAFT applied directly to base model provides the least advantage compared to ICL adaptation among the fine-tuning adaptation variants. On \pharmasim{}, the acceptance filter admits no candidates for two of the five profiles, preventing adapter training for those profiles. On \tuglet{}, RAFT is significantly worse than SFT on all three metrics. Applied after SFT, RAFT achieves the best nominal value in several columns but does not significantly outperform SFT in either environment. We therefore attribute the adaptation gains to SFT. Because the in-profile classifier is built from $\dlim$ also used for SFT, its reward provides little additional information. It also evaluates aggregate features without considering individual actions, allowing action-level errors to be reinforced. For the turn-level model, RAFT reverses the mean SFT gains. One third of its trajectories show mechanistic repetition, which \alps{} largely avoids, but the trajectory-level reward cannot detect.

\begin{leftbar}
\noindent
Overall, the data-sufficient results show that hierarchical generation is the primary driver of profile alignment, SFT consistently improves the simulation quality when real trajectories are available, and RAFT provides no reliable gains with or without SFT.
\end{leftbar}

\subsection{Data-Scarce Setting (Exp2)}
\label{sec:results-data-scarce}

In our second experiment, we assessed what remains achievable when only three real trajectories per profile are available. Table~\ref{tab:results-fewshot} reports \density{}, \coverage{} and judge pass rate for turn-level generation and \alps{} under all adaptation methods. Per-profile results and statistical tests are reported in Appendix Tables~\ref{tab:per-profile-pharmasim-fs},~\ref{tab:per-profile-tuglet-fs}, and~\ref{tab:stats-rq3}. Overall, teacher supervision substantially improves \alps{}, whereas no adaptation method reliably improves the turn-level student.

Under ICL, the three trajectories in $\dfew{}$ are included as in-context examples. For GPT-OSS, \alps{} significantly improves \density{} and judge pass rate over turn-level generation in both environments, as well as \coverage{} on \tuglet{}. For Qwen3, however, the longer prompt limits the effectiveness of ICL, and \alps{} provides no significant advantage. The zero-shot comparison in Appendix~\ref{app:icl} shows that the examples contribute little to \density{} but consistently increase \coverage{}.

Teacher-guided fine-tuning benefits only the \alps{} architecture. Every significant gain over ICL in either environment comes from applying teacher supervision to \alps{}, while no turn-level configuration improves significantly. Of the teacher’s two roles, demonstration provides most of the benefit: SFT on teacher demonstrations significantly improves \density{} and \coverage{} in both environments and judge pass rate on \pharmasim{}. On \tuglet{}, the judge pass rate increases nominally but does not differ significantly from the turn-level counterpart.

Teacher demonstrations convey profile information more reliably than few-shot prompting, which must fit the profile description and three examples into the 8B model's context. SFT instead transfers this information, including signature-consistent behavior, into the model weights. This transfer is particularly effective for \alps{}, whose planner can learn and maintain the trajectory-level structure in the demonstrations. Consequently, the adapted 8B \alps{} performs on par with its 120B teacher. No metric differs significantly on \tuglet{}; on \pharmasim{}, it matches \coverage{}, remains close to \density{}, and achieves a higher judge pass rate. In contrast, teacher demonstrations do not reliably improve the turn-level student.

The teacher reward succeeds where the data-driven reward in section~\ref{sec:results-data-sufficient} fails because it evaluates each candidate against the profile description, its behavioral signatures, and the generated actions, rather than only measuring proximity to the same real trajectories already used for SFT in an aggregated feature space. Teacher-scored RAFT can therefore identify profile-consistent behavior that the in-profile reward cannot distinguish. Demonstrations nevertheless remain the stronger form of guidance: SFT beats  RAFT on the base-model on judge pass rate in both environments because demonstrations add signature-consistent behavior that the prompted model does not sample reliably, whereas a reward can only reselect existing behavior. Applying RAFT after SFT adds no significant gain. Although SFT$\rightarrow$RAFT reaches the best nominal judge pass rate in several conditions, no metric improves significantly over SFT alone, and \coverage{} decreases in both environments. After SFT, the model already produces many trajectories that satisfy the teacher judge, so rejection sampling may retain only the most safely accepted candidates and narrow the demonstration-tuned distribution. The teacher is therefore best used to generate demonstrations.

\begin{leftbar}
\noindent
Overall, the data-scarce results show that teacher-generated demonstrations allow \alps{} to recover most of the profile fidelity lost from limited real data. \alps{} therefore remains the key ingredient for transferring profile-consistent behavior under scarce supervision.
\end{leftbar}
\section{Conclusion}
\label{sec:conclusion}

We introduced \alps{}, a hierarchical simulated student that uses segment-level planning to maintain behavior profiles across long trajectories. Across two environments and two LLMs, it improved profile alignment over turn-level generation and was the only prompted approach to cover every profile. ICL further improved \alps{} in both data settings, but recovering the \coverage{} of real students required SFT on sufficient real student trajectories.

Although these results were consistent across the evaluated settings, further evaluation is needed beyond the two environments and model families studied here. \alps{} also uses hand-designed segmentation rules and plan fields, and a single RAFT cycle. Future work should test automatic segmentation, ways to increase within-profile \coverage{} under scarce data, iterative adaptation under a reward model, transfer across environments, and downstream uses such as testing adaptive tutors before deployment.

Beyond education, profile-conditioned simulation could support pre-deployment testing of long-horizon interactive systems while reducing the need to share real interaction logs. However, limited within-profile \coverage{} under data scarcity and risks of misuse mean that simulated populations should complement studies with real users and be validated before deployment. Overall, \alps{} provides a practical basis for simulating diverse, profile-aligned behavior across long interactions in education, with potential to extended to other human-centered domains.
\bibliography{aaai2027}
\appendix
\newpage
\textbf{Appendix}

\section{Related Work}
\label{ap:related}

\alps{} draws on four lines of work that fall into two groups.
\emph{Simulated learners} and \emph{human behavior simulation with LLMs} define
the problem, generating student behavior that is both faithful and diverse, and informs our
evaluation by supplying the evidence on where such simulations break down. \emph{Hierarchical agents} and \emph{LLM-based rewards} offer the
means: a two-level architecture for long horizons, and a reward that scores an
entire trajectory against a target goal. These two groups' strengths have not
yet been combined. Simulators that hold a behavior profile do so over short,
language-centered tasks, while agents that stay coherent over long horizons
do so in pursuit of a task goal. \alps{} targets what neither achieves: holding
a behavior profile across hundreds of actions.

\subsection{Simulated Learners}
\label{ap:related-simlearners}
Simulated learners are computational models that generate learner-like actions, responses, or trajectories in educational environments, enabling controlled and scalable experimentation when real student data are scarce, costly, or ethically constrained \cite{DBLP:journals/aiedu/KaserA24,DBLP:conf/edm/MacLellanHPK16}.
Such models have long supported the design and evaluation of instructional policies by encoding assumptions about learner knowledge, misconceptions, and progress~\cite{DBLP:conf/edm/MacLellanHPK16}, and thereby adaptive instruction~\cite{McEneaney2022CurriculumMA, VanLehn1994,Mertz1997UsingAS}, error diagnosis~\cite{DBLP:conf/aied/WeitekampSM25}, and tutor development~\cite{Maclellan2014SimStudentAE}. They have also served as interactive counterparts for teachers and students: teachable agents in learning-by-teaching settings \cite{DBLP:journals/aai/BiswasLSV05,DBLP:conf/aied/MacLellanMK13}, and virtual peers that support collaborative learning and reflection \cite{DBLP:journals/aiedu/Pareto14,DBLP:conf/naacl/ZhangZYGZHJCLLHL25,Vizcano2005ASS}.

A 2024 review of roughly 200 studies since 2010~\cite{DBLP:journals/aiedu/KaserA24} reports growing adoption alongside two persistent challenges: narrow learner representations that capture only a slice of behavior, and high data requirements for realism and validation. Validation itself often remains limited, leaving the behavioral fidelity of simulated traces open.
Both challenges stem from trade-offs in rule-based, probabilistic, and heavily parameterized simulators such as SimStudent~\cite{Maclellan2014SimStudentAE} and the Apprentice Learner architecture~\cite{DBLP:conf/edm/MacLellanHPK16,DBLP:journals/aiedu/MacLellanK22}. These offer strong control and interpretability, but encode learners through a small set of predefined states or parameters, which limits the range of behavior profiles they express. Broadening that range demands substantial interaction data for parameter fitting, calibration of learner variability, and validation beyond surface outcomes \cite{DBLP:conf/edm/ZhangM21}.

Generative approaches based on LLMs can address both challenges by increasing expressivity and reducing dependence on hand-crafted structure (e.g., \cite{DBLP:conf/edm/NguyenTS24,DBLP:conf/naacl/ZhangZYGZHJCLLHL25,liu-etal-2024-personality}).
They now span a wide range of educational settings. These include personality-aware student simulation for conversational tutoring systems, which combines cognitive and non-cognitive traits such as the Big Five to generate plausible responses and trigger adaptive scaffolding \cite{liu-etal-2024-personality}; generative students that support item evaluation and MCQ response generation \cite{Lu2024}; and TeachTune, where teachers evaluate pedagogical agents against diverse simulated student profiles in a tutoring system \cite{DBLP:conf/chi/JinYPLWK25}. Others pair LLMs with GANs to synthesize tabular student data \cite{lak/synthetic-data}, simulate classroom conversation with iterative reflection \cite{DBLP:conf/chi/XuWPDHZ25}, align simulated students to item response theory for difficulty prediction \cite{DBLP:journals/corr/abs-2507-05129}, coordinate role-based agents for classroom dialogue in SimClass \cite{DBLP:conf/naacl/ZhangZYGZHJCLLHL25}, and model students in-context to synthesize visual-programming behavior from few observations \cite{DBLP:conf/edm/NguyenTS24}.

Despite these gains, LLM-based simulators are built for short, language-centered exchanges, and three difficulties surface once a trajectory runs to hundreds of actions.
 
First, a behavior profile is hard to hold in long-horizons. Prompted simulators are sensitive to wording and lose trajectory-level consistency across multi-step interactions, succeeding instead on isolated responses or code outputs \cite{DBLP:conf/edm/RadmehrSK24,DBLP:conf/naacl/ZhangZYGZHJCLLHL25}, and over long interactions they drift from the profile they were assigned \cite{DBLP:journals/corr/abs-2511-00222}. A related failure is a mismatch in competence: a strong model asked to act as a partial-knowledge learner produces errors real students do not make. Measured directly, simulators instructed to hold a misconception abandon it under tutoring feedback whether or not the feedback is relevant, and only supervised or reward-based training restores faithful behavior~\cite{Do2026Misconception}.
 
Second, the conditioning signal is rarely observable. Simulators are typically conditioned on generic learner attributes such as knowledge states or personality traits \cite{liu-etal-2024-personality,DBLP:conf/chi/JinYPLWK25}, which describe the student but not how the student acts in a particular environment. A generated trajectory therefore cannot be checked against the profile it was meant to express, and behaviors that arise from the available actions and constraints have no place in the specification. One response is to make the target explicit and testable. In Hyp-Mix, domain experts author a learner model as marginalized distributional hypotheses, each relating a learner characteristic to the probability of an action, such as a more persistent learner being less likely to abandon the task as measurements accumulate \cite{Mannekote2025HypMix}. The target is verifiable, but hand-written rather than derived from student data, and stated per action rather than over a trajectory.
 
Third, supervision is scarce in exactly the environments that motivate simulation, and prompting performs poorly under limited supervision with scarce data  \cite{DBLP:journals/corr/abs-2601-04025}. How to obtain faithful, diverse simulated learners when profile-labeled trajectories are few nonetheless remains open.
 
We address all three with \alps{}, which treats learner simulation as profile-conditioned generation of long trajectories. Drift follows from choosing one action at a time under a prompt that must carry the whole profile, so \alps{} instead plans a profile-aligned segment and then generates the actions that realize it: a few segment-level decisions carry the profile, and every action is chosen under the constraints of the current state. Because generic attributes leave nothing to verify, we condition on environment-specific behavior profiles derived from real trajectories, which give generation and evaluation the same observable target. And because supervision is often scarce, we adapt both components from whatever supervision is available and introduce an LLM-based reward that aligns profiles when profile-labeled trajectories are limited.

\subsection{Human Behavior Simulation with LLMs}
\label{ap:related-humansim}
Beyond education, a broader literature simulates human behavior with LLMs. It supplies the generation methods that educational simulators build on, and the evidence on where those simulations fail. Generative agents maintain a persistent character description and act on it across a simulated community, producing behavior that observers rate as human~\cite{DBLP:conf/uist/ParkOCMLB23}. Scaling this to a population, agents built from two-hour interviews with over 1,000 real individuals answer held-out General Social Survey items at $83\%$ of those individuals' own two-week test--retest consistency, against $74\%$ for agents given only demographics \cite{DBLP:journals/corr/abs-2411-10109}.
 
A second literature asks whether such simulations are accurate. On a benchmark of twenty human-behavior datasets, the best model scores $40.8$ out of $100$, and its answers sit closer to a uniform distribution than to the real human one~\cite{Hu2025SimBench}; instruction tuning helps on questions with a clear majority answer and hurts where people disagree. Simulated survey respondents likewise fail to reproduce the summary statistics of real social-science data unless trained to~\cite{Xie2026StatisticalRealism}. Both point to the same failure: LLMs reproduce the average respondent but not the variation between respondents.
 
Two responses to these findings shape our design. The first is to train the simulator rather than prompt it. Evaluating GPT-4o against a user model trained on real conversations, rather than against a prompted assistant playing the user, lowers its measured success from $74.6\%$ to $57.4\%$, so the prompted simulator was flattering the system under test~\cite{Naous2026Flipping}. In education, an evaluation of simulated students on 2,000 real tutoring dialogues across six dimensions finds that prompting reproduces student behavior poorly, that supervised fine-tuning improves most dimensions, and that preference optimization adds little beyond it, which the authors attribute to a sparse reward~\cite{DBLP:journals/corr/abs-2601-04025}. This matches our own result that SFT carries the gains. The second response is to measure coverage rather than assume it. Coverage across a simulated population can collapse even when each individual persona looks right: the model that matches personas most accurately one at a time produces the population that is most stereotyped and least varied~\cite{Xiao2026ChameleonsLimit}. fidelity and coverage therefore have to be measured separately.
 
Two studies show that a profile does not hold on its own over a long interaction, and both were run over dialogue rather than action trajectories. In the first, simulated learners were given profiles that differ in how easily attention is sustained. Asked about themselves, the models reported stable adherence, while observers rated their behavior as drifting over the conversation; scripting the conversation around explicit, task-relevant prompts removed up to $97\%$ of that drift~\cite{Gonnermann2026PersonaStability}. In the second, training against an automatic consistency reward across turns cut persona inconsistency by more than $55\%$ over several roles, including a student~\cite{DBLP:journals/corr/abs-2511-00222}. Drift therefore yields to imposed structure or to training rather than to a larger base model, and self-report does not reveal it.
 
These findings set our evaluation. We report \density{} and \coverage{} separately, each on average and for the worst profile, because a simulator can be faithful on individual trajectories yet cover only a narrow part of a profile. We score whole trajectories with a judge, rather than individual turns or the simulator's own report, because drift appears only over long interactions.

\subsection{Hierarchical Agents}
\label{ap:related-hierarchical}
Hierarchical agents split decision making into two levels. A high-level policy
picks a behavior to follow for many steps, and a low-level policy turns it into
single actions. The split appears in three literatures we draw on: classical RL,
where it was formalized; LLM agents, where it plans long tasks; and behavior
simulation, where it holds a character in place. We review each, then state what remains a gap.

Hierarchical decision-making was first formalized in reinforcement learning. The options framework models the two levels as a semi-Markov decision process in which each option defines where it can begin, the policy followed while active, and its termination condition~\cite{Sutton1999Options}. Later work learns termination conditions or trains low-level policies against goals provided by a higher-level policy; \citet{Pateria2021HRLSurvey} survey these approaches. Across these variants, hierarchy shortens the effective planning horizon, allowing outcomes to be attributed to a small number of high-level decisions rather than hundreds of individual actions.

With LLMs, the high-level policy is usually a planner that writes subgoals in
text or in a fixed format, and a separate executor carries them
out~\cite{Huang2024PlanningSurvey}. Systems in this family decompose a task
further only when the executor fails~\cite{Prasad2024ADaPT}, or train planner and
executor on trajectories annotated with plans after the
fact~\cite{Erdogan2025PlanAndAct}. In the design closest to ours, one LLM
predicts subgoals and a second generates the low-level actions, with the subgoal
model trained against environment
preferences~\cite{Zhao2024EPO,He2024WordsToActions}. Throughout, the plan breaks
a task goal into smaller goals, and the hierarchy is judged by whether the task
succeeds.

A newer line of work treats the long horizon itself as the problem. Requiring the
agent to state a subgoal before acting, then dropping completed subgoals from the
context, doubles success rates on long tasks~\cite{Hu2025HiAgent}, and folding
completed subgoal histories while training with subgoal-level rewards extends
this to RL~\cite{Diao2026HIPIF}. Independent evidence documents the need for such
structure. LLMs lose $39\%$ of their single-turn performance in multi-turn
settings, mostly by becoming unreliable rather than less
capable~\cite{Laban2025LostMultiTurn}, and they drift from their system prompt within
eight rounds as attention to it decays~\cite{Li2024InstructionInstability}.

Hierarchy has also been used for behavior simulation, though rarely as a learned
policy. Generative agents break a daily plan into hour-level and then
minute-level actions, all conditioned on a fixed character
description~\cite{DBLP:conf/uist/ParkOCMLB23}. Persona consistency is otherwise
pursued at the training objective, for example through multi-turn RL against
consistency rewards~\cite{DBLP:journals/corr/abs-2511-00222}. The closest system to ours is BEAGLE, a simulated student for Python programming tasks. A strategist turns each metacognitive segment, such as planning or debugging, into a goal and a directive; an executor implements it as code. To stop a capable model from quietly solving the task, the executor is prompted to implement rather than diagnose, and error traces are withheld while the student is enacting. A knowledge-tracing model separately sets what the student may get wrong~\cite{Wang2026BEAGLE}.

Two properties separate \alps{} from this work. The first is what the plan
contains. Prior planners decompose a task goal, so the hierarchy succeeds when
the task succeeds; the behavior plan in \alps{} specifies \emph{how} the student
should act over the next segment, as target counts and a stopping rule. This
matters because a profile-matching trajectory often contains slow or mistaken
behavior that a success-driven planner is built to avoid. The second is how the
two levels are obtained. \alps{} adapts both from real or teacher trajectories
rather than prompting them, conditions them on observable,
environment-specific behavior profiles rather than a symbolic knowledge state,
and trains them with reward-ranked fine-tuning against a whole-trajectory
judgment, because no per-action signal for profile fidelity exists. Otherwise the
correspondence is exact: the behavior plan $B_k$ acts as an option and the action
generator as the policy inside it, with segment boundaries fixed by the
environment. To our knowledge, no prior work adapts a two-level plan-then-act
policy to match a behavior profile over trajectories of hundreds of actions.
 
\subsection{LLM-based Rewards}
\label{ap:related-rewards}
Recent work uses LLMs as proxies for reward functions in RL, replacing manual reward engineering where objectives are sparse, delayed, or semantically rich. Rather than defining task-specific metrics, these methods have LLMs generate, evaluate, or refine rewards through natural-language prompting, code synthesis, or iterative feedback.
 
Early studies used LLMs as reward models in data-scarce settings: the user describes the desired behavior in natural language, optionally with examples, and the LLM scores trajectories for RL \cite{Kwon2023Reward}. The same idea underpins LLM-as-a-Judge for preference-based scoring \cite{Zheng2023Judge}. Later work had LLMs synthesize executable reward functions by evolutionary code generation \cite{Ma2023Eureka}, or produce dense rewards from natural-language instructions with iterative refinement \cite{Yu2023Text2Reward,Wang2024CARD}. In education, hybrid agents let an LLM constrain the action choices of a classical RL learner, yielding more human-like interaction in a simulated pharmacy task~\cite{PharmaSimText2025,DBLP:conf/edm/RadmehrSK24}.
 
The field has since moved from scalar scores to structured judgments, and from short outputs to long trajectories. Splitting a judgment into weighted rubric criteria improves both the reward and the RL that follows~\cite{gunjal2025rubrics}. Two further findings constrain how such a reward can be used. First, a reward that is misspecified in its high-scoring range does the most harm~\cite{Zhang2026ChasingTail}, and reward-ranked fine-tuning trains on precisely that range. Second, on a benchmark of whole-trajectory preferences, reward models and LLM judges alike lose accuracy sharply as trajectories grow longer~\cite{Wang2026PlanRewardBench}. We therefore keep the teacher reward a checklist with a written reason per item, calibrate its tolerance on $\dfew$, give it a precomputed feature table so it never counts actions itself, and evaluate with a judge whose prompts and inputs do not overlap with the reward.
 
These methods report strong results in robotics, control, games, and negotiation, but target non-educational settings with simple objectives and short horizons. Learner simulation differs: the reward must capture \emph{profile-level} pedagogical behavior and enforce consistency across extended interaction, often where profile-labeled trajectories are limited. We therefore use an LLM-based reward as a trajectory-level signal for RL fine-tuning, scoring each generated interaction sequence against a textual behavior profile and using those scores to align simulated learners toward profile-consistent trajectories.

\section{Environments, Data, and Behavior Profiles}
\label{app:environments}
\label{app:segmentation}

\begin{figure*}[ht]
\centering
\includegraphics[width=0.95\linewidth,trim=0cm 8.3cm 0cm 0cm,clip]{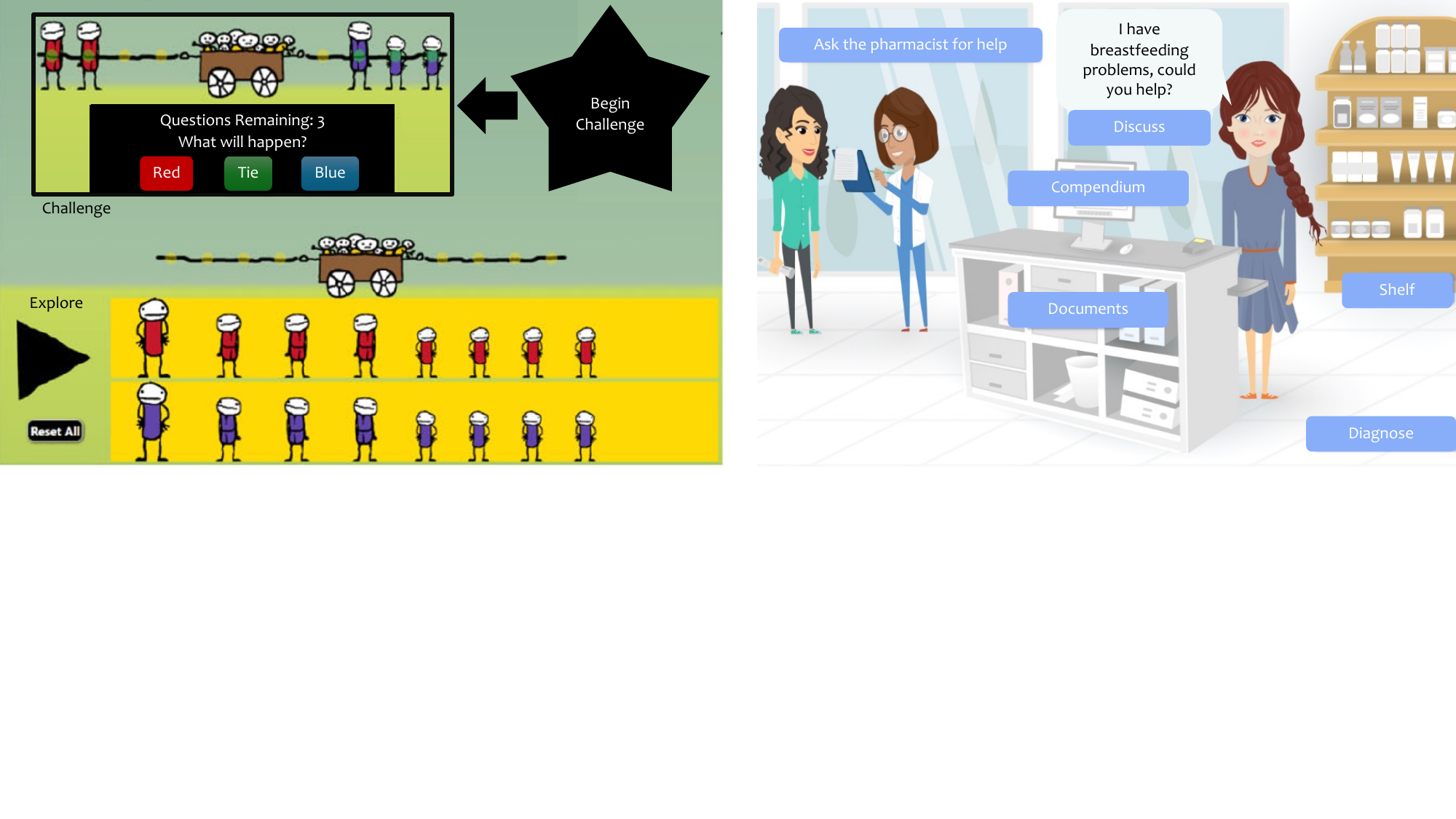}
\caption{User interface of the two learning environments. Left
(\tuglet): an inquiry-based game where learners alternate between
Explore for testing setups and Challenge for predicting outcomes.
Right (\pharmasim): a simulated pharmacy scenario where learners
gather information by discussing with characters and consulting the
compendium, documents, shelf, and pharmacist.}
\label{fig:envs}
\end{figure*}

We evaluate on two open-ended learning environments that span distinct domains, interaction structures, and trajectory lengths. To the best of our knowledge, no publicly available dataset offers comparable breadth and scale of long-horizon student interactions across such diverse environments. \tuglet{} is an inquiry-based game in which students experiment with tug-of-war contests to discover that motion occurs toward the side with the greater net force~\cite{DBLP:journals/aiedu/KaserS20}, whereas \pharmasim{} is a simulated pharmacy environment for practicing diagnostic conversations~\cite{PharmaSimText2025}. Together, they cover two distinct learning topics and interactions ranging from tens to nearly 200 actions. Agent-compatible implementations of both environments will be included in our code release; Appendix~\ref{app:availability} describes the released code, data representations, and generated populations.

For each environment we describe, in order, the learning task, the agent-facing representation used as fixed prompt context, the segmentation rules and behavior plan fields used by the hierarchical generator (Section~\ref{sec:alps-structure}), the dataset and its split, the behavioral features underlying the profile space (Section~\ref{sec:behavior-profiles}), and the resulting behavior profiles. Table~\ref{tab:env-summary} summarizes the environment-level quantities used by the experimental protocol.

\subsection{\tuglet{}}
\label{app:env-tuglet}
As shown in Figure~\ref{fig:envs} (Left), \tuglet{} is an inquiry-based game in which students discover the physics rules governing tug-of-war contests~\cite{DBLP:journals/aiedu/KaserS20}. Characters have three hidden force values: small ($f=1$), medium ($f=2$), and large ($f=3$). The side with the greater total force wins, and equal forces result in a tie.

Students interact through two modes. In \emph{Explore}, they construct and simulate teams of up to four characters per side. The game begins with a simplified Explore mode in which only one small and one large character are available for each side. In \emph{Challenge}, students predict the outcomes of eight configurations of increasing difficulty. An incorrect answer returns the student to Explore, but the student may immediately re-enter Challenge. The game ends after eight consecutive correct answers or 15 minutes. The choice between exploration and repeated challenge attempts results in different inquiry strategies and long interaction trajectories.

For agent interaction, we represent the environment using three action types: simulating a configuration, entering Challenge, and answering a challenge question. Because a simulated student has no wall-clock time, the 15-minute limit of the original game is replaced by an action budget of 200 actions, slightly above the longest real trajectory (185 actions). A trajectory that reaches this budget is completed deterministically by answering the remaining Challenge questions correctly, so that every generated trajectory terminates in the same environment state as a completed real session. The same completion procedure is applied to \alps{} and to the turn-level student.

\paragraph{Agent-facing representation.} We expose \tuglet{} as a turn-based environment in which the agent observes the current mode, game state, interaction history, and target profile and selects one valid action. The following representation is included in the environment-specific prompt context.

\begin{tcolorbox}[
    enhanced,
    breakable,
    colback=gray!3,
    colframe=black!50,
    boxrule=0.5pt,
    arc=1mm,
    title={\tuglet{} Environment Representation}
]
\small
\textbf{Modes.} The student is either in Explore or Challenge mode. An incorrect Challenge answer returns the student to Explore.

\textbf{Environment rules.} The environment uses the following relations: identical teams tie; identical characters on opposite sides cancel; more characters of the same type win; $M>S$; $L>S$; $L>M$; $M=2S$; $L>2S$; $L=3S$; $L=M+S$; $2M>L$; and $S+L=2M$. These rules determine outcomes but are hidden from real students, who must infer them through exploration.

\textbf{Actions.}
\begin{itemize}
    \item \texttt{explore(L,R)} constructs and simulates two teams; valid only in Explore.
    \item \texttt{start\_challenge()} enters Challenge; valid only in Explore.
    \item \texttt{submit\_answer(L/R/T)} predicts whether the left side wins, the right side wins, or the teams tie; valid only in Challenge.
\end{itemize}

\textbf{Exploration phases.} Before the first Challenge attempt, each side may contain only one small and one large character, with no repetitions. After the first attempt, small, medium, and large characters may be repeated, with at most four characters per side.

\textbf{Exploration quality.} Strong explorations use small, non-mirrored configurations that isolate a force relationship. Weak explorations use mirrored or cluttered configurations. The agent is instructed to match its exploration quality to the target profile.

\textbf{Termination.} A session is completed after eight consecutive correct Challenge answers.
\end{tcolorbox}

\paragraph{Segmentation and planning.} For hierarchical generation, a segment consists of one Explore phase followed by one Challenge phase. It ends when the student answers a Challenge question incorrectly or completes the game. Segment boundaries are therefore determined by the action sequence alone, which makes the segmentation map deterministic and identical for real and generated trajectories. The segment plan specifies the numbers of explorations, strong explorations, and Challenge questions; the rules explored and explored strongly; and how the segment ends. A deterministic normalizer corrects violations of the planning schema, for example a planned number of strong explorations that exceeds the planned number of explorations, and malformed outputs are retried. Real trajectories in $\dlim$ contain a median of 5 segments and at most 26, so a typical session is planned through a handful of planner calls and even the longest sessions, which span up to 185 actions, require fewer than 30.

Before planning the next segment, the planner receives compact summaries of the previously
generated segments. These summaries retain the behavioral information needed to maintain the
assigned profile without including the full action history:

\begin{lstlisting}[basicstyle=\scriptsize\ttfamily,breaklines=true,numbers=none]
Segment 1 (simplified mode):
{
  "n_exp": 8,
  "n_strong": 2,
  "explored_rules": [
    "cancellation",
    "identical_teams",
    "large_wins_small",
    "non-empty_wins_empty"
  ],
  "explored_strong_rules": [
    "large_wins_small",
    "non-empty_wins_empty"
  ],
  "n_chal": 5,
  "end": "wrong"
}
Segment 2 (full mode):
{
  "n_exp": 2,
  "n_strong": 0,
  "explored_rules": [
    "medium_wins_small",
    "two_small_equals_medium"
  ],
  "explored_strong_rules": [],
  "n_chal": 4,
  "end": "wrong"
}
\end{lstlisting}
\paragraph{Dataset.} The \tuglet{} dataset comprises trajectories from 249 students. The original data pool contains 288 recorded sessions from two cohorts of 159 and 129 students. Following~\cite{DBLP:journals/aiedu/KaserS20}, we excluded 27 students who did not complete the game. We split the remaining 261 students into development and held-out test sets using a 60/40 split stratified by the profile labels from an initial clustering (seed 52). We then re-ran clustering on the development set to ensure the held-out test set is not involved in any stage of development. Following that clustering we then excluded 12 development-set students with negative clustering silhouettes, because their profile assignment is ambiguous and they would provide unreliable supervision, and we then labeled the test set using the weighted KNN classifier described in Section~\ref{sec:alps-adaptation}. This yields a development set $\dlim$ of 140 students and a held-out test set $\dtest$ of 109 students. Real trajectories have a median of 41 actions, a 90th percentile of 124, and a maximum of 185. For the data-scarce setting, $\dfew$ contains three trajectories per profile sampled uniformly from $\dlim$ (18 trajectories in total), and the same $\dfew$ is used for ICL, teacher demonstrations, and the teacher reward.
\begin{table}[t]
\centering
\footnotesize
\begin{tabular}{@{}lcc@{}}
\toprule
& \tuglet{} & \pharmasim{} \\
\midrule
Action types & 3 & 7 \\
Behavior profiles & 6 & 5 \\
Students ($\dlim$ / $\dtest$) & 140 / 109 & 140 / 94 \\
Trajectories in $\dfew$ & 18 & 15 \\
Median trajectory length & 41 & 31 \\
90th percentile length & 124 & 39 \\
Maximum trajectory length & 185 & 61 \\
Action budget & 200 & 60 \\
Segments per trajectory (median / max) & 5 / 26 & 2 / 9 \\
Behavioral features & 24 & 14 \\
\bottomrule
\end{tabular}
\caption{Environment-level quantities. The action budget is the generation limit referenced in Section~\ref{sec:protocol}; trajectories that reach it are completed with the deterministic procedure described below. Segment counts are measured on the real trajectories in $\dlim$.}
\label{tab:env-summary}
\end{table}

\paragraph{Behavioral features.} Following~\cite{DBLP:journals/aiedu/KaserS20}, each trajectory is reduced to a 24-dimensional vector with three feature groups of eight dimensions each, indexed by game level $L=1..8$: challenge questions answered, explores performed, and strong explores performed. Dimension $L$ of a group holds the cumulative count at the moment level $L$ was first reached; levels never reached carry a sentinel that is masked out of every distance computation. An exploration is \emph{strong} if its configuration is small enough to interpret and its outcome isolates exactly one rule that the student has not previously explored. Distances are computed separately within each group and summed, as in Section~\ref{sec:behavior-profiles}, and the same feature space is used for clustering, the in-profile reward (Appendix~\ref{app:rewards}), and the \density{} and \coverage{} metrics.

\begin{table*}[t]
\centering
\small
\begin{tabular}{@{}L{2.6cm}L{2.2cm}L{2.2cm}L{2.8cm}L{2.0cm}L{3.9cm}@{}}
\toprule
\textbf{Profile} & \textbf{Exploration volume} & \textbf{Strong-explore ratio} & \textbf{Phase concentration} & \textbf{Session length} & \textbf{Distinguishing behavior} \\
\midrule
P1: Thorough Learner & substantial & highest & middle (L3--L5) & moderate & most middle-phase exploration \\
\addlinespace
P2: Efficient Learner & modest & second highest & opening (L1), then tapering & short & brief but high-quality exploration \\
\addlinespace
P3: Challenge-First & lowest & near zero & almost entirely opening (L1) & short & advances mainly through Challenge attempts \\
\addlinespace
P4: Early Burst & large initial burst & lowest & opening (L1) & moderate & largest opening-phase burst \\
\addlinespace
P5: Persistent Grinder & modest & low & mostly early and middle, with sparse late exploration & long & continues exploring during the final grind \\
\addlinespace
P6: Stubborn Grinder & lowest per segment & near zero & sparse throughout & long & highest Challenge-question volume \\
\bottomrule
\end{tabular}
\caption{Qualitative summary of the six \tuglet{} behavior profiles. Properties are stated relative to the other profiles rather than through fixed numeric thresholds. The complete model-facing descriptions and hard and soft signatures will be provided in our code release.}
\label{tab:tuglet-profile-summary}
\end{table*}

\paragraph{Profiles.} We adopt the six behavior profiles established in prior work~\cite{DBLP:journals/aiedu/KaserS20} and reuse their published descriptions as $d_p$; profile labels for individual students are obtained by re-running the clustering procedure of Section~\ref{sec:behavior-profiles} on the feature space above, which recovers the same six profiles. They differ primarily in exploration volume, the proportion of strong rule-isolating explorations, the phase in which exploration occurs, and session length. As summarized in Table~\ref{tab:tuglet-profile-summary}, P1 explores extensively and systematically, particularly during the middle levels. P2 performs a brief period of high-quality exploration and completes the game efficiently. P3 performs almost no exploration and advances primarily through Challenge attempts. P4 concentrates a large, low-quality exploration burst in the initial simplified phase. P5 enters a prolonged final-phase grind but continues to explore occasionally, whereas P6 grinds primarily through trial and error with minimal exploration. Table~\ref{tab:profile-counts} reports the corresponding development- and test-set sizes.

Each model-facing profile representation contains a natural-language description $d_p$, a set of behavioral signatures $\mathcal{S}_p$, and aspects of behavior left unconstrained. Hard signatures state behaviors that hold for every student in the profile, whereas soft signatures state tendencies that are typical of the profile but not universal. Signatures are derived with the procedure of Section~\ref{sec:behavior-profiles} and reviewed by a domain expert. The data-scarce representations state each profile's behavior through ordinal contrasts against the other profiles and definitional signatures, and contain no fitted numeric statistics; the only numbers they carry are fixed environment properties. The data-sufficient representations additionally include quantitative summaries estimated from $\dlim$: whole-session ranges and means for segment count, exploration count, strong-exploration count, and strong-exploration ratio; a per-phase breakdown of segments, explores, strong explores, and Challenge answers for each of L1, L1$\rightarrow$L3, L3$\rightarrow$L5, and L5$\rightarrow$L8; explicit numeric contrasts against neighboring profiles; and a calibration key mapping the qualitative magnitude terms onto numeric bands, for example explores per phase as negligible (0), minimal (1--2), moderate (3--6), extensive (7--16), and heavy (17--40). Our code release will provide the exact profile representations supplied to the models.

\subsection{\pharmasim{}}
\label{app:env-pharmasim}
\pharmasim{} simulates a pharmacy consultation in which a mother visits the pharmacy with her baby and reports a breastfeeding problem. The student questions the mother and the baby and consults the available resources to determine the cause of the problem. The student may question either character, consult three research tools (a medicine compendium, shelf products, and health documents), and request hints from a pharmacist. We adapt the existing PharmaSimText implementation for agent interaction~\cite{PharmaSimText2025}. The environment defines seven action types: questioning a character about a topic, consulting one of the three research tools, requesting a pharmacist hint, ending the current segment, and ending the session. Research relevance is defined against a fixed set of scenario-relevant targets comprising one medicine, five documents, and five products. Generation is limited to an action budget of 60 actions; a trajectory that reaches the budget is terminated deterministically with a session-ending action, using the same procedure for \alps{} and the turn-level student.

\paragraph{Agent-facing representation.} The following description summarizes the \pharmasim{} environment information provided in the prompt context. The exact component-specific prompts will be included in our code release.

\begin{tcolorbox}[
    enhanced,
    breakable,
    colback=gray!3,
    colframe=black!50,
    boxrule=0.5pt,
    arc=1mm,
    title={\pharmasim{} Environment Representation}
]
\small
\textbf{Task.} A mother visits a pharmacy with a breastfeeding problem. The student interacts with the mother and the baby and consults the available resources to determine the cause of the problem.

\textbf{Characters.} The student may question the mother or the baby about an available topic. Questions are represented as character--topic pairs.

\textbf{Research tools.} The student may consult a medicine compendium, shelf products, or health documents. Each tool contains both scenario-relevant and irrelevant resources. The scenario-relevant targets comprise one medicine, five shelf products, and five health documents.

\textbf{Pharmacist hints.} The student may ask the pharmacist for help. The environment provides two distinct hints, although the student may request hints more than twice.

\textbf{Actions.} At each step, the student selects one of seven action types: question a character about a topic, consult one of the three research tools, request a pharmacist hint, end the current segment, or end the session.

\end{tcolorbox}

\paragraph{Segmentation and planning.} For hierarchical generation, a segment consists of a block of discussion actions targeting the patient followed by an optional tail containing discussion actions not targeting the patient, research or hint actions. It ends with an explicit segment- or session-ending action, so segment boundaries are again determined by the action sequence alone and the segmentation map is deterministic and identical for real and generated trajectories. The segment plan specifies the topics discussed about each main character mother or her baby, the number of side-character questions, per-tool research and relevance targets, pharmacist-hint usage, the ordering of the  tail, and whether the segment is final. A deterministic normalizer corrects violations of the planning schema, for example a planned number of relevant document actions that exceeds the planned number of document actions, and malformed outputs are retried. Real trajectories in $\dlim$ contain a median of 2 segments and at most 9.

As in \tuglet{}, the planner receives compact summaries of the previously generated segments,
which retain the behavioral information needed to maintain the assigned profile without
including the full action history:

\begin{lstlisting}[basicstyle=\scriptsize\ttfamily,breaklines=true,numbers=none]
Segment 1:
{
  "mother_topics": ["symptoms"],
  "baby_topics": ["symptoms"],
  "n_other": 0,
  "n_doc": 0,
  "n_doc_relevant": 0,
  "n_shelf": 0,
  "n_shelf_relevant": 0,
  "n_compendium": 0,
  "n_compendium_relevant": 0,
  "n_hint": 1,
  "tail_lead_category": "baby",
  "tail_composition": "blocked",
  "end": "segment"
}
\end{lstlisting}

\paragraph{Dataset.} The dataset contains 234 recorded sessions. We divide these into a development set $\dlim$ of 140 students and a held-out test set $\dtest$ of 94 students using a profile-stratified split with random seed 42, which leaves at least five test students in every profile. Profiles are derived exclusively from $\dlim$: we cluster the development trajectories with the procedure of Section~\ref{sec:behavior-profiles}, and the BIC averaged over 200 seeds selects five clusters. The stratification labels for the initial split are taken from a preliminary clustering of all 234 trajectories, and the trajectories in $\dtest$ are labeled out of sample using the weighted KNN classifier trained on $\dlim$, so no test trajectory contributes to any profile definition, demonstration, or reward. Real trajectories have a median of 31 actions and a maximum of 61. For the data-scarce setting, $\dfew$ contains three trajectories per profile sampled uniformly from $\dlim$ (15 trajectories in total).

\paragraph{Behavioral features.} We represent each trajectory using 14 features organized into five groups (Table~\ref{tab:pharmasim-features}). Fractions are set to zero when their denominator is zero. Dissimilarities are computed separately within each group and summed, preventing larger groups from dominating the representation; the same feature space is used for clustering, for the in-profile reward and the KNN profile classifier (Appendix~\ref{app:rewards}), and for the \density{} and \coverage{} metrics.

\begin{table}[htbp]
\centering
\footnotesize
\begin{tabular}{@{}lp{0.58\columnwidth}@{}}
\toprule
Group & Features \\
\midrule
Discussion effort & mother share, baby share, and number of discussion actions \\
Research volume & numbers of document, shelf-product, and compendium actions \\
Research relevance & fraction of relevant actions for each research tool (3) \\
Hint usage & number of pharmacist hints \\
Topic variety & topic coverage and repetition rate for each character (4) \\
\bottomrule
\end{tabular}
\caption{The 14 \pharmasim{} behavioral features organized into five groups.}
\label{tab:pharmasim-features}
\end{table}

\paragraph{Profiles.} We derive five behavior profiles using the procedure in Section~\ref{sec:behavior-profiles}. They differ in how discussion is distributed between the mother and baby, the breadth and repetition of topic coverage, pharmacist-hint usage, research behavior, and session structure. Table~\ref{tab:pharmasim-profile-summary} summarizes these differences.

\begin{table*}[t]
\centering
\small
\begin{tabular}{@{}L{3.2cm}L{3.0cm}L{2.7cm}L{3.5cm}L{3.8cm}@{}}
\toprule
\textbf{Profile} & \textbf{Discussion} & \textbf{Hints} & \textbf{Coverage and repetition} & \textbf{Distinguishing behavior} \\
\midrule
P1: Hint-Heavy Deep Engagement & both characters, somewhat mother-weighted & highest; distributed across segments & broad coverage with limited repetition & substantial engagement over multiple varied segments \\
\addlinespace
P2: Mother-Only & mother only & not defining & narrow, short discussion & never questions the baby and remains within one segment \\
\addlinespace
P3: Hint-Seeking Wide Coverage & both characters, mother first & always at least one & broad coverage with almost no repetition & routinely checks understanding with the pharmacist \\
\addlinespace
P4: Independent Balanced Learner & balanced across both characters & none & broadest coverage with the least repetition & works independently and targets relevant documents \\
\addlinespace
P5: Baby-Only & baby only & not defining & very limited discussion & never questions the mother and ends after one short segment \\
\bottomrule
\end{tabular}
\caption{Qualitative summary of the five \pharmasim{} behavior profiles. Our code release will provide the complete model-facing descriptions and hard and soft behavioral signatures.}
\label{tab:pharmasim-profile-summary}
\end{table*}

Each model-facing profile representation contains a natural-language description $d_p$, a set of behavioral signatures $\mathcal{S}_p$, and aspects of behavior left unconstrained. Hard signatures state behaviors exhibited by every student in the profile and use absolute constraints such as ``never'' or ``always,'' whereas soft signatures describe tendencies that are typical of the profile but not universal. The data-scarce representations contain no fitted numeric statistics, while the data-sufficient representations additionally contain quantitative summary statistics estimated from $\dlim$: for each of the behavioral features, namely the discussion counts and mother and baby shares, the document, shelf, and compendium counts and their case-relevance rates, the hint count, and the mother and baby topic-coverage scores, a coarse band (none / low / typical / high) together with a numeric range, central value, or proportion, plus the typical segment count and tail composition. The two topic-repeat-rate features are the only behavioral features not given numerically. Our code release will provide the exact representations supplied to the models.

\begin{table}[t]
\centering
\begin{tabular}{lcc@{\hspace{2em}}cc}
\toprule
& \multicolumn{2}{c}{\tuglet{}} & \multicolumn{2}{c}{\pharmasim{}} \\
\cmidrule(r{2em}){2-3}\cmidrule{4-5}
Profile & $\dlim$ & $\dtest$ & $\dlim$ & $\dtest$ \\
\midrule
P1 & 20 & 19 & 8  & 5  \\
P2 & 44 & 39 & 17 & 11 \\
P3 & 40 & 25 & 42 & 29 \\
P4 & 8  & 5  & 65 & 43 \\
P5 & 12 & 10 & 8  & 6  \\
P6 & 16 & 11 & --- & --- \\
\midrule
Total & 140 & 109 & 140 & 94 \\
\bottomrule
\end{tabular}
\caption{Students per behavior profile in the development set $\dlim$ and held-out test set $\dtest$. Each student contributes one trajectory. Every profile contains at least five test students. Following Section~\ref{sec:evaluation}, \density{} and \coverage{} use $N{=}1$ neighbor for the profiles with fewer than ten test students (\tuglet{} P4, \pharmasim{} P1 and P5) and $N{=}3$ otherwise.}
\label{tab:profile-counts}
\end{table}

\section{Implementation Details}
\label{app:experimental-details}
This appendix reports the details needed to reproduce our experiments. We first describe
the two rewards used to filter candidate trajectories during adaptation
(Appendix~\ref{app:rewards}): an in-profile classifier reward for the data-sufficient
setting and the teacher reward \tr{} for the data-scarce setting. We then describe how
teacher demonstrations were generated and converted into training examples for SFT and
RAFT (Appendix~\ref{app:teacher-data}), and list the fine-tuning hyperparameters and the
seeds used for the three replicas (Appendix~\ref{app:hyperparameters}).

\subsection{Adaptation Rewards}
\label{app:rewards}

We used two rewards to filter candidate trajectories during adaptation. Both were constructed from $\dlim$ or $\dfew$ and used no information from $\dtest$. The environment-specific behavioral feature spaces are defined in Appendix~\ref{app:environments}.

\paragraph{In-profile reward.} In the data-sufficient setting, we fit a weighted $k$-nearest-neighbor classifier using the environment-specific behavioral features. We selected $k$ by leave-one-out cross-validation on balanced accuracy and weighted votes by inverse class frequency and inverse distance. A candidate was accepted when its target-profile vote share exceeded $0.5$ and it passed a within-profile outlier gate. The gate required its $k$th-nearest within-profile distance not to exceed the 95th percentile of the corresponding distances among real students of that profile.

\paragraph{Teacher reward (\tr).} In the data-scarce setting, GPT-OSS-120B evaluated each candidate using the qualitative target-profile representation, three same-profile references, contrast references from the two closest profiles, and a feature table comparing the candidate with these references (Prompt~\ref{prompt:teacher-assessment}). 
Same-profile references were drawn from a fixed pool of three real students per profile sampled from $\dlim$ with seed 42. Contrast profiles were the two profiles whose centroids were closest to the target profile in the standardized behavioral feature space.
It answered a fixed checklist of binary realism, profile-match, and signature questions and provided a one-sentence justification for each answer. To accommodate occasional item-level errors, we used a soft acceptance threshold rather than requiring every check to pass. In \tuglet{}, acceptance required at least three of four realism checks and three of four profile-match checks to pass. In \pharmasim{}, it required at least three of four realism checks and four of six profile-match checks. In both environments, acceptance additionally required no hard-signature violation and at most one soft-signature violation. We calibrated these thresholds on $\dfew$ and then held them fixed across profiles. The evaluation judge in Appendix~\ref{app:judge} was constructed separately and was not used as an adaptation reward.

\subsection{Teacher Demonstrations and Training Data}
\label{app:teacher-data}

The teacher generated demonstrations using the same  architecture as the simulated student being adapted. Generation used temperature $0.7$, top-$p=0.8$, and one sample per call. The teacher was conditioned on the qualitative profile representation and the three trajectories in $\dfew$. Candidate demonstrations accepted by \tr{} formed the teacher-demonstration set $\td$.

Accepted trajectories were converted into training examples using the segmentation rules in Appendix~\ref{app:segmentation}. Each segment provided one planner example whose target plan was extracted from the realized actions, and each action provided one action-generator example conditioned on that plan. Action-generator examples were retained only when the realized segment sufficiently matched its plan. We required at least $0.8$ field agreement. Later segments were up-sampled to reduce their underrepresentation. In-context examples were removed from the prompts when constructing fine-tuning data.

For SFT, we split the available training examples 80/20 into training and validation sets and selected the checkpoint with the lowest validation loss.

For RAFT, we performed one generate--filter--train round. Candidate generation was capped at 1,000 trajectories per profile in the data-sufficient setting and 500 per profile due to cost constraints in prompting the teacher LLM for rewarding in the data-scarce setting, with at most 100 accepted trajectories retained per profile. When no candidate was accepted for a profile, we retained the corresponding base model and marked that profile in the result tables.

\begin{table}[htbp]
\centering
\footnotesize
\setlength{\tabcolsep}{3pt}
\begin{tabular}{@{}lp{3.15cm}ccc@{}}
\toprule
Environment & Quantity & R1 & R2 & R3 \\
\midrule
\multirow{4}{*}{\tuglet{}}
  & Training seed & 4096 & 4097 & 4098 \\
  & Data-split seed            & 42   & 43   & 44   \\
  & Generation seed  & 42   & 142  & 242  \\
\midrule
\multirow{4}{*}{\pharmasim{}}
  & Training seed & 4096 & 1234 & 5678 \\
  & Data-split seed            & 42   & 43   & 44   \\
  & Generation seed  & 42   & 142  & 242  \\
\bottomrule
\end{tabular}
\caption{Random seeds for the three replicas.}
\label{tab:seeds}
\end{table}

\subsection{Fine-Tuning Hyperparameters}
\label{app:hyperparameters}

All adapters were trained on Qwen3-8B. Loss was computed on response tokens only. We trained one LoRA adapter per profile, environment, architecture component, and adaptation condition. In SFT$\rightarrow$RAFT, RAFT continued from the corresponding SFT adapter.
\begin{table}[htbp]
\centering
\footnotesize
\setlength{\tabcolsep}{3pt}
\begin{tabular}{p{3.1cm}p{1.6cm}p{1.6cm}}
\toprule
Setting & \tuglet{} & \pharmasim{} \\
\midrule
Base model & \multicolumn{2}{l}{Qwen3-8B} \\
LoRA $r$ / $\alpha$ / dropout & \multicolumn{2}{l}{32 / 64 / 0} \\
LoRA variant & \multicolumn{2}{l}{rsLoRA} \\
Target modules & \multicolumn{2}{l}{all attn.\ \& MLP proj.} \\
Optimizer & \multicolumn{2}{l}{AdamW} \\
Learning rate & \multicolumn{2}{l}{$2\times10^{-5}$} \\
Schedule & \multicolumn{2}{l}{cosine, 25 warm-up} \\
Weight decay & \multicolumn{2}{l}{$0.001$} \\
Batch/device & 16 & 4 (eval: 8) \\
Epochs (max) & 8 & 8 \\
Early-stop patience & 4 & 20 \\
Planner ctx.\ window & 8{,}192 & 4{,}096 \\
Action-gen.\ window & \multicolumn{2}{l}{4{,}096} \\
Segment weighting & \multicolumn{2}{l}{inverse frequency} \\
Weight power / max & \multicolumn{2}{l}{$0.5$ / $4.0$} \\
Checkpoint sel. & \multicolumn{2}{l}{lowest val.\ loss} \\
\bottomrule
\end{tabular}
\caption{Fine-tuning hyperparameters. The \tuglet{} planner used gradient accumulation to preserve an effective batch size of 16.}
\label{tab:hparams}
\end{table}

We ran three replicas of each condition, varying the training seed, training--validation split, and generation seed. All conditions used the same decoding parameters during population generation.
 Table~\ref{tab:seeds} reports the training, data-split, generation seeds used for each replica. 
 
\subsection{Judge Construction}
\label{app:judge}
The evaluation judge was constructed independently of both adaptation rewards, so that no
part of the reward signal was reused to score trajectories. It received abstracted action
trajectories rather than the original interface text. Its inputs were the profile
descriptions and signatures, the candidate trajectory, and a table of quantities measured
from that trajectory. All judge calls used \texttt{openai/gpt-oss-120b} at reasoning effort
\emph{high} and temperature $0.6$, with one vote per trajectory and structured outputs
enforced per stage.

\paragraph{Blind profile identification ($G$).} The judge first received the candidate
trajectory and the descriptions of all profiles, without being told which profile had been
assigned. It selected the profile that best matched the trajectory. This stage passed when
the selected profile matched the assigned one.

\paragraph{Signature checks ($H$).} The assigned profile was then revealed, and the judge
evaluated each of its signatures separately. Each judgment consisted of a binary answer and
a short justification citing evidence from the trajectory.

\paragraph{Mechanical checks ($R$).} Deterministic checks identified generation artifacts
absent from real trajectories. Thresholds were calibrated on the real students of each
environment and then fixed across all evaluated conditions. Each threshold was set at the
corresponding real-student extreme, so that any violation falls outside the observed range
of human behavior. The environment- and profile-specific signatures and checks will be given in
our code release.

\paragraph{Verdict.} A trajectory passed the judge only when its assigned profile was
identified correctly, all signature checks passed, and no mechanical check fired; that is,
$V = G \cdot H \cdot R$.

\section{Prompt Templates}
\label{app:prompt-templates}
The segment planner and action generator used
Prompt~\ref{prompt:segment-planner} and
Prompt~\ref{prompt:action-generator}, respectively. The turn-level
baseline used Prompt~\ref{prompt:turn-level}. In the data-scarce
setting, the teacher used Prompt~\ref{prompt:teacher-demonstration}
to generate demonstrations and Prompt~\ref{prompt:teacher-assessment}
to score candidate trajectories. We evaluated profile alignment using the LLM-judge template in
Prompt~\ref{prompt:evaluation-judge}.

Text enclosed in angle brackets denotes environment-, profile-, or
trajectory-specific information inserted at runtime. The same templates
were used in both environments with the corresponding
environment-specific representations and output schemas. 
These are prompt templates and the per environment examples can be found in the code supplemantary material.  Both architectures use a reason-then-act output contract. The \alps{}
planner prompt contains the environment description, the segment and
plan-schema definitions with worked examples, and the profile block;
the action-generator prompt contains the action definitions, the
current plan, running plan-consumption counters, and the environment
state, and never contains the profile representation. The turn-level
student's prompt contains the same profile representation the planner
receives, the environment description, the current state, the valid
actions, and running behavior summaries.

Two profile-block variants exist. The \emph{numeric} variant embeds
the quantitative statistics $q_p$ and is used in the data-sufficient
setting. The \emph{qualitative} variant replaces all numbers with
qualitative signatures and appends the $\dfew$ examples; it is used
by the teacher and in all data-scarce conditions. Full prompt templates
will be provided in our code release.
\begin{promptbox}{Segment-Planner Prompt Template}{prompt:segment-planner}
\small\ttfamily

You are the segment planner for a simulated student interacting with
the following learning environment.

\medskip
Environment description:

<environment description, action semantics, and segment definition>

\medskip
Target behavior profile:

<profile description>

\medskip
Session history:

<all previously generated segment plans, labelled by segment number>

\medskip
Running totals:

<cumulative counts across all prior segments and current phase>

\medskip
Next segment:

<segment number and available context>

\medskip
Plan the simulated student's behavior during the next segment.
Specify the intended quantity, quality, and structural properties
of the relevant actions.

Reason first, then return the plan:

<REASONING> ... </REASONING>
<answer> <environment-specific segment-plan schema> </answer>
\end{promptbox}

\begin{promptbox}{Action-Generator Prompt Template}{prompt:action-generator}
\small\ttfamily

You are generating the next action of a simulated student interacting
with the following learning environment.

\medskip
Environment description:

<environment description, allowed actions, and parameter constraints>

\medskip
Target behavior profile:

<profile description>

\medskip
Prior segments:

<summaries of previously completed segments>

\medskip
Current segment plan:

<segment plan>

\medskip
Current segment progress:

<actions completed so far against the plan's targets>

\medskip
Current environment state:

<environment state, mode, learned information, and running totals>

\medskip
Recent turns this segment:

<current-segment action history>

\medskip
Generate exactly one student action for the next turn, consistent with
the current segment plan and the environment's constraints on the
currently allowed actions.

Reason first, then return the action:

<REASONING> ... </REASONING>
<answer> <environment-specific action schema> </answer>
\end{promptbox}

\begin{promptbox}{Turn-Level Simulator Prompt Template}{prompt:turn-level}
\small\ttfamily

You are a simulated student interacting with the following learning
environment.

\medskip
Environment description:

<environment description, allowed actions, and parameter constraints>

\medskip
Target behavior profile:

<profile description>

\medskip
Current environment state and student progress:

<environment state, learned information, and running totals>

\medskip
Interaction history:

<previous states and generated actions>

\medskip
Generate exactly one action that is consistent with the target
behavior profile and the current environment state.

Reason first, then return the action:

<REASONING> ... </REASONING>
<answer> <environment-specific action schema> </answer>
\end{promptbox}

\begin{promptbox}{Teacher-Demonstration Prompt Template}{prompt:teacher-demonstration}
\small\ttfamily
D.1-D.3 +

\medskip
Few-shot examples:

<three real profile-labeled trajectories>

\end{promptbox}

\begin{promptbox}{Teacher Trajectory-Assessment Prompt Template}{prompt:teacher-assessment}
\small\ttfamily

Assess whether the generated trajectory faithfully represents the
specified behavior profile.

\medskip
Environment description:

<environment description>

\medskip
Target behavior profile:

<profile description, including its signature behaviors>

\medskip
Reference trajectories:

<three real trajectories from the target profile, each annotated with
signature notes>

\medskip
Contrast trajectories:

<two real trajectories from the most similar profiles, labeled>

\medskip
Feature table:

<per-axis behavioral counts for the candidate, the references and the
contrasts, with the reference min--max range per axis>

\medskip
Generated trajectory:

<generated student trajectory>

\medskip
Read the feature table and the per-segment summaries rather than
re-deriving counts from the trajectory. Answer every item below.

\medskip
Realism" judged against all real trajectories shown, references and
contrasts alike:

R1. the balance between high- and low-quality actions, and the overall
action volume, sit within the range the real trajectories show;

R2. repeated or unsuccessful attempts stay in character --- neither an
implausibly flawless run nor mechanical repetition beyond any real
trajectory;

R3. activity comes in episodes tied to task progress, rather than at a
rigid constant rate throughout;

R4. progression through the task tracks the student's actions the way
the real trajectories show.

\medskip
Profile match: judged against the single nearest reference:

TYP. the candidate's whole feature row could belong to one real student
of this profile, with leniency per axis set by whether the profile
description marks that axis as a signature or as incidental;

DISC. the candidate shares this profile's defining behaviors more than
any contrast profile's;

DIST. the structure of the session and the distribution of behavior across
the early, middle and late phases match the nearest reference and the
profile description;

TERM. the session concludes with an amount of effort neither cut short nor
extending well past where the nearest reference stops.

\medskip
Signature match: against this profile's defining signatures only:

SIG1--SIG$k$. each listed signature behavior holds.

\medskip
Answer every item "yes" or "no", each with one sentence of evidence
citing segment numbers or the feature table.

Return only:

<teacher-assessment schema>
\end{promptbox}

\begin{promptbox}{Profile-Alignment Judge Prompt Template}{prompt:evaluation-judge}
\small\ttfamily

The judge runs as two independent calls per trajectory. Each receives its own system
prompt and returns its own structured output; the assigned profile is withheld in the
first call and revealed only in the second.

\hrulefill

\textbf{Call 1: Blind profile identification}

\medskip
You are an expert rater of <environment> trajectories.

<environment description>

\medskip
The behavior profiles and their signatures:

<descriptions of all profiles, with their behavioral signatures>

\medskip
You can consult the following decision tree:

<decision tree over observable action counts>

\medskip
Trajectory to evaluate:

<rendered action sequence + measured facts>

\medskip
Classify the trajectory blind: give the single
most likely profile.

\begin{verbatim}
{
  "reasoning": "<short chain of evidence>",
  "guessed_profile": <profile identifier>
}
\end{verbatim}

\hrulefill

\textbf{Call 2: Signature checklist (assigned profile revealed)}

\medskip
<environment description>

\medskip
The assigned profile of this trajectory is <assigned profile>.

\medskip
Trajectory to evaluate:

<rendered action sequence + measured facts>

\medskip
For EACH signature of the assigned profile below, decide whether it is PRESENT in the
trajectory. Return one entry per signature, copying the signature text verbatim, with a
one-sentence justification citing the evidence. Judge each signature on its own merits
and only from the evidence in the trajectory.

<signatures of the assigned profile, each with its operational criterion>

\begin{verbatim}
{
  "signatures": [
    {"signature": "<verbatim>", 
    "present": <bool>, 
    "evidence": "<one sentence>"}
  ],
  "comment": "<overall note>"
}
\end{verbatim}

\end{promptbox}

\section{Additional Results}
\label{app:results}
This appendix reports the results that the main text summarizes:
the per-profile breakdown of both experiments
(Section~\ref{app:per-profile}); the effect of the two contents
of the profile prompt (Section~\ref{app:icl}); how profile alignment evolves
along a trajectory (Section~\ref{app:drift}); the ablation of the two \alps{}
adapters (Section~\ref{app:components-abl}); and the statistical tests behind
every significance claim (Section~\ref{sec:stats-tests}).

\subsection{Results per Profile}
\label{app:per-profile}
Tables~\ref{tab:per-profile-pharmasim}
and~\ref{tab:per-profile-tuglet} break the data-sufficient results
down by behavior profile, and
Tables~\ref{tab:per-profile-pharmasim-fs}
and~\ref{tab:per-profile-tuglet-fs} do the same for the data-scarce
setting. The four tables share one layout, and their rows follow
Tables~\ref{tab:results-limited} and~\ref{tab:results-fewshot}: each
row averaged over profiles reproduces the corresponding \emph{mean}
column of the main table. The tables also carry additional ICL rows
for the two ablated prompts; the aggregate prompt grid for both environments is
given in Tables~\ref{tab:prompt-variants-pharmasim}
and~\ref{tab:prompt-variants-tuglet}.
Profiles scored with $N{=}1$ (\pharmasim{}~P1 and~P5,
\tuglet{}~P4) saturate: any generator that reaches such a profile
records $100.0$ \coverage{}, and its \density{} is likewise optimistic.

Aggregate means hide which profiles a simulator abandons, and the
per-profile tables show that the two architectures fail in different places.
Under prompting, turn-level generation concentrates almost all of its mass on a
single \pharmasim{} profile and scores zero \density{} and zero judge pass rate
on three of the other four, whereas \alps{} reaches every profile
(Table~\ref{tab:per-profile-pharmasim}). On \tuglet{} the abandoned profile is
P1, the Thorough Learner, which every Qwen3 turn-level condition leaves near
zero on both metrics (Table~\ref{tab:per-profile-tuglet}). The worst-profile
columns of the main tables therefore report a structural failure rather than a
weak tail.
 
Fine-tuning does not lift the profiles uniformly, and it relocates
the bottleneck rather than removing it. On \pharmasim{}, adaptation repairs the
profile that prompting handles worst and leaves a different one, P3, as the
weakest. On \tuglet{} the bottleneck is P6, the Stubborn Grinder, and there
adaptation makes matters worse: both distributional metrics fall below their
prompted values while every other profile rises. Both bottleneck profiles are
defined by what the student does \emph{not} do, which a planner can express only
by suppressing behavior over many consecutive segments.
 
In the data-scarce setting, teacher demonstrations lift \alps{} on
every \pharmasim{} profile in \density{} and judge pass rate, and most sharply on
the profiles where prompting is weakest (Table~\ref{tab:per-profile-pharmasim-fs}).
P3 is again the exception, in an informative way: it is the profile on which the
demonstration-trained arms remain weakest distributionally (\density{} $18.4$ for
\alps{}~(SFT), against $92.5$ or above on every other profile), and the only
profile on which the reward-trained arm overtakes them on either distributional
metric (\density{} $33.4$ vs.\ $18.4$; \coverage{} $14.2$ vs.\ $12.2$), while
falling behind on all four others. The two teacher signals are therefore weakest
on different profiles: demonstrations on P3, the reward on P4.
 
\begin{table*}[t]
    \centering
    \footnotesize
    \setlength{\tabcolsep}{3pt}
    \renewcommand{\arraystretch}{0.8}
    \setlength{\aboverulesep}{0.05ex}
    \setlength{\belowrulesep}{0.15ex}
    {\footnotesize
    \begin{tabular*}{\textwidth}{@{\extracolsep{\fill}} c l l l l ccccc @{}}
        \toprule[1.2pt]
        & Model & Architecture & Adaptation & Signal Source & P1 & P2 & P3 & P4 & P5 \\
        \midrule[1pt]
        \multicolumn{10}{@{}l}{\textit{Clipped Density$\uparrow$}} \\
        \multirow{18}{*}{\rotatebox[origin=c]{90}{\textit{\pharmasim{}}}}
        & \multirow{6}{*}{GPT-OSS} & \multirow{3}{*}{Turn-level} & ICL & 3-shot & 0.0\sd{0.0} & 12.5\sd{1.4} & 0.0\sd{0.0} & 0.0\sd{0.0} & 100.0\sd{0.0} \\
        &  &  & ICL & 0-shot numeric & 1.7\sd{1.2} & 31.3\sd{4.3} & 0.0\sd{0.0} & 0.0\sd{0.0} & 100.0\sd{0.0} \\
        &  &  & ICL & 0-shot & 0.0\sd{0.0} & 31.5\sd{4.4} & 0.0\sd{0.0} & 0.0\sd{0.0} & 100.0\sd{0.0} \\
        \cmidrule(lr){4-10}
        &  & \multirow{3}{*}{\alps} & ICL & 3-shot & 100.0\sd{0.0} & 68.6\sd{3.5} & 58.5\sd{9.9} & 52.7\sd{11.2} & 100.0\sd{0.0} \\
        &  &  & ICL & 0-shot numeric & 100.0\sd{0.0} & 61.0\sd{2.0} & 66.6\sd{4.4} & 100.0\sd{0.0} & 100.0\sd{0.0} \\
        &  &  & ICL & 0-shot & 59.2\sd{5.1} & 100.0\sd{0.0} & 16.9\sd{6.8} & 100.0\sd{0.0} & 100.0\sd{0.0} \\
        \cmidrule(l){3-10}
        & \multirow{12}{*}{Qwen3} & \multirow{6}{*}{Turn-level} & ICL & 3-shot & 1.7\sd{1.2} & 84.7\sd{3.9} & 0.2\sd{0.3} & 0.3\sd{0.4} & 100.0\sd{0.0} \\
        &  &  & ICL & 0-shot numeric & 0.0\sd{0.0} & 26.2\sd{4.6} & 0.0\sd{0.0} & 0.0\sd{0.0} & 100.0\sd{0.0} \\
        &  &  & ICL & 0-shot & 0.0\sd{0.0} & 19.2\sd{6.2} & 0.0\sd{0.0} & 0.0\sd{0.0} & 100.0\sd{0.0} \\
        &  &  & RAFT & in-profile$^{\dagger}$ & \textit{0.0}\sd{0.0} & 41.1\sd{14.1} & 0.0\sd{0.0} & \textit{0.0}\sd{0.0} & 64.2\sd{5.5} \\
        &  &  & SFT & $\dlim$ & 38.0\sd{12.9} & 22.7\sd{8.4} & 28.4\sd{9.9} & 8.6\sd{5.2} & 90.7\sd{12.0} \\
        &  &  & SFT${\scriptstyle\rightarrow}$RAFT & $\dlim{,}$ in-profile & 31.3\sd{13.6} & 12.7\sd{1.4} & 23.8\sd{13.8} & 2.9\sd{2.9} & 75.9\sd{29.9} \\
        \cmidrule(lr){4-10}
        &  & \multirow{6}{*}{\alps} & ICL & 3-shot & 6.7\sd{1.2} & 29.9\sd{3.5} & 7.9\sd{0.9} & 9.2\sd{6.0} & 100.0\sd{0.0} \\
        &  &  & ICL & 0-shot numeric & 27.5\sd{7.1} & 37.9\sd{1.2} & 5.4\sd{1.4} & 10.5\sd{6.9} & 100.0\sd{0.0} \\
        &  &  & ICL & 0-shot & 0.0\sd{0.0} & 84.2\sd{9.0} & 0.2\sd{0.3} & 10.0\sd{2.2} & 100.0\sd{0.0} \\
        &  &  & RAFT & in-profile & 37.7\sd{22.3} & 16.1\sd{2.0} & 9.3\sd{3.6} & 19.7\sd{11.4} & 100.0\sd{0.0} \\
        &  &  & SFT & $\dlim$ & 78.4\sd{14.4} & 59.9\sd{10.1} & 30.8\sd{6.3} & 62.8\sd{20.9} & 92.5\sd{10.6} \\
        &  &  & SFT${\scriptstyle\rightarrow}$RAFT & $\dlim{,}$ in-profile & 92.7\sd{9.6} & 62.2\sd{11.0} & 36.2\sd{7.5} & 65.3\sd{21.7} & 92.3\sd{10.0} \\
        \specialrule{1.2pt}{0.2ex}{0.2ex}
        \multicolumn{10}{@{}l}{\textit{Clipped Coverage$\uparrow$}} \\
        \multirow{18}{*}{\rotatebox[origin=c]{90}{\textit{\pharmasim{}}}}
        & \multirow{6}{*}{GPT-OSS} & \multirow{3}{*}{Turn-level} & ICL & 3-shot & 0.0\sd{0.0} & 3.0\sd{0.0} & 1.0\sd{1.4} & 3.0\sd{0.0} & 4.0\sd{2.1} \\
        &  &  & ICL & 0-shot numeric & 1.3\sd{0.9} & 6.6\sd{0.5} & 0.0\sd{0.0} & 1.0\sd{1.4} & 7.3\sd{6.1} \\
        &  &  & ICL & 0-shot & 0.0\sd{0.0} & 5.3\sd{0.5} & 0.0\sd{0.0} & 3.0\sd{0.0} & 11.6\sd{6.1} \\
        \cmidrule(lr){4-10}
        &  & \multirow{3}{*}{\alps} & ICL & 3-shot & 5.8\sd{2.0} & 5.9\sd{0.0} & 17.8\sd{1.4} & 15.8\sd{1.6} & 3.0\sd{0.0} \\
        &  &  & ICL & 0-shot numeric & 6.9\sd{0.0} & 5.6\sd{0.5} & 19.8\sd{0.8} & 17.5\sd{0.5} & 3.0\sd{0.0} \\
        &  &  & ICL & 0-shot & 9.9\sd{4.2} & 5.0\sd{0.0} & 4.3\sd{0.5} & 11.2\sd{0.5} & 3.0\sd{0.0} \\
        \cmidrule(l){3-10}
        & \multirow{12}{*}{Qwen3} & \multirow{6}{*}{Turn-level} & ICL & 3-shot & 2.6\sd{0.5} & 3.6\sd{0.9} & 3.0\sd{0.0} & 2.0\sd{1.4} & 9.9\sd{4.2} \\
        &  &  & ICL & 0-shot numeric & 0.7\sd{0.9} & 44.6\sd{39.4} & 1.0\sd{1.4} & 0.0\sd{0.0} & 40.9\sd{41.9} \\
        &  &  & ICL & 0-shot & 0.0\sd{0.0} & 11.6\sd{2.3} & 0.0\sd{0.0} & 0.0\sd{0.0} & 71.9\sd{39.7} \\
        &  &  & RAFT & in-profile$^{\dagger}$ & \textit{0.7\sd{0.9}} & 46.3\sd{38.4} & 1.1\sd{1.5} & \textit{0.0}\sd{0.0} & 20.5\sd{7.5} \\
        &  &  & SFT & $\dlim$ & 69.0\sd{43.9} & 18.7\sd{13.0} & 18.8\sd{6.4} & 8.3\sd{3.8} & 100.0\sd{0.0} \\
        &  &  & SFT${\scriptstyle\rightarrow}$RAFT & $\dlim{,}$ in-profile & 39.5\sd{43.1} & 7.5\sd{1.3} & 12.8\sd{6.0} & 4.4\sd{4.1} & 71.9\sd{39.7} \\
        \cmidrule(lr){4-10}
        &  & \multirow{6}{*}{\alps} & ICL & 3-shot & 7.3\sd{6.1} & 3.0\sd{0.0} & 10.2\sd{1.2} & 10.9\sd{3.7} & 9.9\sd{4.2} \\
        &  &  & ICL & 0-shot numeric & 3.0\sd{0.0} & 3.0\sd{0.0} & 11.2\sd{0.9} & 9.6\sd{3.3} & 3.0\sd{0.0} \\
        &  &  & ICL & 0-shot & 2.3\sd{0.5} & 3.3\sd{0.5} & 3.0\sd{0.0} & 7.6\sd{0.9} & 3.0\sd{0.0} \\
        &  &  & RAFT & in-profile & 4.5\sd{1.9} & 3.0\sd{0.0} & 11.0\sd{1.7} & 15.3\sd{6.1} & 3.0\sd{0.0} \\
        &  &  & SFT & $\dlim$ & 72.3\sd{39.2} & 17.9\sd{3.0} & 31.0\sd{11.2} & 33.6\sd{8.2} & 100.0\sd{0.0} \\
        &  &  & SFT${\scriptstyle\rightarrow}$RAFT & $\dlim{,}$ in-profile & 100.0\sd{0.0} & 22.5\sd{3.5} & 34.1\sd{18.4} & 34.2\sd{5.5} & 100.0\sd{0.0} \\
        \specialrule{1.2pt}{0.2ex}{0.2ex}
        \multicolumn{10}{@{}l}{\textit{Judge pass rate$\uparrow$}} \\
        \multirow{18}{*}{\rotatebox[origin=c]{90}{\textit{\pharmasim{}}}}
        & \multirow{6}{*}{GPT-OSS} & \multirow{3}{*}{Turn-level} & ICL & 3-shot & 0.0\sd{0.0} & 98.0\sd{0.8} & 0.0\sd{0.0} & 0.0\sd{0.0} & 100.0\sd{0.0} \\
        &  &  & ICL & 0-shot numeric & 0.3\sd{0.5} & 89.0\sd{2.2} & 0.0\sd{0.0} & 0.0\sd{0.0} & 99.3\sd{0.5} \\
        &  &  & ICL & 0-shot & 0.0\sd{0.0} & 86.7\sd{1.7} & 0.0\sd{0.0} & 0.0\sd{0.0} & 100.0\sd{0.0} \\
        \cmidrule(lr){4-10}
        &  & \multirow{3}{*}{\alps} & ICL & 3-shot & 86.9\sd{1.5} & 100.0\sd{0.0} & 98.3\sd{1.7} & 92.7\sd{1.7} & 100.0\sd{0.0} \\
        &  &  & ICL & 0-shot numeric & 69.7\sd{5.6} & 97.7\sd{0.5} & 79.0\sd{2.9} & 47.7\sd{3.3} & 100.0\sd{0.0} \\
        &  &  & ICL & 0-shot & 60.7\sd{0.5} & 98.3\sd{1.2} & 59.3\sd{3.9} & 63.0\sd{4.3} & 100.0\sd{0.0} \\
        \cmidrule(l){3-10}
        & \multirow{12}{*}{Qwen3} & \multirow{6}{*}{Turn-level} & ICL & 3-shot & 1.0\sd{0.8} & 99.7\sd{0.5} & 19.7\sd{2.9} & 12.7\sd{2.6} & 100.0\sd{0.0} \\
        &  &  & ICL & 0-shot numeric & 0.0\sd{0.0} & 40.0\sd{2.2} & 6.3\sd{1.2} & 0.0\sd{0.0} & 92.0\sd{1.4} \\
        &  &  & ICL & 0-shot & 0.0\sd{0.0} & 31.3\sd{8.2} & 10.3\sd{2.1} & 0.0\sd{0.0} & 90.7\sd{2.9} \\
        &  &  & RAFT & in-profile$^{\dagger}$ & \textit{0.0}\sd{0.0} & 43.4\sd{11.2} & 11.4\sd{1.3} & \textit{0.0}\sd{0.0} & 86.0\sd{1.4} \\
        &  &  & SFT & $\dlim$ & 28.0\sd{14.5} & 57.9\sd{13.4} & 62.0\sd{12.8} & 32.4\sd{13.3} & 85.6\sd{2.1} \\
        &  &  & SFT${\scriptstyle\rightarrow}$RAFT & $\dlim{,}$ in-profile & 20.4\sd{15.2} & 35.8\sd{16.9} & 48.9\sd{25.7} & 24.8\sd{19.8} & 91.4\sd{5.9} \\
        \cmidrule(lr){4-10}
        &  & \multirow{6}{*}{\alps} & ICL & 3-shot & 18.0\sd{5.9} & 61.0\sd{4.5} & 41.0\sd{3.3} & 8.7\sd{3.4} & 81.0\sd{2.2} \\
        &  &  & ICL & 0-shot numeric & 49.7\sd{4.0} & 99.7\sd{0.5} & 43.3\sd{3.4} & 53.7\sd{2.9} & 90.0\sd{1.4} \\
        &  &  & ICL & 0-shot & 11.3\sd{2.6} & 97.0\sd{1.4} & 3.3\sd{1.2} & 36.0\sd{1.6} & 95.7\sd{1.7} \\
        &  &  & RAFT & in-profile & 56.0\sd{8.2} & 100.0\sd{0.0} & 40.6\sd{6.5} & 46.9\sd{8.8} & 99.3\sd{0.9} \\
        &  &  & SFT & $\dlim$ & 34.7\sd{18.7} & 92.3\sd{2.4} & 56.8\sd{14.9} & 86.9\sd{7.2} & 92.6\sd{7.1} \\
        &  &  & SFT${\scriptstyle\rightarrow}$RAFT & $\dlim{,}$ in-profile & 41.6\sd{20.4} & 93.3\sd{2.0} & 61.4\sd{5.3} & 90.0\sd{7.3} & 92.5\sd{6.0} \\
        \bottomrule[1.2pt]
    \end{tabular*}
    }
    \captionsetup{skip=3pt}
    \caption{\textbf{Exp1: Data-sufficient setting, per profile (\pharmasim{}).}
Simulation quality per behavior profile, as percentages
(mean$_{\pm\mathrm{std}}$ over three seeds, 100 trajectories per profile
per seed). Rows follow Table~\ref{tab:results-limited}, and each row
averaged over the five profiles reproduces the corresponding \emph{mean}
column there. Rows in the GPT-OSS block are the prompted GPT-OSS
reference; all other rows are Qwen3. The ICL rows add two ablated prompts to the \emph{3-shot} condition of the main table: \emph{0-shot numeric} (numeric profile descriptions, no examples) and \emph{0-shot} (neither).
$^{\dagger}$~Turn-level RAFT produced no trajectories for profiles~1 and~4 in any replica, so no adapter was trained for them; those cells (\textit{italic}) are filled per replica with the prompted turn-level baseline from the \emph{0-shot numeric} condition.}
\label{tab:per-profile-pharmasim}
\end{table*}

\begin{table*}[t]
    \centering
    \footnotesize
    \setlength{\tabcolsep}{3pt}
    \renewcommand{\arraystretch}{0.8}
    \setlength{\aboverulesep}{0.05ex}
    \setlength{\belowrulesep}{0.15ex}
    {\footnotesize
    \begin{tabular*}{\textwidth}{@{\extracolsep{\fill}} c l l l l cccccc @{}}
        \toprule[1.2pt]
        & Model & Architecture & Adaptation & Signal Source & P1 & P2 & P3 & P4 & P5 & P6 \\
        \midrule[1pt]
        \multicolumn{11}{@{}l}{\textit{Clipped Density$\uparrow$}} \\
        \multirow{18}{*}{\rotatebox[origin=c]{90}{\textit{\tuglet{}}}}
        & \multirow{6}{*}{GPT-OSS} & \multirow{3}{*}{Turn-level} & ICL & 3-shot & 2.0\sd{2.0} & 33.7\sd{9.6} & 65.2\sd{3.3} & 21.7\sd{12.5} & 83.6\sd{11.1} & 86.2\sd{9.3} \\
        &  &  & ICL & 0-shot numeric & 6.5\sd{1.3} & 36.0\sd{5.2} & 70.6\sd{2.7} & 28.3\sd{7.2} & 4.8\sd{0.3} & 0.0\sd{0.0} \\
        &  &  & ICL & 0-shot & 3.4\sd{2.0} & 26.6\sd{3.6} & 68.0\sd{1.4} & 75.8\sd{15.0} & 19.0\sd{10.3} & 3.4\sd{3.3} \\
        \cmidrule(lr){4-11}
        &  & \multirow{3}{*}{\alps} & ICL & 3-shot & 42.4\sd{6.5} & 100.0\sd{0.0} & 100.0\sd{0.0} & 100.0\sd{0.0} & 100.0\sd{0.0} & 61.3\sd{4.0} \\
        &  &  & ICL & 0-shot numeric & 21.0\sd{2.4} & 83.9\sd{5.7} & 100.0\sd{0.0} & 100.0\sd{0.0} & 93.8\sd{8.8} & 63.6\sd{1.9} \\
        &  &  & ICL & 0-shot & 23.8\sd{4.4} & 34.5\sd{5.4} & 89.4\sd{3.0} & 18.3\sd{5.9} & 90.5\sd{4.4} & 0.0\sd{0.0} \\
        \cmidrule(l){3-11}
        & \multirow{12}{*}{Qwen3} & \multirow{6}{*}{Turn-level} & ICL & 3-shot & 0.4\sd{0.6} & 16.7\sd{6.0} & 47.7\sd{6.9} & 98.3\sd{2.4} & 28.1\sd{6.1} & 99.1\sd{1.2} \\
        &  &  & ICL & 0-shot numeric & 1.6\sd{0.7} & 26.4\sd{1.6} & 4.7\sd{1.0} & 0.0\sd{0.0} & 0.0\sd{0.0} & 1.0\sd{1.0} \\
        &  &  & ICL & 0-shot & 1.4\sd{0.3} & 19.3\sd{1.1} & 10.2\sd{2.4} & 32.5\sd{5.4} & 17.1\sd{3.6} & 100.0\sd{0.0} \\
        &  &  & RAFT & in-profile & 2.0\sd{2.0} & 10.6\sd{0.5} & 6.8\sd{0.9} & 14.2\sd{5.1} & \textit{0.0\sd{0.0}} & \textit{1.0\sd{1.0}} \\
        &  &  & SFT & $\dlim$ & 0.0\sd{0.0} & 23.4\sd{4.5} & 77.8\sd{10.3} & 3.3\sd{3.1} & 57.1\sd{1.0} & 11.8\sd{1.3} \\
        &  &  & SFT${\scriptstyle\rightarrow}$RAFT & $\dlim{,}$ in-profile & 4.2\sd{3.0} & 30.1\sd{8.5} & 73.9\sd{6.1} & 25.8\sd{20.4} & 44.8\sd{20.1} & 12.0\sd{7.1} \\
        \cmidrule(lr){4-11}
        &  & \multirow{6}{*}{\alps} & ICL & 3-shot & 18.0\sd{3.0} & 14.8\sd{1.3} & 8.3\sd{1.5} & 4.2\sd{2.4} & 21.4\sd{6.3} & 27.5\sd{2.8} \\
        &  &  & ICL & 0-shot numeric & 12.3\sd{1.1} & 60.1\sd{5.3} & 29.5\sd{4.4} & 30.8\sd{7.7} & 12.6\sd{5.3} & 7.9\sd{2.3} \\
        &  &  & ICL & 0-shot & 3.2\sd{1.6} & 22.7\sd{0.6} & 54.5\sd{1.9} & 0.8\sd{1.2} & 6.9\sd{1.8} & 1.3\sd{1.3} \\
        &  &  & RAFT & in-profile & 12.3\sd{6.1} & 66.0\sd{3.9} & 59.7\sd{16.9} & 45.8\sd{7.2} & 22.9\sd{9.1} & 2.1\sd{1.3} \\
        &  &  & SFT & $\dlim$ & 31.5\sd{5.3} & 50.4\sd{14.2} & 58.5\sd{14.2} & 59.2\sd{21.4} & 55.7\sd{6.9} & 22.8\sd{5.0} \\
        &  &  & SFT${\scriptstyle\rightarrow}$RAFT & $\dlim{,}$ in-profile & 23.9\sd{6.5} & 64.6\sd{5.8} & 68.6\sd{22.3} & 53.3\sd{17.4} & 95.7\sd{4.2} & 11.3\sd{7.1} \\
        \specialrule{1.2pt}{0.2ex}{0.2ex}
        \multicolumn{11}{@{}l}{\textit{Clipped Coverage$\uparrow$}} \\
        \multirow{18}{*}{\rotatebox[origin=c]{90}{\textit{\tuglet{}}}}
        & \multirow{6}{*}{GPT-OSS} & \multirow{3}{*}{Turn-level} & ICL & 3-shot & 3.3\sd{0.5} & 10.6\sd{0.5} & 11.2\sd{1.7} & 67.7\sd{45.7} & 100.0\sd{0.0} & 100.0\sd{0.0} \\
        &  &  & ICL & 0-shot numeric & 6.3\sd{1.2} & 9.2\sd{0.9} & 6.6\sd{0.5} & 15.8\sd{0.0} & 3.0\sd{0.0} & 1.0\sd{1.4} \\
        &  &  & ICL & 0-shot & 4.3\sd{0.9} & 8.3\sd{0.5} & 6.3\sd{0.5} & 40.9\sd{41.9} & 4.6\sd{1.2} & 6.6\sd{6.2} \\
        \cmidrule(lr){4-11}
        &  & \multirow{3}{*}{\alps} & ICL & 3-shot & 11.9\sd{0.8} & 29.4\sd{0.9} & 14.5\sd{0.5} & 43.9\sd{39.7} & 100.0\sd{0.0} & 9.9\sd{0.0} \\
        &  &  & ICL & 0-shot numeric & 9.9\sd{1.4} & 33.7\sd{0.8} & 16.8\sd{5.7} & 100.0\sd{0.0} & 15.2\sd{3.3} & 10.9\sd{0.8} \\
        &  &  & ICL & 0-shot & 9.2\sd{0.5} & 17.2\sd{3.3} & 5.6\sd{0.5} & 4.3\sd{1.9} & 18.2\sd{4.1} & 0.0\sd{0.0} \\
        \cmidrule(l){3-11}
        & \multirow{12}{*}{Qwen3} & \multirow{6}{*}{Turn-level} & ICL & 3-shot & 1.0\sd{1.4} & 13.9\sd{2.4} & 100.0\sd{0.0} & 100.0\sd{0.0} & 100.0\sd{0.0} & 100.0\sd{0.0} \\
        &  &  & ICL & 0-shot numeric & 3.3\sd{0.5} & 16.2\sd{0.9} & 5.9\sd{0.8} & 0.0\sd{0.0} & 0.0\sd{0.0} & 3.0\sd{0.0} \\
        &  &  & ICL & 0-shot & 4.6\sd{2.3} & 13.9\sd{1.6} & 12.9\sd{4.0} & 100.0\sd{0.0} & 100.0\sd{0.0} & 100.0\sd{0.0} \\
        &  &  & RAFT & in-profile & 2.3\sd{1.7} & 7.3\sd{0.9} & 4.6\sd{1.2} & 15.8\sd{0.0} & \textit{0.0\sd{0.0}} & \textit{3.0\sd{0.0}} \\
        &  &  & SFT & $\dlim$ & 1.0\sd{1.4} & 18.2\sd{2.3} & 22.4\sd{2.0} & 3.6\sd{2.3} & 11.6\sd{2.6} & 9.2\sd{0.9} \\
        &  &  & SFT${\scriptstyle\rightarrow}$RAFT & $\dlim{,}$ in-profile & 5.6\sd{4.1} & 19.8\sd{8.2} & 15.5\sd{2.0} & 4.0\sd{2.1} & 11.9\sd{2.9} & 8.3\sd{4.9} \\
        \cmidrule(lr){4-11}
        &  & \multirow{6}{*}{\alps} & ICL & 3-shot & 8.3\sd{1.9} & 6.9\sd{0.8} & 4.3\sd{0.5} & 3.0\sd{0.0} & 5.9\sd{1.4} & 52.8\sd{33.9} \\
        &  &  & ICL & 0-shot numeric & 8.3\sd{0.9} & 5.6\sd{0.5} & 10.2\sd{1.9} & 15.8\sd{0.0} & 5.0\sd{0.8} & 6.3\sd{0.5} \\
        &  &  & ICL & 0-shot & 4.3\sd{1.2} & 3.6\sd{0.5} & 10.2\sd{1.7} & 0.7\sd{0.9} & 3.0\sd{0.0} & 2.3\sd{1.7} \\
        &  &  & RAFT & in-profile & 6.6\sd{1.7} & 5.9\sd{0.8} & 10.6\sd{1.2} & 9.9\sd{4.2} & 7.9\sd{2.1} & 4.0\sd{0.8} \\
        &  &  & SFT & $\dlim$ & 42.2\sd{15.9} & 43.9\sd{7.9} & 28.7\sd{3.7} & 100.0\sd{0.0} & 54.5\sd{32.2} & 16.8\sd{4.0} \\
        &  &  & SFT${\scriptstyle\rightarrow}$RAFT & $\dlim{,}$ in-profile & 22.8\sd{9.9} & 53.8\sd{8.9} & 25.1\sd{2.5} & 71.9\sd{39.7} & 100.0\sd{0.0} & 5.0\sd{0.8} \\
        \specialrule{1.2pt}{0.2ex}{0.2ex}
        \multicolumn{11}{@{}l}{\textit{Judge pass rate$\uparrow$}} \\
        \multirow{18}{*}{\rotatebox[origin=c]{90}{\textit{\tuglet{}}}}
        & \multirow{6}{*}{GPT-OSS} & \multirow{3}{*}{Turn-level} & ICL & 3-shot & 1.0\sd{0.0} & 22.7\sd{0.9} & 89.0\sd{1.4} & 11.0\sd{0.8} & 0.0\sd{0.0} & 0.0\sd{0.0} \\
        &  &  & ICL & 0-shot numeric & 5.3\sd{1.2} & 19.0\sd{2.8} & 99.7\sd{0.5} & 3.0\sd{0.8} & 0.0\sd{0.0} & 0.7\sd{0.9} \\
        &  &  & ICL & 0-shot & 0.7\sd{0.9} & 8.7\sd{1.7} & 98.7\sd{0.5} & 8.7\sd{2.1} & 0.0\sd{0.0} & 0.3\sd{0.5} \\
        \cmidrule(lr){4-11}
        &  & \multirow{3}{*}{\alps} & ICL & 3-shot & 67.3\sd{6.8} & 92.0\sd{2.4} & 95.7\sd{1.7} & 76.7\sd{1.9} & 11.0\sd{1.6} & 75.7\sd{2.9} \\
        &  &  & ICL & 0-shot numeric & 57.0\sd{1.4} & 74.0\sd{5.7} & 91.7\sd{1.9} & 74.0\sd{2.8} & 7.0\sd{1.6} & 72.3\sd{4.9} \\
        &  &  & ICL & 0-shot & 51.3\sd{10.5} & 26.3\sd{5.4} & 100.0\sd{0.0} & 0.0\sd{0.0} & 3.0\sd{1.4} & 0.7\sd{0.5} \\
        \cmidrule(l){3-11}
        & \multirow{12}{*}{Qwen3} & \multirow{6}{*}{Turn-level} & ICL & 3-shot & 0.0\sd{0.0} & 15.7\sd{0.9} & 25.7\sd{6.2} & 0.0\sd{0.0} & 0.0\sd{0.0} & 0.0\sd{0.0} \\
        &  &  & ICL & 0-shot numeric & 0.7\sd{0.5} & 37.7\sd{4.9} & 14.7\sd{4.6} & 0.0\sd{0.0} & 0.0\sd{0.0} & 14.7\sd{0.5} \\
        &  &  & ICL & 0-shot & 0.3\sd{0.5} & 22.7\sd{1.7} & 25.0\sd{5.4} & 0.7\sd{0.9} & 0.0\sd{0.0} & 0.7\sd{0.9} \\
        &  &  & RAFT & in-profile & 0.3\sd{0.5} & 16.0\sd{2.8} & 12.3\sd{2.1} & 2.7\sd{2.1} & \textit{0.0\sd{0.0}} & \textit{14.7\sd{0.5}} \\
        &  &  & SFT & $\dlim$ & 0.0\sd{0.0} & 25.3\sd{4.5} & 73.3\sd{5.6} & 0.7\sd{0.9} & 2.0\sd{2.2} & 14.0\sd{1.6} \\
        &  &  & SFT${\scriptstyle\rightarrow}$RAFT & $\dlim{,}$ in-profile & 2.0\sd{2.8} & 34.3\sd{15.7} & 91.0\sd{6.4} & 1.7\sd{2.4} & 1.0\sd{0.8} & 12.7\sd{2.5} \\
        \cmidrule(lr){4-11}
        &  & \multirow{6}{*}{\alps} & ICL & 3-shot & 39.3\sd{0.5} & 6.3\sd{0.5} & 9.7\sd{0.9} & 11.0\sd{6.2} & 4.7\sd{2.1} & 21.7\sd{3.9} \\
        &  &  & ICL & 0-shot numeric & 14.3\sd{0.5} & 7.0\sd{2.9} & 38.7\sd{4.7} & 12.0\sd{2.9} & 22.0\sd{2.9} & 20.3\sd{4.5} \\
        &  &  & ICL & 0-shot & 15.7\sd{1.7} & 2.3\sd{0.5} & 58.7\sd{1.2} & 0.7\sd{0.5} & 0.3\sd{0.5} & 0.0\sd{0.0} \\
        &  &  & RAFT & in-profile & 18.0\sd{3.6} & 4.3\sd{2.5} & 65.0\sd{19.6} & 22.0\sd{9.9} & 13.7\sd{1.7} & 32.3\sd{3.3} \\
        &  &  & SFT & $\dlim$ & 40.0\sd{5.9} & 50.7\sd{9.0} & 58.0\sd{9.0} & 45.0\sd{12.2} & 56.0\sd{7.5} & 35.7\sd{10.4} \\
        &  &  & SFT${\scriptstyle\rightarrow}$RAFT & $\dlim{,}$ in-profile & 15.7\sd{13.5} & 57.7\sd{5.6} & 59.3\sd{20.4} & 41.0\sd{12.0} & 42.3\sd{9.0} & 31.3\sd{13.3} \\
        \bottomrule[1.2pt]
    \end{tabular*}
    }
    \captionsetup{skip=3pt}
    \caption{\textbf{Exp1: Data-sufficient setting, per profile (\tuglet{}).}
Simulation quality per behavior profile, as percentages
(mean$_{\pm\mathrm{std}}$ over three seeds, 100 trajectories per profile
per seed). Rows follow Table~\ref{tab:results-limited}, and each row
averaged over the six profiles reproduces the corresponding \emph{mean}
column there. The ICL rows add two ablated
prompts to the \emph{3-shot} condition of the main table: \emph{0-shot numeric}
(numeric profile descriptions, no examples) and \emph{0-shot} (neither).
Turn-level RAFT produced no trajectories for profiles~5 and~6 in any replica, so no adapter was trained for them; those cells (\textit{italic}) are filled per replica with the prompted turn-level baseline from the \emph{0-shot numeric} condition.}
    \label{tab:per-profile-tuglet}
\end{table*}

\begin{table*}[t]
    \centering
    \footnotesize
    \setlength{\tabcolsep}{3pt}
    \renewcommand{\arraystretch}{0.94}
    \setlength{\aboverulesep}{0.05ex}
    \setlength{\belowrulesep}{0.15ex}
    {\footnotesize
    \begin{tabular*}{\textwidth}{@{\extracolsep{\fill}} c l l l l ccccc @{}}
        \toprule[1.2pt]
        & Model & Architecture & Adaptation & Signal Source & P1 & P2 & P3 & P4 & P5 \\
        \midrule[1pt]
        \multicolumn{10}{@{}l}{\textit{Clipped Density$\uparrow$}} \\
        \multirow{14}{*}{\rotatebox[origin=c]{90}{\textit{\pharmasim{}}}}
        & \multirow{4}{*}{GPT-OSS} & \multirow{2}{*}{Turn-level} & ICL & $\dfew$ & 0.0\sd{0.0} & 14.7\sd{1.1} & 0.0\sd{0.0} & 0.0\sd{0.0} & 100.0\sd{0.0} \\
        &  &  & ICL & 0-shot & 0.0\sd{0.0} & 31.5\sd{4.4} & 0.0\sd{0.0} & 0.0\sd{0.0} & 100.0\sd{0.0} \\
        \cmidrule(lr){4-10}
        &  & \multirow{2}{*}{\alps} & ICL & $\dfew$ & 100.0\sd{0.0} & 100.0\sd{0.0} & 18.4\sd{3.3} & 100.0\sd{0.0} & 100.0\sd{0.0} \\
        &  &  & ICL & 0-shot & 59.2\sd{5.1} & 100.0\sd{0.0} & 16.9\sd{6.8} & 100.0\sd{0.0} & 100.0\sd{0.0} \\
        \cmidrule(l){3-10}
        & \multirow{10}{*}{Qwen3} & \multirow{5}{*}{Turn-level} & ICL & $\dfew$ & 0.0\sd{0.0} & 84.0\sd{1.9} & 0.0\sd{0.0} & 0.0\sd{0.0} & 100.0\sd{0.0} \\
        &  &  & ICL & 0-shot & 0.0\sd{0.0} & 19.2\sd{6.2} & 0.0\sd{0.0} & 0.0\sd{0.0} & 100.0\sd{0.0} \\
        &  &  & RAFT & TR$^{\dagger}$ & \textit{0.0}\sd{0.0} & 83.6\sd{11.9} & 13.5\sd{3.8} & 13.1\sd{4.5} & 100.0\sd{0.0} \\
        &  &  & SFT & TD$^{\dagger}$ & \textit{0.0}\sd{0.0} & 50.2\sd{23.9} & \textit{0.0}\sd{0.0} & \textit{0.0}\sd{0.0} & 100.0\sd{0.0} \\
        &  &  & SFT${\scriptstyle\rightarrow}$RAFT & TD${\scriptstyle\rightarrow}$TR$^{\dagger}$ & \textit{0.0}\sd{0.0} & 52.2\sd{36.4} & \textit{0.0}\sd{0.0} & \textit{0.0}\sd{0.0} & 100.0\sd{0.0} \\
        \cmidrule(lr){4-10}
        &  & \multirow{5}{*}{\alps} & ICL & $\dfew$ & 17.5\sd{4.1} & 75.0\sd{1.6} & 3.2\sd{0.9} & 8.7\sd{2.2} & 82.5\sd{2.0} \\
        &  &  & ICL & 0-shot & 0.0\sd{0.0} & 84.2\sd{9.0} & 0.2\sd{0.3} & 10.0\sd{2.2} & 100.0\sd{0.0} \\
        &  &  & RAFT & TR & 71.7\sd{28.6} & 100.0\sd{0.0} & 33.4\sd{4.1} & 17.7\sd{11.1} & 84.2\sd{13.3} \\
        &  &  & SFT & TD & 92.5\sd{7.4} & 100.0\sd{0.0} & 18.4\sd{3.8} & 100.0\sd{0.0} & 100.0\sd{0.0} \\
        &  &  & SFT${\scriptstyle\rightarrow}$RAFT & TD${\scriptstyle\rightarrow}$TR & 89.2\sd{15.3} & 100.0\sd{0.0} & 13.5\sd{4.9} & 100.0\sd{0.0} & 100.0\sd{0.0} \\
        \specialrule{1.2pt}{0.2ex}{0.2ex}
        \multicolumn{10}{@{}l}{\textit{Clipped Coverage$\uparrow$}} \\
        \multirow{14}{*}{\rotatebox[origin=c]{90}{\textit{\pharmasim{}}}}
        & \multirow{4}{*}{GPT-OSS} & \multirow{2}{*}{Turn-level} & ICL & $\dfew$ & 0.0\sd{0.0} & 3.0\sd{0.0} & 2.0\sd{1.4} & 2.0\sd{1.4} & 2.6\sd{0.5} \\
        &  &  & ICL & 0-shot & 0.0\sd{0.0} & 5.3\sd{0.5} & 0.0\sd{0.0} & 3.0\sd{0.0} & 11.6\sd{6.1} \\
        \cmidrule(lr){4-10}
        &  & \multirow{2}{*}{\alps} & ICL & $\dfew$ & 100.0\sd{0.0} & 5.0\sd{0.8} & 10.9\sd{1.4} & 10.6\sd{0.9} & 3.0\sd{0.0} \\
        &  &  & ICL & 0-shot & 9.9\sd{4.2} & 5.0\sd{0.0} & 4.3\sd{0.5} & 11.2\sd{0.5} & 3.0\sd{0.0} \\
        \cmidrule(l){3-10}
        & \multirow{10}{*}{Qwen3} & \multirow{5}{*}{Turn-level} & ICL & $\dfew$ & 2.0\sd{0.0} & 3.6\sd{0.5} & 3.0\sd{0.0} & 1.0\sd{1.4} & 38.0\sd{43.9} \\
        &  &  & ICL & 0-shot & 0.0\sd{0.0} & 11.6\sd{2.3} & 0.0\sd{0.0} & 0.0\sd{0.0} & 71.9\sd{39.7} \\
        &  &  & RAFT & TR$^{\dagger}$ & \textit{2.0}\sd{0.0} & 4.3\sd{0.5} & 9.3\sd{3.8} & 10.6\sd{2.0} & 43.9\sd{39.7} \\
        &  &  & SFT & TD$^{\dagger}$ & \textit{2.0}\sd{0.0} & 4.0\sd{0.8} & \textit{3.0}\sd{0.0} & \textit{1.0}\sd{1.4} & 2.3\sd{0.5} \\
        &  &  & SFT${\scriptstyle\rightarrow}$RAFT & TD${\scriptstyle\rightarrow}$TR$^{\dagger}$ & \textit{2.0}\sd{0.0} & 3.6\sd{0.5} & \textit{3.0}\sd{0.0} & \textit{1.0}\sd{1.4} & 2.0\sd{0.0} \\
        \cmidrule(lr){4-10}
        &  & \multirow{5}{*}{\alps} & ICL & $\dfew$ & 4.3\sd{1.9} & 4.6\sd{0.5} & 6.3\sd{0.5} & 7.6\sd{1.7} & 3.0\sd{0.0} \\
        &  &  & ICL & 0-shot & 2.3\sd{0.5} & 3.3\sd{0.5} & 3.0\sd{0.0} & 7.6\sd{0.9} & 3.0\sd{0.0} \\
        &  &  & RAFT & TR & 43.9\sd{39.7} & 3.0\sd{0.0} & 14.2\sd{1.7} & 10.2\sd{4.1} & 2.6\sd{0.5} \\
        &  &  & SFT & TD & 100.0\sd{0.0} & 3.6\sd{0.5} & 12.2\sd{1.7} & 11.2\sd{0.5} & 3.0\sd{0.0} \\
        &  &  & SFT${\scriptstyle\rightarrow}$RAFT & TD${\scriptstyle\rightarrow}$TR & 100.0\sd{0.0} & 3.0\sd{0.0} & 8.6\sd{1.9} & 10.6\sd{1.2} & 3.0\sd{0.0} \\
        \specialrule{1.2pt}{0.2ex}{0.2ex}
        \multicolumn{10}{@{}l}{\textit{Judge pass rate$\uparrow$}} \\
        \multirow{14}{*}{\rotatebox[origin=c]{90}{\textit{\pharmasim{}}}}
        & \multirow{4}{*}{GPT-OSS} & \multirow{2}{*}{Turn-level} & ICL & $\dfew$ & 0.0\sd{0.0} & 98.0\sd{0.8} & 0.3\sd{0.5} & 0.0\sd{0.0} & 100.0\sd{0.0} \\
        &  &  & ICL & 0-shot & 0.0\sd{0.0} & 86.7\sd{1.7} & 0.0\sd{0.0} & 0.0\sd{0.0} & 100.0\sd{0.0} \\
        \cmidrule(lr){4-10}
        &  & \multirow{2}{*}{\alps} & ICL & $\dfew$ & 93.7\sd{0.5} & 67.0\sd{2.2} & 92.3\sd{3.7} & 99.3\sd{0.5} & 100.0\sd{0.0} \\
        &  &  & ICL & 0-shot & 60.7\sd{0.5} & 98.3\sd{1.2} & 59.3\sd{3.9} & 63.0\sd{4.3} & 100.0\sd{0.0} \\
        \cmidrule(l){3-10}
        & \multirow{10}{*}{Qwen3} & \multirow{5}{*}{Turn-level} & ICL & $\dfew$ & 0.7\sd{0.5} & 99.7\sd{0.5} & 20.3\sd{0.9} & 11.0\sd{3.3} & 100.0\sd{0.0} \\
        &  &  & ICL & 0-shot & 0.0\sd{0.0} & 31.3\sd{8.2} & 10.3\sd{2.1} & 0.0\sd{0.0} & 90.7\sd{2.9} \\
        &  &  & RAFT & TR$^{\dagger}$ & \textit{0.7}\sd{0.5} & 97.3\sd{3.1} & 76.3\sd{7.3} & 57.0\sd{13.5} & 100.0\sd{0.0} \\
        &  &  & SFT & TD$^{\dagger}$ & \textit{0.7}\sd{0.5} & 99.7\sd{0.5} & \textit{20.3}\sd{0.9} & \textit{11.0}\sd{3.3} & 100.0\sd{0.0} \\
        &  &  & SFT${\scriptstyle\rightarrow}$RAFT & TD${\scriptstyle\rightarrow}$TR$^{\dagger}$ & \textit{0.7}\sd{0.5} & 100.0\sd{0.0} & \textit{20.3}\sd{0.9} & \textit{11.0}\sd{3.3} & 100.0\sd{0.0} \\
        \cmidrule(lr){4-10}
        &  & \multirow{5}{*}{\alps} & ICL & $\dfew$ & 10.3\sd{2.6} & 59.3\sd{1.7} & 32.0\sd{2.9} & 9.3\sd{1.7} & 99.0\sd{0.8} \\
        &  &  & ICL & 0-shot & 11.3\sd{2.6} & 97.0\sd{1.4} & 3.3\sd{1.2} & 36.0\sd{1.6} & 95.7\sd{1.7} \\
        &  &  & RAFT & TR & 47.7\sd{12.3} & 89.0\sd{10.4} & 55.7\sd{6.0} & 54.0\sd{18.5} & 99.7\sd{0.5} \\
        &  &  & SFT & TD & 86.3\sd{6.1} & 98.7\sd{1.9} & 96.3\sd{2.6} & 99.0\sd{0.8} & 100.0\sd{0.0} \\
        &  &  & SFT${\scriptstyle\rightarrow}$RAFT & TD${\scriptstyle\rightarrow}$TR & 90.7\sd{3.9} & 98.7\sd{1.2} & 95.3\sd{3.4} & 99.7\sd{0.5} & 100.0\sd{0.0} \\
        \bottomrule[1.2pt]
    \end{tabular*}
    }
    \captionsetup{skip=3pt}
    \caption{\textbf{Exp2: Data-scarce setting, per profile (\pharmasim{}).}
Simulation quality per behavior profile, as percentages
(mean$_{\pm\mathrm{std}}$ over three seeds, 100 trajectories per profile
per seed). The ICL rows report the $\dfew$
prompt used in the main table and a \emph{0-shot} variant without the
three example trajectories. 
$^{\dagger}$~Turn-level RAFT produced no trajectories for profile~1, and
turn-level SFT and SFT$\rightarrow$RAFT none for profiles~1, 3 and~4, in
any replica; those cells (\textit{italic}) are filled per replica with
the matched prompted turn-level baseline.}
\label{tab:per-profile-pharmasim-fs}
\end{table*}

\begin{table*}[t]
    \centering
    \footnotesize
    \setlength{\tabcolsep}{3pt}
    \renewcommand{\arraystretch}{0.85}
    \setlength{\aboverulesep}{0.05ex}
    \setlength{\belowrulesep}{0.15ex}
    {\footnotesize
    \begin{tabular*}{\textwidth}{@{\extracolsep{\fill}} c l l l l cccccc @{}}
        \toprule[1.2pt]
        & Model & Architecture & Adaptation & Signal Source & P1 & P2 & P3 & P4 & P5 & P6 \\
        \midrule[1pt]
        \multicolumn{11}{@{}l}{\textit{Clipped Density$\uparrow$}} \\
        \multirow{14}{*}{\rotatebox[origin=c]{90}{\textit{\tuglet{}}}}
        & \multirow{4}{*}{GPT-OSS} & \multirow{2}{*}{Turn-level} & ICL & $\dfew$ & 2.4\sd{0.5} & 26.6\sd{4.9} & 64.8\sd{2.5} & 70.8\sd{12.3} & 6.4\sd{1.5} & 17.0\sd{19.1} \\
        &  &  & ICL & 0-shot & 3.4\sd{2.0} & 26.6\sd{3.6} & 68.0\sd{1.4} & 75.8\sd{15.0} & 19.0\sd{10.3} & 3.4\sd{3.3} \\
        \cmidrule(lr){4-11}
        &  & \multirow{2}{*}{\alps} & ICL & $\dfew$ & 33.2\sd{3.4} & 38.4\sd{5.5} & 97.3\sd{0.3} & 100.0\sd{0.0} & 84.5\sd{3.8} & 0.0\sd{0.0} \\
        &  &  & ICL & 0-shot & 23.8\sd{4.4} & 34.5\sd{5.4} & 89.4\sd{3.0} & 18.3\sd{5.9} & 90.5\sd{4.4} & 0.0\sd{0.0} \\
        \cmidrule(l){3-11}
        & \multirow{10}{*}{Qwen3} & \multirow{5}{*}{Turn-level} & ICL & $\dfew$ & 9.3\sd{4.1} & 45.1\sd{1.6} & 15.5\sd{0.7} & 0.0\sd{0.0} & 0.0\sd{0.0} & 8.1\sd{3.0} \\
        &  &  & ICL & 0-shot & 1.4\sd{0.3} & 19.3\sd{1.1} & 10.2\sd{2.4} & 32.5\sd{5.4} & 17.1\sd{3.6} & 100.0\sd{0.0} \\
        &  &  & RAFT & TR & 0.6\sd{0.8} & 28.4\sd{21.6} & 36.7\sd{7.4} & 47.5\sd{10.6} & 0.0\sd{0.0} & 12.3\sd{3.6} \\
        &  &  & SFT & TD & 3.2\sd{3.0} & 26.7\sd{11.2} & 62.5\sd{10.9} & 0.0\sd{0.0} & 34.3\sd{5.2} & 0.0\sd{0.0} \\
        &  &  & SFT${\scriptstyle\rightarrow}$RAFT & TD${\scriptstyle\rightarrow}$TR & 3.8\sd{3.2} & 16.5\sd{7.6} & 59.8\sd{7.0} & 0.0\sd{0.0} & 0.0\sd{0.0} & 22.0\sd{8.7} \\
        \cmidrule(lr){4-11}
        &  & \multirow{5}{*}{\alps} & ICL & $\dfew$ & 12.7\sd{1.0} & 37.8\sd{8.6} & 50.2\sd{6.6} & 14.2\sd{2.4} & 12.9\sd{3.8} & 1.0\sd{1.5} \\
        &  &  & ICL & 0-shot & 3.2\sd{1.6} & 22.7\sd{0.6} & 54.5\sd{1.9} & 0.8\sd{1.2} & 6.9\sd{1.8} & 1.3\sd{1.3} \\
        &  &  & RAFT & TR & 26.7\sd{5.5} & 60.7\sd{24.1} & 78.0\sd{12.4} & 21.7\sd{1.2} & 19.0\sd{4.7} & 1.0\sd{1.5} \\
        &  &  & SFT & TD & 40.2\sd{8.0} & 31.5\sd{6.1} & 95.1\sd{3.9} & 100.0\sd{0.0} & 76.0\sd{17.4} & 2.4\sd{1.9} \\
        &  &  & SFT${\scriptstyle\rightarrow}$RAFT & TD${\scriptstyle\rightarrow}$TR & 43.5\sd{12.8} & 29.5\sd{13.6} & 99.4\sd{0.8} & 100.0\sd{0.0} & 67.4\sd{37.4} & 2.9\sd{2.4} \\
        \specialrule{1.2pt}{0.2ex}{0.2ex}
        \multicolumn{11}{@{}l}{\textit{Clipped Coverage$\uparrow$}} \\
        \multirow{14}{*}{\rotatebox[origin=c]{90}{\textit{\tuglet{}}}}
        & \multirow{4}{*}{GPT-OSS} & \multirow{2}{*}{Turn-level} & ICL & $\dfew$ & 3.0\sd{0.0} & 8.3\sd{1.2} & 6.6\sd{0.5} & 12.9\sd{4.2} & 3.0\sd{0.0} & 36.6\sd{44.8} \\
        &  &  & ICL & 0-shot & 4.3\sd{0.9} & 8.3\sd{0.5} & 6.3\sd{0.5} & 40.9\sd{41.9} & 4.6\sd{1.2} & 6.6\sd{6.2} \\
        \cmidrule(lr){4-11}
        &  & \multirow{2}{*}{\alps} & ICL & $\dfew$ & 17.2\sd{2.5} & 15.5\sd{3.4} & 5.6\sd{0.5} & 43.9\sd{39.7} & 14.2\sd{1.2} & 3.0\sd{0.0} \\
        &  &  & ICL & 0-shot & 9.2\sd{0.5} & 17.2\sd{3.3} & 5.6\sd{0.5} & 4.3\sd{1.9} & 18.2\sd{4.1} & 0.0\sd{0.0} \\
        \cmidrule(l){3-11}
        & \multirow{10}{*}{Qwen3} & \multirow{5}{*}{Turn-level} & ICL & $\dfew$ & 9.6\sd{1.9} & 17.2\sd{1.9} & 8.3\sd{1.2} & 0.0\sd{0.0} & 2.0\sd{1.4} & 6.3\sd{0.9} \\
        &  &  & ICL & 0-shot & 4.6\sd{2.3} & 13.9\sd{1.6} & 12.9\sd{4.0} & 100.0\sd{0.0} & 100.0\sd{0.0} & 100.0\sd{0.0} \\
        &  &  & RAFT & TR & 2.3\sd{1.7} & 11.6\sd{6.1} & 27.1\sd{22.3} & 100.0\sd{0.0} & 2.0\sd{1.4} & 29.4\sd{10.3} \\
        &  &  & SFT & TD & 5.0\sd{2.8} & 8.3\sd{1.0} & 5.3\sd{0.5} & 1.3\sd{0.9} & 3.0\sd{0.0} & 0.0\sd{0.0} \\
        &  &  & SFT${\scriptstyle\rightarrow}$RAFT & TD${\scriptstyle\rightarrow}$TR & 5.0\sd{1.6} & 7.3\sd{2.5} & 5.0\sd{0.8} & 0.7\sd{0.9} & 2.0\sd{1.4} & 16.8\sd{7.0} \\
        \cmidrule(lr){4-11}
        &  & \multirow{5}{*}{\alps} & ICL & $\dfew$ & 6.9\sd{0.8} & 6.9\sd{1.4} & 10.2\sd{1.7} & 15.8\sd{0.0} & 3.0\sd{0.0} & 1.7\sd{2.3} \\
        &  &  & ICL & 0-shot & 4.3\sd{1.2} & 3.6\sd{0.5} & 10.2\sd{1.7} & 0.7\sd{0.9} & 3.0\sd{0.0} & 2.3\sd{1.7} \\
        &  &  & RAFT & TR & 8.3\sd{1.7} & 6.9\sd{0.8} & 11.2\sd{1.9} & 15.8\sd{0.0} & 36.6\sd{44.8} & 1.7\sd{2.3} \\
        &  &  & SFT & TD & 21.5\sd{6.9} & 13.2\sd{1.7} & 5.0\sd{0.0} & 71.9\sd{39.7} & 100.0\sd{0.0} & 3.3\sd{0.5} \\
        &  &  & SFT${\scriptstyle\rightarrow}$RAFT & TD${\scriptstyle\rightarrow}$TR & 18.5\sd{6.5} & 9.9\sd{1.6} & 5.0\sd{0.0} & 71.9\sd{39.7} & 69.6\sd{42.9} & 3.3\sd{0.5} \\
        \specialrule{1.2pt}{0.2ex}{0.2ex}
        \multicolumn{11}{@{}l}{\textit{Judge pass rate$\uparrow$}} \\
        \multirow{14}{*}{\rotatebox[origin=c]{90}{\textit{\tuglet{}}}}
        & \multirow{4}{*}{GPT-OSS} & \multirow{2}{*}{Turn-level} & ICL & $\dfew$ & 3.0\sd{1.6} & 10.3\sd{2.9} & 99.0\sd{0.8} & 10.7\sd{3.3} & 0.0\sd{0.0} & 0.0\sd{0.0} \\
        &  &  & ICL & 0-shot & 0.7\sd{0.9} & 8.7\sd{1.7} & 98.7\sd{0.5} & 8.7\sd{2.1} & 0.0\sd{0.0} & 0.3\sd{0.5} \\
        \cmidrule(lr){4-11}
        &  & \multirow{2}{*}{\alps} & ICL & $\dfew$ & 58.0\sd{0.8} & 47.7\sd{3.4} & 100.0\sd{0.0} & 43.7\sd{1.9} & 3.0\sd{0.8} & 7.0\sd{1.4} \\
        &  &  & ICL & 0-shot & 51.3\sd{10.5} & 26.3\sd{5.4} & 100.0\sd{0.0} & 0.0\sd{0.0} & 3.0\sd{1.4} & 0.7\sd{0.5} \\
        \cmidrule(l){3-11}
        & \multirow{10}{*}{Qwen3} & \multirow{5}{*}{Turn-level} & ICL & $\dfew$ & 7.7\sd{2.1} & 45.0\sd{3.3} & 47.7\sd{10.1} & 0.0\sd{0.0} & 0.0\sd{0.0} & 5.3\sd{2.9} \\
        &  &  & ICL & 0-shot & 0.3\sd{0.5} & 22.7\sd{1.7} & 25.0\sd{5.4} & 0.7\sd{0.9} & 0.0\sd{0.0} & 0.7\sd{0.9} \\
        &  &  & RAFT & TR & 0.3\sd{0.5} & 22.0\sd{14.4} & 62.7\sd{24.1} & 0.0\sd{0.0} & 0.0\sd{0.0} & 0.0\sd{0.0} \\
        &  &  & SFT & TD & 4.0\sd{3.3} & 33.5\sd{12.9} & 97.3\sd{2.1} & 61.2\sd{6.8} & 0.0\sd{0.0} & 0.3\sd{0.5} \\
        &  &  & SFT${\scriptstyle\rightarrow}$RAFT & TD${\scriptstyle\rightarrow}$TR & 2.0\sd{0.0} & 21.7\sd{14.8} & 99.7\sd{0.5} & 70.7\sd{12.1} & 0.0\sd{0.0} & 0.0\sd{0.0} \\
        \cmidrule(lr){4-11}
        &  & \multirow{5}{*}{\alps} & ICL & $\dfew$ & 29.0\sd{2.4} & 4.3\sd{2.4} & 51.7\sd{4.9} & 8.3\sd{0.9} & 0.0\sd{0.0} & 0.0\sd{0.0} \\
        &  &  & ICL & 0-shot & 15.7\sd{1.7} & 2.3\sd{0.5} & 58.7\sd{1.2} & 0.7\sd{0.5} & 0.3\sd{0.5} & 0.0\sd{0.0} \\
        &  &  & RAFT & TR & 39.0\sd{3.6} & 6.0\sd{1.4} & 75.7\sd{3.9} & 14.3\sd{5.9} & 2.3\sd{2.6} & 0.0\sd{0.0} \\
        &  &  & SFT & TD & 68.7\sd{9.5} & 37.7\sd{8.3} & 99.7\sd{0.5} & 48.3\sd{1.2} & 16.3\sd{4.6} & 25.0\sd{14.2} \\
        &  &  & SFT${\scriptstyle\rightarrow}$RAFT & TD${\scriptstyle\rightarrow}$TR & 73.3\sd{9.0} & 38.0\sd{17.0} & 100.0\sd{0.0} & 42.0\sd{2.2} & 12.7\sd{4.2} & 40.7\sd{10.1} \\
        \bottomrule[1.2pt]
    \end{tabular*}
    }
    \captionsetup{skip=3pt}
    \caption{\textbf{Exp2: Data-scarce setting, per profile (\tuglet{}).}
Simulation quality per behavior profile, as percentages
(mean$_{\pm\mathrm{std}}$ over three seeds, 100 trajectories per profile per seed). Rows follow Table~\ref{tab:results-fewshot}, and each row averaged over the six profiles reproduces the corresponding \emph{mean} column there. The ICL rows report the $\dfew$ prompt used in the main table and a \emph{0-shot} variant with neither the three example trajectories nor numeric profile descriptions.}
    \label{tab:per-profile-tuglet-fs}
\end{table*}
\begin{table*}[t]
\centering
\footnotesize
\setlength{\tabcolsep}{3pt}
\renewcommand{\arraystretch}{0.95}
\begin{tabular*}{\textwidth}{@{\extracolsep{\fill}} l l cc cc cc cc @{}}
\toprule[1.2pt]
Model & Architecture & Numeric & Examples & \multicolumn{2}{c}{Clipped Density$\uparrow$} & \multicolumn{2}{c}{Clipped Coverage$\uparrow$} & \multicolumn{2}{c}{Judge pass rate$\uparrow$} \\
\cmidrule(lr){5-6} \cmidrule(lr){7-8} \cmidrule(l){9-10}
 & & & & mean & worst & mean & worst & mean & worst \\
\midrule[1pt]
\multirow{4}{*}{GPT-OSS} & \multirow{4}{*}{Turn-level} & -- & -- & 26.3\sd{0.9} & 0.0\sd{0.0} & 4.0\sd{1.3} & 0.0\sd{0.0} & 37.3\sd{0.3} & 0.0\sd{0.0} \\
 &  & $\checkmark$ & -- & 26.6\sd{0.6} & 0.0\sd{0.0} & 3.2\sd{1.2} & 0.0\sd{0.0} & 37.7\sd{0.3} & 0.0\sd{0.0} \\
 &  & -- & $\checkmark$ & 22.9\sd{0.2} & 0.0\sd{0.0} & 1.9\sd{0.2} & 0.0\sd{0.0} & 39.7\sd{0.1} & 0.0\sd{0.0} \\
 &  & $\checkmark$ & $\checkmark$ & 22.5\sd{0.3} & 0.0\sd{0.0} & 2.2\sd{0.7} & 0.0\sd{0.0} & 39.6\sd{0.2} & 0.0\sd{0.0} \\
\cmidrule(lr){2-10}
 & \multirow{4}{*}{\alps{}} & -- & -- & 75.2\sd{1.7} & 16.9\sd{6.8} & 6.7\sd{0.9} & 3.0\sd{0.0} & 76.3\sd{1.1} & 58.0\sd{2.9} \\
 &  & $\checkmark$ & -- & 85.5\sd{0.5} & 59.9\sd{0.4} & 10.6\sd{0.1} & 3.0\sd{0.0} & 78.8\sd{2.3} & 47.7\sd{3.3} \\
 &  & -- & $\checkmark$ & 83.7\sd{0.7} & 18.4\sd{3.3} & 25.9\sd{0.2} & 3.0\sd{0.0} & 90.5\sd{0.7} & 67.0\sd{2.2} \\
 &  & $\checkmark$ & $\checkmark$ & 76.0\sd{2.3} & 47.9\sd{4.5} & 9.7\sd{0.3} & 3.0\sd{0.0} & 95.6\sd{0.8} & 86.9\sd{1.5} \\
\midrule[1pt]
\multirow{4}{*}{Qwen3} & \multirow{4}{*}{Turn-level} & -- & -- & 23.9\sd{1.2} & 0.0\sd{0.0} & 16.7\sd{8.2} & 0.0\sd{0.0} & 26.5\sd{0.8} & 0.0\sd{0.0} \\
 &  & $\checkmark$ & -- & 25.2\sd{0.9} & 0.0\sd{0.0} & 17.4\sd{16.0} & 0.0\sd{0.0} & 27.7\sd{0.6} & 0.0\sd{0.0} \\
 &  & -- & $\checkmark$ & 36.8\sd{0.4} & 0.0\sd{0.0} & 9.5\sd{8.7} & 0.7\sd{0.9} & 46.3\sd{0.8} & 0.7\sd{0.5} \\
 &  & $\checkmark$ & $\checkmark$ & 37.4\sd{0.6} & 0.0\sd{0.0} & 4.2\sd{1.3} & 2.0\sd{1.4} & 46.6\sd{0.3} & 1.0\sd{0.8} \\
\cmidrule(lr){2-10}
 & \multirow{4}{*}{\alps{}} & -- & -- & 38.9\sd{2.2} & 0.0\sd{0.0} & 3.8\sd{0.2} & 2.3\sd{0.5} & 48.7\sd{1.4} & 3.3\sd{1.2} \\
 &  & $\checkmark$ & -- & 36.3\sd{2.7} & 3.7\sd{1.0} & 5.9\sd{0.5} & 3.0\sd{0.0} & 67.3\sd{0.7} & 42.0\sd{1.6} \\
 &  & -- & $\checkmark$ & 37.4\sd{1.7} & 3.2\sd{0.9} & 5.1\sd{0.7} & 3.0\sd{0.0} & 42.0\sd{0.2} & 8.0\sd{0.8} \\
 &  & $\checkmark$ & $\checkmark$ & 30.7\sd{0.8} & 5.2\sd{2.6} & 8.3\sd{1.4} & 3.0\sd{0.0} & 41.9\sd{1.2} & 8.7\sd{3.4} \\
\bottomrule[1.2pt]
\end{tabular*}
\caption{\textbf{Prompt-variant comparison (\pharmasim{}).} The two prompt
factors are crossed: \emph{Numeric} indicates numeric profile descriptions
(explicit count ranges per behavior), \emph{Examples} indicates three worked
example trajectories in context. Values are percentages
(mean$_{\pm\mathrm{std}}$ over three seeds); \emph{mean} averages across the
five behavior profiles and \emph{worst} reports the lowest-scoring.}
\label{tab:prompt-variants-pharmasim}
\end{table*}

\begin{table*}[t]
\centering
\footnotesize
\setlength{\tabcolsep}{3pt}
\renewcommand{\arraystretch}{0.95}
\begin{tabular*}{\textwidth}{@{\extracolsep{\fill}} l l cc cc cc cc @{}}
\toprule[1.2pt]
Model & Architecture & Numeric & Examples & \multicolumn{2}{c}{Clipped Density$\uparrow$} & \multicolumn{2}{c}{Clipped Coverage$\uparrow$} & \multicolumn{2}{c}{Judge pass rate$\uparrow$} \\
\cmidrule(lr){5-6} \cmidrule(lr){7-8} \cmidrule(l){9-10}
 & & & & mean & worst & mean & worst & mean & worst \\
\midrule[1pt]
\multirow{8}{*}{GPT-OSS} & \multirow{4}{*}{Turn-level} & -- & -- & 32.7\sd{3.6} & 1.6\sd{1.8} & 11.8\sd{6.6} & 2.6\sd{2.0} & 19.5\sd{0.3} & 0.0\sd{0.0} \\
 &  & $\checkmark$ & -- & 24.4\sd{1.1} & 0.0\sd{0.0} & 7.0\sd{0.3} & 1.0\sd{1.4} & 21.3\sd{0.1} & 0.0\sd{0.0} \\
 &  & -- & $\checkmark$ & 31.3\sd{5.4} & 2.4\sd{0.5} & 11.7\sd{6.7} & 3.0\sd{0.0} & 20.5\sd{1.2} & 0.0\sd{0.0} \\
 &  & $\checkmark$ & $\checkmark$ & 48.7\sd{4.0} & 2.0\sd{2.0} & 48.8\sd{7.8} & 3.0\sd{0.0} & 20.6\sd{0.3} & 0.0\sd{0.0} \\
\cmidrule(lr){2-10}
 & \multirow{4}{*}{\alps{}} & -- & -- & 42.7\sd{0.7} & 0.0\sd{0.0} & 9.1\sd{1.2} & 0.0\sd{0.0} & 30.2\sd{1.8} & 0.0\sd{0.0} \\
 &  & $\checkmark$ & -- & 77.1\sd{2.2} & 21.0\sd{2.4} & 31.1\sd{0.9} & 9.2\sd{0.9} & 62.7\sd{0.5} & 7.0\sd{1.6} \\
 &  & -- & $\checkmark$ & 58.9\sd{1.5} & 0.0\sd{0.0} & 16.6\sd{6.9} & 3.0\sd{0.0} & 43.2\sd{1.0} & 3.0\sd{0.8} \\
 &  & $\checkmark$ & $\checkmark$ & 83.9\sd{1.1} & 42.4\sd{6.5} & 34.9\sd{6.7} & 9.9\sd{0.0} & 69.7\sd{0.8} & 11.0\sd{1.6} \\
\midrule[1pt]
\multirow{8}{*}{Qwen3} & \multirow{4}{*}{Turn-level} & -- & -- & 30.1\sd{2.1} & 1.4\sd{0.3} & 55.2\sd{0.8} & 4.6\sd{2.3} & 8.2\sd{1.2} & 0.0\sd{0.0} \\
 &  & $\checkmark$ & -- & 5.6\sd{0.3} & 0.0\sd{0.0} & 4.7\sd{0.1} & 0.0\sd{0.0} & 11.3\sd{1.5} & 0.0\sd{0.0} \\
 &  & -- & $\checkmark$ & 13.0\sd{0.7} & 0.0\sd{0.0} & 7.2\sd{0.4} & 0.0\sd{0.0} & 17.6\sd{1.5} & 0.0\sd{0.0} \\
 &  & $\checkmark$ & $\checkmark$ & 48.4\sd{2.3} & 0.4\sd{0.6} & 69.1\sd{0.2} & 1.0\sd{1.4} & 6.9\sd{0.9} & 0.0\sd{0.0} \\
\cmidrule(lr){2-10}
 & \multirow{4}{*}{\alps{}} & -- & -- & 14.9\sd{0.5} & 0.3\sd{0.4} & 4.0\sd{0.5} & 0.7\sd{0.9} & 12.9\sd{0.3} & 0.0\sd{0.0} \\
 &  & $\checkmark$ & -- & 25.5\sd{0.7} & 7.5\sd{2.2} & 8.5\sd{0.6} & 5.0\sd{0.8} & 19.1\sd{1.3} & 7.0\sd{2.9} \\
 &  & -- & $\checkmark$ & 21.5\sd{1.7} & 1.0\sd{1.5} & 7.4\sd{0.0} & 1.0\sd{1.4} & 15.6\sd{0.7} & 0.0\sd{0.0} \\
 &  & $\checkmark$ & $\checkmark$ & 15.7\sd{0.2} & 4.2\sd{2.4} & 13.5\sd{5.4} & 3.0\sd{0.0} & 15.4\sd{0.4} & 3.3\sd{1.2} \\
\bottomrule[1.2pt]
\end{tabular*}
\caption{\textbf{Prompt-variant comparison (\tuglet{}).} The two prompt
factors are crossed: \emph{Numeric} indicates numeric profile descriptions
(explicit count ranges per behavior), \emph{Examples} indicates three worked
example trajectories in context. Values are percentages
(mean$_{\pm\mathrm{std}}$ over three seeds); \emph{mean} averages across the six
behavior profiles and \emph{worst} reports the lowest-scoring.}
\label{tab:prompt-variants-tuglet}
\end{table*}

\subsection{Profile Drift over the Trajectory}
\label{app:drift}

We assess whether turn-level generation drifts away from the assigned profile over the course of a trajectory, and whether the segment planner prevents this drift. Each trajectory is scored at ten positions: the fraction of its own actions completed is binned into deciles, and the behavioral features of the corresponding prefix are scored by the in-profile classifier against the real basis. We report the vote share assigned to the target profile, set to zero when the within-profile outlier gate fires, so a prefix that leaves the real-data manifold counts as unaligned rather than as a weak match. As a reference, the same procedure is applied to the real test students, scored leave-one-out so that a trajectory is never its own nearest neighbor.
 
\begin{figure}
\centering
\includegraphics[width=\linewidth]{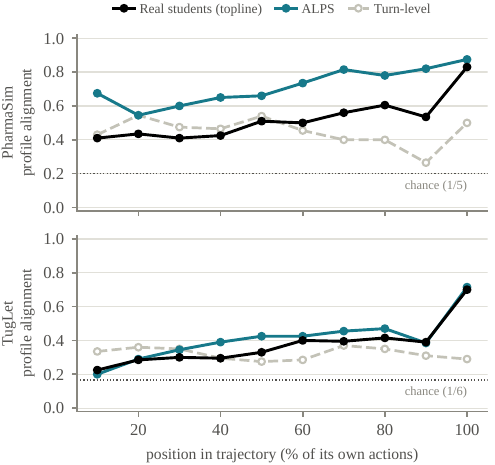}
\caption{\textbf{Profile alignment along the trajectory
(GPT-OSS-120B).} Target-profile vote share of the in-profile classifier,
outlier-masked, against position in the trajectory, for \pharmasim{} (top,
chance $=1/5$) and \tuglet{} (bottom, chance $=1/6$). Real students (black)
align increasingly with their profile as the trajectory unfolds. Turn-level
generation (gray, dashed) stagnates or decays toward chance, whereas \alps{}
(teal) tracks the real trend.}
\label{fig:drift}
\end{figure}

Figure~\ref{fig:drift} shows the results for GPT-OSS-120B ICL in both
environments. Real students' alignment increases with trajectory position:
profiles are expressed most clearly late in the interaction. The turn-level
student does not follow this pattern; its alignment stagnates or decays toward
chance as the trajectory grows long, ending well below the real
students. \alps{} instead tracks the rising alignment of the real
students, closing most of the gap by the final position in
\pharmasim{} and matching the real curve in \tuglet{}. The two architectures
therefore separate progressively, which is the pattern the segment planner is meant to produce: a profile carried by a few segment-level decisions survives a long trajectory, whereas one re-established at every action does not. Because the gap opens only late, it is invisible to any evaluation that scores short interactions or single turns.

\subsection{Component Ablation}
\label{app:components-abl}
 
To locate the source of the fine-tuning gains, we ablate the two
adapters of \alps{} (SFT$\rightarrow$RAFT) separately, running the
removed component prompted while the other keeps its adapter
(Table~\ref{tab:ablation}).
 
The planner adapter carries profile alignment. Removing it returns
\pharmasim{} to roughly prompted performance and collapses \tuglet{} on all
three metrics. The larger \tuglet{} drop is consistent with co-adaptation: the
adapted generator is trained on plans produced by the adapted planner, and the
prompted planner supplies plans it has not seen. The generator adapter instead
broadens coverage within a profile. Removing it lowers \coverage{} and
worst-profile \density{} in both environments, while mean \density{} changes
little in either direction. The two components are therefore complementary
rather than redundant: the planner decides which behavior is generated, and the
generator determines how much of the real within-profile variation that
behavior spans.
How far that complementarity goes differs by environment. On
\pharmasim{} the planner-free model falls back to its prompted level, so the
planner supplies the adaptation gain and the generator refines it; on \tuglet{}
it falls below the prompted level on every metric, so there the two adapters are
co-dependent rather than merely complementary.

\begin{table*}[t]
\centering
\small
\setlength{\tabcolsep}{3pt}
\begin{tabular}{@{}lcccccc@{}}
\toprule
& \multicolumn{2}{c}{Clipped Density} & \multicolumn{2}{c}{Clipped Coverage} & \multicolumn{2}{c}{Judge} \\
\cmidrule(lr){2-3} \cmidrule(lr){4-5} \cmidrule(l){6-7}
Configuration & mean & worst & mean & worst & mean & worst \\
\midrule
\multicolumn{7}{@{}l}{\textsc{PharmaSim}} \\
\alps{} (full)        & \textbf{69.7}$\pm$3.1 & \textbf{33.4}$\pm$5.8 & \textbf{58.1}$\pm$2.8 & \textbf{21.2}$\pm$5.0 & \textbf{75.8}$\pm$3.8 & \textbf{41.6}$\pm$20.4 \\
\;- generator adapter & 67.0$\pm$5.3 & 23.7$\pm$4.5 & 49.7$\pm$8.4 & 17.1$\pm$3.8 & 71.2$\pm$5.6 & 40.4$\pm$17.7 \\
\;- planner adapter   & 35.3$\pm$0.9 & 4.5$\pm$2.2 & 6.8$\pm$0.8  & 3.0$\pm$0.0 & 67.0$\pm$2.8 & 35.9$\pm$2.2 \\
\midrule
\multicolumn{7}{@{}l}{\textsc{\tuglet{}}} \\
\alps{} (full)        & 52.9$\pm$7.1 & \textbf{9.1}$\pm$4.1 & \textbf{46.4}$\pm$7.3 & 5.0$\pm$0.8 & 41.2$\pm$5.3 & 11.0$\pm$7.3 \\
\;- generator adapter & \textbf{56.8}$\pm$5.9 & 6.0$\pm$4.5 & 33.6$\pm$1.0 & \textbf{5.6}$\pm$1.2 & \textbf{45.5}$\pm$4.4 & \textbf{12.0}$\pm$2.5 \\
\;- planner adapter   & 5.3$\pm$1.9 & 0.7$\pm$0.5 & 9.9$\pm$8.7 & 2.0$\pm$1.4 & 7.7$\pm$3.2 & 2.0$\pm$0.8 \\
\bottomrule
\end{tabular}
\caption{Adapter ablation of \alps{} (SFT$\rightarrow$RAFT) in the
data-sufficient setting. ``- generator adapter'' keeps only the planner
adapter and ``- planner adapter'' only the generator adapter; the
removed component runs prompted. Percentages,
mean$_{\pm\text{std}}$ over 3 replicas; \emph{mean} and \emph{worst}
aggregate over profiles; \textbf{bold} marks the best value per
column within each environment.}
\label{tab:ablation}
\end{table*}

\subsection{ICL Prompt Content}
\label{app:icl}
\label{app:prompt-analysis}
 
We vary the two contents of the
profile prompt independently: the numeric profile descriptions, which state
count ranges per behavior, and the three worked example trajectories.
Tables~\ref{tab:prompt-variants-pharmasim} and~\ref{tab:prompt-variants-tuglet}
report the four combinations for both models and both architectures.
The grid answers three questions: which content helps which model,
whether the two metric families agree, and what no ICL variant fixes.
 
Both contents help the 120B model, and on \tuglet{} they compose:
numbers alone and examples alone each raise the \alps{} judge pass rate over the
bare prompt, and the two together give the best cell of that table in every
column. On \pharmasim{} they do not combine the same way. Adding
examples raises the judge pass rate whichever description accompanies them,
numeric descriptions alone give the highest \density{}, and examples without
numeric descriptions the highest \coverage{}, so each content is best on a
different metric and no prompt is best on all three.
 
For the 8B student the ordering inverts. Numeric descriptions
\emph{without} examples give the best \alps{} judge pass rate in both
environments, and adding the three examples lowers it while leaving \density{}
essentially unchanged. The pattern is consistent with context overload happening by adding examples rather
than with uninformative examples: the same examples that improve the larger
model degrade the smaller one, and they degrade it most on the judge, which is
the measure most sensitive to whether the planner still respects the profile.
 
The effect on the turn-level student runs the other way. GPT-OSS
turn-level is insensitive to the prompt on \pharmasim{}, whereas Qwen3
turn-level depends on the examples alone: they raise its \density{} in both data
settings while halving its \coverage{}, and numeric descriptions contribute
almost nothing. The two metrics move in opposite directions, so the examples
concentrate the turn-level population near typical real behavior at the cost of
spanning fewer real students. The per-profile view shows what the examples supply: for
Qwen3 turn-level the three-shot prompt is best on every \pharmasim{} profile,
whereas for Qwen3 \alps{} the best prompt differs by profile
(Table~\ref{tab:per-profile-pharmasim}). Examples give the unstructured
architecture the pacing information that the behavior plan in \alps{} otherwise provides.
 
 No ICL variant repairs the worst profile. Every turn-level configuration in
the \pharmasim{} grid scores zero worst-profile \density{} under every prompt,
and every turn-level configuration in the \tuglet{} grid scores zero
worst-profile judge under every prompt in both models, eight of eight cells;
\alps{} clears zero in five of its eight \tuglet{} cells.

\newcommand{\statsnote}[1]{%
Wilcoxon signed-rank over paired profile$\times$seed cells, exact null. Zero differences
are retained in $n$ (\texttt{zero\_method=pratt}), since ties arise from ceiling effects
where both arms saturate a metric on a profile. Tests are one-sided where the claim is
directional and two-sided for the RAFT contrasts and the reference comparison. $n$ gives
total pairs and, in parentheses, non-tied pairs. $\Delta$ is the mean paired difference in
percentage points with a percentile-bootstrap 95\%\ CI (#1 resamples). Holm correction is
applied within each (contrast block, metric) family. \textbf{Bold} marks
$p_{\text{Holm}}<.05$. A \textsuperscript{\ddag} on the $n$ column marks at least 40\%\ tied pairs, and a
\textsuperscript{\S} on $\Delta$ marks $|\Delta|<0.05$. Rows with either flag should be read from $\Delta$ and the non-tied count
rather than from $p$.}

\subsection{Statistical Tests}
\label{sec:stats-tests}

Tables~\ref{tab:stats-exp1}, \ref{tab:stats-exp1-tuglet}, \ref{tab:stats-exp2}
and~\ref{tab:stats-exp2-tuglet} report every test behind the significance statements in the
main text, one per environment and data setting.

\paragraph{Unit of analysis.} One test is one contrast $\times$ metric over paired
(profile, seed) cells, giving $n=15$ on \pharmasim{} and $n=18$ on \tuglet{}. Pairing
differences out profile difficulty and seed variation. We take the cell rather than the
profile as the unit because the three replicas are not repetitions of one fit. Each varies
the training seed, the training and validation split, and the generation and evaluation
seeds, so the spread across replicas is the run-to-run variation a practitioner would meet,
and aggregating it away would discard evidence the experiment was designed to collect.

\paragraph{Fallback cells.} For twelve (condition, profile) combinations the acceptance
filter admitted no candidate in any replica, so no adapter was trained and the deployed
system falls back to the matched prompted model in all three replicas. On \pharmasim{}
these are turn-level RAFT (P1, P4) in the data-sufficient setting and turn-level RAFT (P1),
SFT (P1, P3, P4) and SFT$\rightarrow$RAFT (P1, P3, P4) in the data-scarce setting. On
\tuglet{} they are turn-level RAFT (P5, P6) and \alps{}~(RAFT) on P6. We report these cells
as the method delivers them, in the aggregate columns of the main tables and in the tests
alike, so that a condition is scored on what it produces end to end rather than only on the
profiles for which it trained an adapter. Under this intention-to-treat protocol the
resulting zero differences are observations, so they are retained in $n$ (\texttt{pratt}).

\paragraph{Families.} Holm correction is applied within each (contrast block, metric)
family. The architecture contrasts form two blocks in both experiments, one under prompting
(GPT-OSS and Qwen3, two contrasts) and one at each adapted stage (SFT, RAFT and
SFT$\rightarrow$RAFT, three contrasts). Adaptation contrasts form a separate block per
experiment, and the prompt-robustness contrast forms its own. The subheadings of each table
give the membership, which is identical across experiments and environments.

\paragraph{Tied pairs.} Every row reports the number of non-tied pairs alongside $n$. Rows
in which at least 40\%\ of pairs are tied carry \textsuperscript{\ddag}, and rows with
$|\Delta|<0.05$ carry \textsuperscript{\S}. Rows with either flag are read together with
$\Delta$ and the non-tied count.

\paragraph{Markers in the main tables.} Every marker set in
Tables~\ref{tab:results-limited} and~\ref{tab:results-fewshot} corresponds to a significant
test here. Five further cells reach significance under the declared families:
\tuglet{} \alps{}~(RAFT) \coverage{} and \pharmasim{} \alps{}~(RAFT)
\density{} in the data-sufficient setting, and \tuglet{} \alps{}~(ICL) \density{},
\tuglet{} GPT-OSS \alps{} \coverage{} and \tuglet{} \alps{}~(SFT) judge in the data-scarce
setting, all with $.024 \le p_{\text{Holm}} \le .046$. The tests reported here give the
significance status of every cell.

\begin{table*}[t]
\centering
\footnotesize
\setlength{\tabcolsep}{4pt}
\renewcommand{\arraystretch}{0.95}
\begin{tabular*}{\textwidth}{@{\extracolsep{\fill}} l L{5.0cm} l r r c r r @{}}
\toprule
& Contrast & Metric & $n$ (inf.) & $\Delta$ & 95\%\ CI & $p$ & $p_{\text{Holm}}$ \\
\midrule
\multicolumn{8}{@{}l}{\textit{\pharmasim{} Exp1: architecture under prompting (2 contrasts)}} \\
& E1.1 GPT-OSS: \alps{} vs.\ Turn & \density{} & 15 (12) & $+53.5$ & $[+36.9, +69.5]$ & $.0004$ & \textbf{$<$.001} \\
&  & \coverage{} & 15 (14) & $+7.5$ & $[+4.2, +10.9]$ & $.0008$ & \textbf{.0015} \\
&  & judge & 15 (12) & $+56.0$ & $[+31.9, +78.8]$ & $.0004$ & \textbf{$<$.001} \\
& E1.2 Qwen3: \alps{} vs.\ Turn & \density{} & 15 (12) & $-6.6$ & $[-19.8, +4.7]$ & $.2622$ & $.2622$ \\
&  & \coverage{} & 15 (10) & $+4.0$ & $[+1.0, +6.9]$ & $.0677$ & $.0677$ \\
&  & judge & 15 (15) & $-4.7$ & $[-16.2, +6.7]$ & $.7556$ & $.7556$ \\
\addlinespace[3pt]
\multicolumn{8}{@{}l}{\textit{\pharmasim{} Exp1: adaptation, Qwen3 (8 contrasts)}} \\
& E1.3 \alps{} SFT vs.\ \alps{} ICL & \density{} & 15 (13) & $+34.2$ & $[+18.6, +50.2]$ & $.0013$ & \textbf{.0105} \\
&  & \coverage{} & 15 (15) & $+42.7$ & $[+26.0, +60.7]$ & $<.0001$ & \textbf{$<$.001} \\
&  & judge & 15 (15) & $+30.7$ & $[+17.4, +45.5]$ & $.0002$ & \textbf{.0015} \\
& E1.4 Turn SFT vs.\ Turn ICL & \density{} & 15 (14) & $+0.3$ & $[-19.0, +18.0]$ & $.3599$ & $1.0000$ \\
&  & \coverage{} & 15 (15) & $+38.7$ & $[+19.9, +59.3]$ & $<.0001$ & \textbf{$<$.001} \\
&  & judge & 15 (15) & $+6.6$ & $[-10.4, +22.9]$ & $.2271$ & $.6814$ \\
& E1.5 \alps{} SFT$\rightarrow$RAFT vs.\ \alps{} SFT & \density{} & 15 (14) & $+4.8$ & $[-0.2, +9.6]$ & $.1070$ & $.4280$ \\
&  & \coverage{} & 15 (10) & $+7.2$ & $[-0.7, +19.4]$ & $.6387$ & $1.0000$ \\
&  & judge & 15 (14) & $+3.1$ & $[+0.2, +6.8]$ & $.1205$ & $.6027$ \\
& E1.6a \alps{} RAFT vs.\ \alps{} ICL & \density{} & 15 (12) & $+5.2$ & $[-3.1, +15.4]$ & $.8469$ & $1.0000$ \\
&  & \coverage{} & 15 (15) & $-0.6$ & $[-3.7, +2.3]$ & $.9341$ & $1.0000$ \\
&  & judge & 15 (15) & $+26.6$ & $[+17.6, +34.8]$ & $.0003$ & \textbf{.0018} \\
& E1.6b Turn RAFT vs.\ Turn ICL & \density{} & 15 (7)\textsuperscript{\ddag} & $-9.1$ & $[-20.1, -0.6]$ & $.0020$ & \textbf{.0141} \\
&  & \coverage{} & 15 (14) & $+19.8$ & $[+2.2, +39.8]$ & $.5995$ & $1.0000$ \\
&  & judge & 15 (9)\textsuperscript{\ddag} & $-15.7$ & $[-27.6, -5.9]$ & $<.0001$ & \textbf{$<$.001} \\
& E1.7 \alps{} ICL vs.\ Turn SFT$\rightarrow$RAFT & \density{} & 15 (14) & $+1.4$ & $[-10.5, +14.5]$ & $.3808$ & $1.0000$ \\
&  & \coverage{} & 15 (15) & $-19.0$ & $[-37.3, -2.9]$ & $.9723$ & $1.0000$ \\
&  & judge & 15 (15) & $-2.3$ & $[-14.3, +10.3]$ & $.7193$ & $.7193$ \\
& E1.11 \alps{} SFT vs.\ \alps{} RAFT & \density{} & 15 (13) & $+29.0$ & $[+13.9, +43.7]$ & $.0051$ & \textbf{.0256} \\
&  & \coverage{} & 15 (15) & $+43.4$ & $[+24.4, +63.5]$ & $<.0001$ & \textbf{$<$.001} \\
&  & judge & 15 (15) & $+4.1$ & $[-8.9, +17.5]$ & $.3394$ & $.6814$ \\
& E1.12 \alps{} SFT$\rightarrow$RAFT vs.\ \alps{} RAFT & \density{} & 15 (14) & $+33.8$ & $[+18.1, +49.5]$ & $.0021$ & \textbf{.0141} \\
&  & \coverage{} & 15 (15) & $+50.5$ & $[+31.1, +70.2]$ & $<.0001$ & \textbf{$<$.001} \\
&  & judge & 15 (15) & $+7.2$ & $[-5.5, +20.3]$ & $.1514$ & $.6056$ \\
\addlinespace[3pt]
\multicolumn{8}{@{}l}{\textit{\pharmasim{} Exp1: architecture at each adapted stage, Qwen3 (3 contrasts)}} \\
& E1.8 \alps{} SFT vs.\ Turn SFT & \density{} & 15 (14) & $+27.2$ & $[+13.9, +41.3]$ & $.0006$ & \textbf{.0012} \\
&  & \coverage{} & 15 (11) & $+8.0$ & $[-10.1, +25.7]$ & $.0844$ & $.1688$ \\
&  & judge & 15 (15) & $+19.5$ & $[+5.4, +34.5]$ & $.0177$ & \textbf{.0177} \\
& E1.9 \alps{} SFT$\rightarrow$RAFT vs.\ Turn SFT$\rightarrow$RAFT & \density{} & 15 (15) & $+40.4$ & $[+26.2, +55.1]$ & $.0002$ & \textbf{$<$.001} \\
&  & \coverage{} & 15 (12) & $+30.9$ & $[+16.4, +47.5]$ & $.0004$ & \textbf{.0013} \\
&  & judge & 15 (15) & $+31.5$ & $[+14.3, +49.2]$ & $.0034$ & \textbf{.0067} \\
& E1.10 \alps{} RAFT vs.\ Turn RAFT & \density{} & 15 (12) & $+7.6$ & $[-4.3, +20.3]$ & $.1384$ & $.1384$ \\
&  & \coverage{} & 15 (15) & $-16.4$ & $[-38.9, +3.0]$ & $.6807$ & $.6807$ \\
&  & judge & 15 (15) & $+37.7$ & $[+28.6, +46.9]$ & $<.0001$ & \textbf{$<$.001} \\
\bottomrule
\end{tabular*}
\caption{\textbf{Exp1: data-sufficient test results, \pharmasim{}.} \statsnote{$2{\times}10^{4}$}}
\label{tab:stats-exp1}\label{tab:stats-rq12}
\end{table*}
\begin{table*}[t]
\centering
\footnotesize
\setlength{\tabcolsep}{4pt}
\renewcommand{\arraystretch}{0.95}
\begin{tabular*}{\textwidth}{@{\extracolsep{\fill}} l L{5.0cm} l r r c r r @{}}
\toprule
& Contrast & Metric & $n$ (inf.) & $\Delta$ & 95\%\ CI & $p$ & $p_{\text{Holm}}$ \\
\midrule
\multicolumn{8}{@{}l}{\textit{\tuglet{} Exp1: architecture under prompting (2 contrasts)}} \\
& E1.1 GPT-OSS: \alps{} vs.\ Turn & \density{} & 18 (18) & $+35.2$ & $[+18.6, +50.9]$ & $.0006$ & \textbf{.0013} \\
&  & \coverage{} & 18 (15) & $-13.9$ & $[-36.6, +9.0]$ & $.4325$ & $.8650$ \\
&  & judge & 18 (18) & $+49.1$ & $[+35.3, +62.1]$ & $<.0001$ & \textbf{$<$.001} \\
& E1.2 Qwen3: \alps{} vs.\ Turn & \density{} & 18 (18) & $-32.7$ & $[-51.6, -14.6]$ & $.9962$ & $.9962$ \\
&  & \coverage{} & 18 (17) & $-55.6$ & $[-76.1, -34.5]$ & $.9997$ & $.9997$ \\
&  & judge & 18 (18) & $+8.6$ & $[-0.2, +17.5]$ & $.0770$ & $.0770$ \\
\addlinespace[3pt]
\multicolumn{8}{@{}l}{\textit{\tuglet{} Exp1: adaptation, Qwen3 (8 contrasts)}} \\
& E1.3 \alps{} SFT vs.\ \alps{} ICL & \density{} & 18 (18) & $+30.6$ & $[+19.7, +42.0]$ & $<.0001$ & \textbf{$<$.001} \\
&  & \coverage{} & 18 (18) & $+34.2$ & $[+13.4, +53.5]$ & $.0010$ & \textbf{.0048} \\
&  & judge & 18 (18) & $+32.1$ & $[+22.1, +41.8]$ & $<.0001$ & \textbf{$<$.001} \\
& E1.4 Turn SFT vs.\ Turn ICL & \density{} & 18 (16) & $-19.5$ & $[-44.4, +4.1]$ & $.7789$ & $1.0000$ \\
&  & \coverage{} & 18 (17) & $-58.1$ & $[-77.4, -37.7]$ & $.9996$ & $1.0000$ \\
&  & judge & 18 (12) & $+12.3$ & $[+5.2, +20.9]$ & $.0017$ & \textbf{.0118} \\
& E1.5 \alps{} SFT$\rightarrow$RAFT vs.\ \alps{} SFT & \density{} & 18 (17) & $+6.6$ & $[-3.0, +16.5]$ & $.2837$ & $.8512$ \\
&  & \coverage{} & 18 (15) & $-1.3$ & $[-16.3, +14.0]$ & $.3038$ & $1.0000$ \\
&  & judge & 18 (18) & $-6.3$ & $[-13.3, +0.7]$ & $.0987$ & $.3950$ \\
& E1.6a \alps{} RAFT vs.\ \alps{} ICL & \density{} & 18 (17) & $+19.1$ & $[+4.3, +34.0]$ & $.0342$ & $.1712$ \\
&  & \coverage{} & 18 (17) & $-6.1$ & $[-18.8, +2.6]$ & $.8317$ & $1.0000$ \\
&  & judge & 18 (18) & $+10.4$ & $[-0.3, +22.8]$ & $.1187$ & $.3950$ \\
& E1.6b Turn RAFT vs.\ Turn ICL & \density{} & 18 (18) & $-42.8$ & $[-60.5, -25.8]$ & $.0003$ & \textbf{.0018} \\
&  & \coverage{} & 18 (18) & $-63.8$ & $[-83.3, -43.2]$ & $.0001$ & \textbf{$<$.001} \\
&  & judge & 18 (9)\textsuperscript{\ddag} & $-1.7$ & $[-4.8, +0.7]$ & $.3465$ & $.6931$ \\
& E1.11 \alps{} SFT vs.\ \alps{} RAFT & \density{} & 18 (18) & $+11.6$ & $[+0.5, +21.5]$ & $.0665$ & $.2661$ \\
&  & \coverage{} & 18 (18) & $+40.2$ & $[+27.5, +54.1]$ & $<.0001$ & \textbf{$<$.001} \\
&  & judge & 18 (18) & $+21.7$ & $[+10.6, +32.1]$ & $.0028$ & \textbf{.0168} \\
& E1.12 \alps{} SFT$\rightarrow$RAFT vs.\ \alps{} RAFT & \density{} & 18 (18) & $+18.1$ & $[+4.5, +32.6]$ & $.0237$ & $.1420$ \\
&  & \coverage{} & 18 (18) & $+38.9$ & $[+23.2, +55.7]$ & $<.0001$ & \textbf{$<$.001} \\
&  & judge & 18 (18) & $+15.3$ & $[+2.4, +27.6]$ & $.0342$ & $.1712$ \\
& E1.7 \alps{} ICL vs.\ Turn SFT$\rightarrow$RAFT & \density{} & 18 (18) & $-16.1$ & $[-30.9, -2.4]$ & $.9506$ & $1.0000$ \\
&  & \coverage{} & 18 (17) & $+2.7$ & $[-6.5, +16.1]$ & $.8154$ & $1.0000$ \\
&  & judge & 18 (18) & $-8.3$ & $[-27.3, +8.4]$ & $.5000$ & $.6931$ \\
\addlinespace[3pt]
\multicolumn{8}{@{}l}{\textit{\tuglet{} Exp1: architecture at each adapted stage, Qwen3 (3 contrasts)}} \\
& E1.8 \alps{} SFT vs.\ Turn SFT & \density{} & 18 (17) & $+17.4$ & $[+4.9, +30.7]$ & $.0134$ & \textbf{.0134} \\
&  & \coverage{} & 18 (18) & $+36.7$ & $[+21.7, +52.9]$ & $<.0001$ & \textbf{$<$.001} \\
&  & judge & 18 (18) & $+28.3$ & $[+16.6, +39.3]$ & $.0003$ & \textbf{$<$.001} \\
& E1.9 \alps{} SFT$\rightarrow$RAFT vs.\ Turn SFT$\rightarrow$RAFT & \density{} & 18 (18) & $+21.1$ & $[+8.8, +33.2]$ & $.0020$ & \textbf{.0040} \\
&  & \coverage{} & 18 (18) & $+35.6$ & $[+19.3, +53.1]$ & $<.0001$ & \textbf{$<$.001} \\
&  & judge & 18 (18) & $+17.4$ & $[+3.8, +29.4]$ & $.0118$ & \textbf{.0118} \\
& E1.10 \alps{} RAFT vs.\ Turn RAFT & \density{} & 18 (18) & $+29.2$ & $[+19.2, +39.8]$ & $<.0001$ & \textbf{$<$.001} \\
&  & \coverage{} & 18 (16) & $+2.1$ & $[-0.3, +4.3]$ & $.0449$ & \textbf{.0449} \\
&  & judge & 18 (18) & $+20.7$ & $[+11.1, +30.9]$ & $.0003$ & \textbf{$<$.001} \\
\addlinespace[3pt]
\multicolumn{8}{@{}l}{\textit{\tuglet{} Exp1: prompt robustness (1 contrast)}} \\
& E1.13 \alps{} vs.\ Turn, numeric prompt (no examples) & \density{} & 18 (18) & $+19.9$ & $[+14.7, +25.1]$ & $<.0001$ & \textbf{$<$.001} \\
&  & \coverage{} & 18 (18) & $+3.8$ & $[+0.1, +7.3]$ & $.0300$ & \textbf{.0300} \\
&  & judge & 18 (17) & $+7.8$ & $[-1.6, +15.9]$ & $.0594$ & $.0594$ \\
\bottomrule
\end{tabular*}
\caption{\textbf{Exp1: data-sufficient test results, \tuglet{}.} \statsnote{$2{\times}10^{4}$}}
\label{tab:stats-exp1-tuglet}
\end{table*}
\begin{table*}[t]
\centering
\footnotesize
\setlength{\tabcolsep}{4pt}
\renewcommand{\arraystretch}{0.95}
\begin{tabular*}{\textwidth}{@{\extracolsep{\fill}} l L{5.0cm} l r r c r r @{}}
\toprule
& Contrast & Metric & $n$ (inf.) & $\Delta$ & 95\%\ CI & $p$ & $p_{\text{Holm}}$ \\
\midrule
\multicolumn{8}{@{}l}{\textit{\pharmasim{} Exp2: teacher guidance, Qwen3 (8 contrasts)}} \\
& E2.1 \alps{} SFT(\td) vs.\ \alps{} ICL & \density{} & 15 (15) & $+44.8$ & $[+29.2, +61.8]$ & $<.0001$ & \textbf{$<$.001} \\
&  & \coverage{} & 15 (11) & $+20.9$ & $[+2.8, +40.3]$ & $.0177$ & $.1060$ \\
&  & judge & 15 (14) & $+54.1$ & $[+37.8, +69.1]$ & $<.0001$ & \textbf{$<$.001} \\
& E2.2 Turn SFT(\td) vs.\ Turn ICL & \density{} & 15 (3)\textsuperscript{\ddag} & $-6.7$ & $[-17.2, +0.0]$ & $<.0001$ & \textbf{$<$.001} \\
&  & \coverage{} & 15 (5)\textsuperscript{\ddag} & $-7.1$ & $[-20.3, +0.0]$ & $.0043$ & \textbf{.0299} \\
&  & judge & 15 (2)\textsuperscript{\ddag} & $+0.0$\textsuperscript{\S} & $[-0.2, +0.2]$ & $.0084$ & \textbf{.0334} \\
& E2.3 \alps{} SFT$\rightarrow$RAFT vs.\ \alps{} SFT & \density{} & 15 (5)\textsuperscript{\ddag} & $-1.6$ & $[-6.3, +2.3]$ & $.0067$ & \textbf{.0269} \\
&  & \coverage{} & 15 (8)\textsuperscript{\ddag} & $-1.0$ & $[-1.8, -0.3]$ & $.0026$ & \textbf{.0210} \\
&  & judge & 15 (7)\textsuperscript{\ddag} & $+0.8$ & $[-0.3, +2.1]$ & $.9780$ & $.9780$ \\
& E2.4 \alps{} RAFT(\tr) vs.\ \alps{} ICL & \density{} & 15 (15) & $+24.0$ & $[+12.6, +36.5]$ & $.0009$ & \textbf{.0051} \\
&  & \coverage{} & 15 (12) & $+9.6$ & $[+1.5, +23.2]$ & $.1876$ & $.7505$ \\
&  & judge & 15 (13) & $+27.2$ & $[+17.9, +37.3]$ & $.0003$ & \textbf{.0021} \\
& E2.5 Turn RAFT(\tr) vs.\ Turn ICL & \density{} & 15 (9)\textsuperscript{\ddag} & $+5.2$ & $[+0.7, +9.7]$ & $.2524$ & $.5387$ \\
&  & \coverage{} & 15 (11) & $+4.5$ & $[-12.9, +22.0]$ & $.0413$ & $.2063$ \\
&  & judge & 15 (9)\textsuperscript{\ddag} & $+19.9$ & $[+7.1, +33.9]$ & $.2293$ & $.6879$ \\
& E2.6 \alps{} SFT(\td) vs.\ \alps{} RAFT(\tr) & \density{} & 15 (10) & $+20.8$ & $[+2.9, +40.5]$ & $.2444$ & $.5387$ \\
&  & \coverage{} & 15 (11) & $+11.2$ & $[-0.5, +28.1]$ & $.2807$ & $.8421$ \\
&  & judge & 15 (13) & $+26.9$ & $[+16.1, +38.1]$ & $.0006$ & \textbf{.0035} \\
& E2.7 Qwen3 \alps{} SFT(\td) vs.\ GPT-OSS \alps{} ICL & \density{} & 15 (5)\textsuperscript{\ddag} & $-1.5$ & $[-4.1, +0.1]$ & $.0054$ & \textbf{.0269} \\
&  & \coverage{} & 15 (8)\textsuperscript{\ddag} & $+0.1$ & $[-0.5, +0.7]$ & $.8469$ & $1.0000$ \\
&  & judge & 15 (12) & $+5.6$ & $[-0.9, +13.0]$ & $.2769$ & $.6879$ \\
& E2.13 \alps{} SFT$\rightarrow$RAFT vs.\ \alps{} RAFT(\tr) & \density{} & 15 (11) & $+19.2$ & $[-0.1, +39.9]$ & $.1796$ & $.5387$ \\
&  & \coverage{} & 15 (9)\textsuperscript{\ddag} & $+10.2$ & $[-1.8, +27.3]$ & $.6606$ & $1.0000$ \\
&  & judge & 15 (13) & $+27.7$ & $[+16.7, +39.2]$ & $.0006$ & \textbf{.0035} \\
\addlinespace[3pt]
\multicolumn{8}{@{}l}{\textit{\pharmasim{} Exp2: architecture under prompting (2 contrasts)}} \\
& E2.8 GPT-OSS: \alps{} vs.\ Turn & \density{} & 15 (12) & $+60.8$ & $[+38.8, +81.8]$ & $.0004$ & \textbf{$<$.001} \\
&  & \coverage{} & 15 (13) & $+24.0$ & $[+5.7, +43.8]$ & $.0002$ & \textbf{$<$.001} \\
&  & judge & 15 (12) & $+50.8$ & $[+22.3, +78.2]$ & $.0128$ & \textbf{.0256} \\
& E2.9 Qwen3: \alps{} vs.\ Turn & \density{} & 15 (15) & $+0.6$ & $[-5.8, +7.0]$ & $.4452$ & $.4452$ \\
&  & \coverage{} & 15 (14) & $-4.4$ & $[-18.4, +3.6]$ & $.0938$ & $.0938$ \\
&  & judge & 15 (12) & $-4.3$ & $[-14.5, +4.5]$ & $.5765$ & $.5765$ \\
\addlinespace[3pt]
\multicolumn{8}{@{}l}{\textit{\pharmasim{} Exp2: architecture at each adapted stage, Qwen3 (3 contrasts)}} \\
& E2.10 \alps{} SFT vs.\ Turn SFT & \density{} & 15 (12) & $+52.1$ & $[+31.8, +72.8]$ & $.0004$ & \textbf{.0013} \\
&  & \coverage{} & 15 (12) & $+23.6$ & $[+5.9, +43.2]$ & $.0017$ & \textbf{.0034} \\
&  & judge & 15 (10) & $+49.7$ & $[+27.9, +70.3]$ & $.0128$ & \textbf{.0384} \\
& E2.11 \alps{} RAFT vs.\ Turn RAFT & \density{} & 15 (13) & $+19.3$ & $[+3.6, +37.7]$ & $.0319$ & \textbf{.0319} \\
&  & \coverage{} & 15 (15) & $+0.8$ & $[-18.2, +19.9]$ & $.4452$ & $.4452$ \\
&  & judge & 15 (13) & $+2.9$ & $[-9.4, +16.5]$ & $.5980$ & $.5980$ \\
& E2.12 \alps{} SFT$\rightarrow$RAFT vs.\ Turn SFT$\rightarrow$RAFT & \density{} & 15 (12) & $+50.1$ & $[+28.5, +72.0]$ & $.0004$ & \textbf{.0013} \\
&  & \coverage{} & 15 (14) & $+22.7$ & $[+4.6, +42.4]$ & $.0010$ & \textbf{.0030} \\
&  & judge & 15 (11) & $+50.5$ & $[+28.4, +71.1]$ & $.0128$ & \textbf{.0384} \\
\bottomrule
\end{tabular*}
\caption{\textbf{Exp2: data-scarce test results, \pharmasim{}.} \statsnote{$2{\times}10^{4}$}}
\label{tab:stats-exp2}\label{tab:stats-rq3}
\end{table*}
\begin{table*}[t]
\centering
\footnotesize
\setlength{\tabcolsep}{4pt}
\renewcommand{\arraystretch}{0.95}
\begin{tabular*}{\textwidth}{@{\extracolsep{\fill}} l L{5.0cm} l r r c r r @{}}
\toprule
& Contrast & Metric & $n$ (inf.) & $\Delta$ & 95\%\ CI & $p$ & $p_{\text{Holm}}$ \\
\midrule
\multicolumn{8}{@{}l}{\textit{\tuglet{} Exp2: teacher guidance, Qwen3 (8 contrasts)}} \\
& E2.1 \alps{} SFT(\td) vs.\ \alps{} ICL & \density{} & 18 (18) & $+36.1$ & $[+20.7, +51.6]$ & $.0006$ & \textbf{.0045} \\
&  & \coverage{} & 18 (17) & $+28.4$ & $[+11.2, +47.7]$ & $.0033$ & \textbf{.0263} \\
&  & judge & 18 (18) & $+33.7$ & $[+27.4, +39.6]$ & $<.0001$ & \textbf{$<$.001} \\
& E2.2 Turn SFT(\td) vs.\ Turn ICL & \density{} & 18 (15) & $+8.1$ & $[-3.1, +20.2]$ & $.8317$ & $1.0000$ \\
&  & \coverage{} & 18 (15) & $-3.4$ & $[-5.3, -1.6]$ & $.0034$ & \textbf{.0263} \\
&  & judge & 18 (15) & $+15.1$ & $[+1.7, +29.5]$ & $.5226$ & $1.0000$ \\
& E2.3 \alps{} SFT$\rightarrow$RAFT vs.\ \alps{} SFT & \density{} & 18 (13) & $-0.4$ & $[-6.8, +4.5]$ & $.7019$ & $1.0000$ \\
&  & \coverage{} & 18 (7)\textsuperscript{\ddag} & $-6.1$ & $[-16.7, -0.3]$ & $.0066$ & \textbf{.0395} \\
&  & judge & 18 (16) & $+1.8$ & $[-2.7, +6.6]$ & $.7019$ & $1.0000$ \\
& E2.4 \alps{} RAFT(\tr) vs.\ \alps{} ICL & \density{} & 18 (15) & $+13.1$ & $[+7.3, +19.7]$ & $.0003$ & \textbf{.0026} \\
&  & \coverage{} & 18 (10)\textsuperscript{\ddag} & $+6.0$ & $[+0.1, +17.1]$ & $.5509$ & $.6495$ \\
&  & judge & 18 (14) & $+7.3$ & $[+3.7, +11.7]$ & $.0016$ & \textbf{.0079} \\
& E2.5 Turn RAFT(\tr) vs.\ Turn ICL & \density{} & 18 (15) & $+7.9$ & $[-2.8, +19.1]$ & $.2645$ & $1.0000$ \\
&  & \coverage{} & 18 (14) & $+21.5$ & $[+5.4, +40.6]$ & $.1815$ & $.5444$ \\
&  & judge & 18 (12) & $-3.4$ & $[-11.7, +5.6]$ & $.0208$ & $.0833$ \\
& E2.6 \alps{} SFT(\td) vs.\ \alps{} RAFT(\tr) & \density{} & 18 (18) & $+23.0$ & $[+5.7, +40.1]$ & $.0080$ & \textbf{.0478} \\
&  & \coverage{} & 18 (16) & $+22.4$ & $[+6.7, +40.7]$ & $.0134$ & $.0671$ \\
&  & judge & 18 (18) & $+26.4$ & $[+21.0, +31.4]$ & $<.0001$ & \textbf{$<$.001} \\
& E2.13 \alps{} SFT$\rightarrow$RAFT vs.\ \alps{} RAFT(\tr) & \density{} & 18 (18) & $+22.6$ & $[+4.2, +40.8]$ & $.0237$ & $.1184$ \\
&  & \coverage{} & 18 (15) & $+16.3$ & $[+2.5, +32.5]$ & $.1084$ & $.4335$ \\
&  & judge & 18 (18) & $+28.2$ & $[+21.9, +34.6]$ & $<.0001$ & \textbf{$<$.001} \\
& E2.7 Qwen3 \alps{} SFT(\td) vs.\ GPT-OSS \alps{} ICL & \density{} & 18 (14) & $-1.4$ & $[-6.1, +2.8]$ & $.7019$ & $1.0000$ \\
&  & \coverage{} & 18 (13) & $+19.3$ & $[+4.5, +37.6]$ & $.3247$ & $.6495$ \\
&  & judge & 18 (15) & $+6.1$ & $[+0.2, +11.8]$ & $.1187$ & $.3561$ \\
\addlinespace[3pt]
\multicolumn{8}{@{}l}{\textit{\tuglet{} Exp2: architecture under prompting (2 contrasts)}} \\
& E2.8 GPT-OSS: \alps{} vs.\ Turn & \density{} & 18 (18) & $+27.6$ & $[+13.5, +41.5]$ & $.0014$ & \textbf{.0028} \\
&  & \coverage{} & 18 (17) & $+4.8$ & $[-10.9, +18.6]$ & $.0216$ & \textbf{.0432} \\
&  & judge & 18 (17) & $+22.7$ & $[+13.5, +32.2]$ & $<.0001$ & \textbf{$<$.001} \\
& E2.9 Qwen3: \alps{} vs.\ Turn & \density{} & 18 (18) & $+8.4$ & $[+1.5, +15.9]$ & $.0269$ & \textbf{.0269} \\
&  & \coverage{} & 18 (14) & $+0.2$ & $[-3.5, +4.2]$ & $.7101$ & $.7101$ \\
&  & judge & 18 (13) & $-2.1$ & $[-11.7, +6.4]$ & $.5675$ & $.5675$ \\
\addlinespace[3pt]
\multicolumn{8}{@{}l}{\textit{\tuglet{} Exp2: architecture at each adapted stage, Qwen3 (3 contrasts)}} \\
& E2.10 \alps{} SFT vs.\ Turn SFT & \density{} & 18 (17) & $+36.4$ & $[+21.3, +52.7]$ & $.0002$ & \textbf{$<$.001} \\
&  & \coverage{} & 18 (17) & $+32.0$ & $[+14.3, +52.1]$ & $<.0001$ & \textbf{$<$.001} \\
&  & judge & 18 (17) & $+16.5$ & $[+4.8, +29.3]$ & $.0152$ & \textbf{.0455} \\
& E2.11 \alps{} RAFT vs.\ Turn RAFT & \density{} & 18 (18) & $+13.6$ & $[+0.7, +26.9]$ & $.0241$ & \textbf{.0241} \\
&  & \coverage{} & 18 (18) & $-15.3$ & $[-34.7, +4.5]$ & $.9406$ & $.9406$ \\
&  & judge & 18 (14) & $+8.7$ & $[-0.9, +18.9]$ & $.0837$ & $.0837$ \\
& E2.12 \alps{} SFT$\rightarrow$RAFT vs.\ Turn SFT$\rightarrow$RAFT & \density{} & 18 (18) & $+40.1$ & $[+20.5, +59.5]$ & $.0010$ & \textbf{.0019} \\
&  & \coverage{} & 18 (15) & $+23.6$ & $[+5.9, +43.9]$ & $.0333$ & $.0665$ \\
&  & judge & 18 (16) & $+18.8$ & $[+2.9, +35.0]$ & $.0333$ & $.0665$ \\
\bottomrule
\end{tabular*}
\caption{\textbf{Exp2: data-scarce test results, \tuglet{}.} \statsnote{$2{\times}10^{4}$}}
\label{tab:stats-exp2-tuglet}
\end{table*}

\paragraph{Data-Sufficient Analysis.} Taken together, the data-sufficient tests support four claims and withhold one.
Hierarchical generation is significant under prompting for the 120B model in both
environments but not for the 8B one. With the numeric profile description and no
in-context examples it is significant for the 8B model as well on \tuglet{}, in \density{}
($+19.9$) and \coverage{} ($+3.8$), though not in judge pass rate ($+7.8$, $p=.059$).
Adding SFT to \alps{} improves all three metrics in both environments. At every adapted
stage \alps{} beats its turn-level counterpart on the judge in both environments and on
\density{} in all but one of those six contrasts. What the tests withhold is any
distributional gain from RAFT. No adaptation contrast shows RAFT improving \density{} or
\coverage{} over prompting, SFT significantly outperforms RAFT on \coverage{} in both
environments (E1.11), and on the turn-level student RAFT significantly reduces \density{}
(E1.6b). Two RAFT contrasts do reach significance, and both are architecture comparisons
rather than adaptation gains, since E1.10 holds the adaptation method fixed and therefore
measures the hierarchy. The one adaptation-level RAFT effect is on the judge for \alps{} on
\pharmasim{} (E1.6a, $+26.6$), and it points the opposite way for the turn-level student,
whose reward-selected trajectories lose judge pass rate (E1.6b, $-15.7$).

\paragraph{Data-Scarce Analysis.} The data-scarce tests are simpler. Both teacher roles significantly raise \alps{} on
\density{} and the judge in both environments. The \coverage{} gain from demonstrations is
significant on \tuglet{} ($+28.4$) but not on \pharmasim{} ($+20.9$, $p=.106$).
Demonstrations beat the reward on the judge in both environments and on \density{} on
\tuglet{}. No turn-level contrast improves significantly anywhere, and teacher
demonstrations significantly reduce turn-level \coverage{} in both environments. Applying
RAFT after teacher SFT improves no metric and significantly reduces \coverage{} in both
($-1.0$ on \pharmasim{}, $-6.1$ on \tuglet{}), though both results rest on fewer than half
the pairs. On the architecture side, the hierarchy is significant for GPT-OSS on all three
metrics in both environments and, for Qwen3 under prompting, only on \tuglet{} \density{}
($+8.4$). At the adapted stages \alps{} exceeds its turn-level counterpart on \density{} in
all six contrasts, and on \coverage{} and the judge for SFT in both environments and
additionally for SFT$\rightarrow$RAFT on \pharmasim{}.
\section{Reproducibility Notes}
\label{app:reproducibility}

This section describes the released artifacts, verification of the feature-extraction pipeline, sources of randomness, and computational environment. Together with the implementation and hyperparameter details in Appendix~\ref{app:experimental-details}, these materials support reproduction of the generation and evaluation pipeline without releasing the original student interaction logs.

\subsection{Data and Code Availability}
\label{app:availability}

Our code release will include our code, prompts, agent-compatible implementations of both learning environments, and the student populations generated by our best-performing SFT-adapted models. For \pharmasim{}, we adapted the existing PharmaSimText implementation~\cite{PharmaSimText2025} to support interaction with simulated students.

The full original student interaction logs cannot be released due to data-protection restrictions. However, we released the $\dfew$ examples and to support evaluation of newly generated student populations without exposing these logs, we provide the behavioral feature representations used as the real-data evaluation reference. These representations were obtained through an irreversible transformation and cannot be used to reconstruct individual trajectories. The trained adapters will be released in a subsequent version of the code release.

\subsection{Compute Environment}
\label{app:compute-environment}

Experiments ran on a Linux Kubernetes cluster using NVIDIA A100-SXM4 GPUs with 80\,GB of memory. Each adapter-training job used one GPU, while inference and evaluation deployments used eight GPUs. Jobs requested one CPU and 10\,GiB of RAM. The software environment used Python~3.12 with PyTorch~2.11.0, Transformers~4.57.2, PEFT~0.19.1, TRL~0.23.0, Unsloth~2025.11.1, Accelerate~1.13.0, Datasets~4.8.4, NumPy~2.4.4, SciPy~1.17.1, and scikit-learn~1.8.0. Models were served with vLLM using a GPU-memory-utilization setting of 0.90 or 0.92.

\end{document}